\documentclass[runningheads]{llncs}
\usepackage{float}
\usepackage{amssymb,amsmath,array}
\usepackage{subcaption}
\usepackage{graphicx}
\graphicspath{{./figs/}}
\DeclareGraphicsExtensions{.pdf,.png,.jpg,.jpeg, .eps}
\usepackage{geometry}
\usepackage{fancyhdr}   
\usepackage{tabularx,multicol,booktabs,multirow,makecell}[width=\textwidth]
\usepackage{longtable}
\usepackage{xfrac}
\usepackage[square,sort,comma,numbers]{natbib}
\usepackage{hyperref}

\usepackage{enumitem}
\setlist[enumerate]{noitemsep,topsep=0pt} 
\setlist[itemize]{noitemsep,topsep=0pt}

\usepackage{color}

\begin{document}
\title{Approximate Unlearning in Prototype-based-classifiers}
%
%
\author{Sreejita Ghosh\inst{1,2}\orcidID{0000-0001-7865-2736} \and
Mohammad Mohammadi\inst{3}\orcidID{0000-0002-1984-2107} }
\authorrunning{S.Ghosh et al.}
%
\institute{UMC Groningen, Groningen, The Netherlands\and
Technical University of Eindhoven, Eindhoven, The Netherlands\\
\email{ghosh.sreejita@gmail.com}\and
Tilburg University, Tilburg, The Netherlands\\
\email{mohammadimathstar@gmail.com}}
\setlength{\belowcaptionskip}{-1pt}
\setlength{\abovecaptionskip}{-1pt}
\captionsetup{belowskip=0pt}
\renewcommand{\vec}[1]{\boldsymbol{#1}}
\newcommand{\R}{\rm I\!R}
\maketitle              
\begin{abstract}
The recent data and AI regulations and governance needs have highlighted the need for removing specific data even from models trained on or with them. While retraining the entire model is a theoretically trivial solution, it is computationally expensive, thereby motivating research in Machine Unlearning. This paper introduces a robust approximate unlearning technique for the family of prototype-based classifiers called Generalized Learning Vector Quantization, aimed at removal of effects of specified samples, either due to deletion request or due to them being identified as noisy or corrupt. We first show the impact of the different sizes of the forget set, as well as the characteristics of the data in this set, on the unlearned model. To address the sources of instabilities detected in this unlearning method we propose an adapted version of this technique, such that the unlearning is robust to the characteristics of the forget set. Thereafter using the intrinsic interpretability property of GLVQ and appropriate metrics defined in unlearning literature, we show the efficacy ($\approx0$ deviation of parameters and performance of unlearned model relative to retrained model) and efficiency ($6\times$ for smallest dataset to $10^{5}\times$ in the largest dataset) of the proposed unlearning techniques compared to naive retrained, on six benchmark datasets for tabular data recommended by literature. Although in this work we investigated only binary class problems, the proposed robust unlearning strategy can be readily applied for removing sample effects from any trained GLVQ model, i.e., it does not require any modification in the training scheme of GLVQ.

\end{abstract}
\section{Introduction}
In the recent years Machine Learning (ML) and other AI have become integral to applications in every sector of industry, along with in government. This has been possible due to mass digitization of data and electronic ``book-keeping'' in every human-centric applications, as well as increasing number of services that link personal data of users directly or via third-parties, often unbeknownst to the user (the data source). While this has enabled faster response-times, cost reductions (at least in the short term), and the option of having a \emph{secondary external} memory (``Humans forget, machines remember'' \cite{Villaronga2018304machinesremeber}), such convenience came at the cost of user privacy, and even security when attackers misuse the vulnerabilities of the inference (deployed) models to cause data breaches\cite{zhang2023reviewMU, nguyen2025surveyMU}. These concerns led to data protection laws and regulations worldwide, of which the European Union (EU)'s General Data Protection Regulations (GDPR) is noteworthy, specifically its Article 17 (\emph{the right to be forgotten}) \cite{nguyen2025surveyMU, wang2024comprehensiveMU, zhang2023reviewMU, golatkar2020eternalsunshine}. According to Article 17, the request by any user to have their data deleted must be immediately complied with. While adhering to it is straightforward in terms of deleting raw data from databases, its implementation in terms of meta-data such as ML and AI models that trained on the data poses is a non-trivial problem. The increasing frequency of external attacks on trained/ inference models, and the data-deletion requests from users exercising GDPR Article 17, led to the emergence of a new paradigm called \emph{Machine Unlearning} (MU), which focus on this problem, i.e., on developing techniques that enable machines to \emph{forget} \cite{Villaronga2018304machinesremeber, kumar2026controllingMU, nguyen2025surveyMU, zhang2023reviewMU, wang2024comprehensiveMU, bourtoule2021MU}.   

The recent surveys and review papers\cite{li2025overviewMU, wang2024comprehensiveMU, zhang2023reviewMU, brophy2021UnlearnRF, Villaronga2018304machinesremeber} list \emph{usability} and \emph{fidelity}, in addition to the aforementioned concerns of \emph{security} and \emph{privacy}, as reasons that motivate users to want their data to be deleted, and thereby motivation behind Machine Unlearning. While the naive way to forget such data, be it data related to specific subjects, features, class(es), or tasks that the ML model had learned (types of \emph{unlearning requests}), is to retrain the entire model, it is expensive in terms of time and resources, especially when deletion requests arrive one at a time, and requires access to the original dataset. The high computational and storage costs related to retraining, an \emph{exact unlearning} technique, necessitated research in \emph{approximate unlearning}, as well as \emph{few-shot, zero-shot} and \emph{zero-glance} unlearning\cite{bourtoule2021MU, li2025overviewMU, nguyen2025surveyMU, wang2024comprehensiveMU, zhang2023reviewMU}. These works have also summarized and categorized the unlearning algorithms in terms of their application domain (Recommendation systems, federated learning, lifelong learning, large language models, generative models, graph embedding, gradient-based, tree-based, causal unlearning, statistical query-based, amnesiac unlearning, among others), type of data (tabular, image, text, stream, multi-modal), relevant evaluation metrics (including measures closely oriented with security threats and model vulnerabilities), as well as approach (model-agnostic vs model-intrinsic vs data-driven)\cite{nguyen2025surveyMU, li2025overviewMU}. In this paper we introduce an approximate unlearning strategy, to remove influence of specified samples from the nearest prototype based classifier model called Generalized Learning Vector Quantization (GLVQ)\cite{sato}, whose training set included these samples. During training, GLVQ and its extensions learn representative examples (known as \emph{prototypes}, denoted by $\vec{}w$) of each class in the training set , by updating the prototype towards or away from a presented sample based on the match between the ground-truth label of this sample and the label of the prototypes\cite{Hammer20021059, NaNLVQ,ghosh2020visualisation, ghosh2025interpretable, mohammadi2024prototype}. 
Although this family of classifier is capable of learning from image as well\cite{mohammadi2024generalized, mohammadi2024prototype}, in this paper we focus only on six tabular datasets mentioned in \cite{brophy2021UnlearnRF, wang2024comprehensiveMU, nguyen2025surveyMU}. 

\subsection{Our contribution}\label{ssec:contribution}
Training of GLVQ and its variants involve minimization of a cost-function following a gradient descent strategy (stochastic vs batch, depending on dataset size and application), that direct prototype updates. Thus influence of each training sample on the trained model can be inferred from the gradients between the sample and the prototypes. Our proposed approximate unlearning of GLVQ model involve computing the influence of the specified samples and removing them from the prototypes. As the distance between a sample and the prototypes dictate the update size and direction, and unlearning should minimize catastrophic forgetting in the updated (unlearned) model, our contributions include the following:
\begin{itemize}[label=$\bullet$]
    \item investigating the deviation between the model performances, and parameters (prototypes) of the unlearned and a retrained GLVQ model, when the samples to be unlearned are `randomly` selected, such that there is class homogeneity vs heterogeneity (i.e., set of samples to be forgotten span across different classes); 
    \item investigating the deviation between an unlearned and retrained GLVQ model over different batch sizes and locations of samples to be unlearned, in terms of model parameters and their performance;
    \item compensating for deleting the influence of randomly selected samples on a trained GLVQ model, by adding influence of equal amount of `randomly' selected samples from the retained (or `remembered') dataset, such that these `enforcer sample' set have the same class distribution as that of the unlearned set.  
\end{itemize}
 Although we present our findings based on the distance metric squared euclidean distance, the same unlearning strategy can be extended to other dissimilarities, like cosine\cite{ghosh2020visualisation, ghosh2025interpretable} or chordal distance\cite{mohammadi2024prototype} in a straightforward manner. 
 
\subsection{Organization of the paper}
The rest of this paper is organized as follows: \autoref{sec: rel work} briefly explains Random Forest (RF) and different unlearning strategies in the context of RF, followed by a brief introduction of Generalized LVQ, and its mathematical formulations that we exploit for unlearning. In \autoref{sec:Unlearning} we introduce the unlearning framework for GLVQ models; \autoref{sec:Exp} first explains the evaluation metrics of unlearning techniques applicable for our novel contribution in this paper, the preprocessing steps and experimental setups to evaluate the efficacy of our proposed strategy, the outcomes of which are showcased in \autoref{sec:Res}, and discussed in \autoref{sec:Disc}. Finally in \autoref{sec:Conclusion} we present our concluding remarks along with limitations of our study and future directions. 

\section{Related Works}\label{sec: rel work}
The stochastic gradient descent (SGD) based unlearning in literature are mostly for image-based problems \cite{neel2021descentdelete, thudi2022unrollingSGD, chundawat2023zeroMU, neel2021descentdelete}, and of these \cite{neel2021descentdelete, chundawat2023zeroMU} investigate class removal. Other noteworthy model-intrinsic unlearning strategies for models trained on tabular data and aimed at sample removal, include unlearning in tree-based models DeltaBoost, Hedge-cut for Extreme Boost RF\cite{schelter2021hedgecut}, and DaRE-RF \cite{brophy2021UnlearnRF} and Bayesian Models\cite{nguyen2025surveyMU}. Although DaRE-RF investigated the impact of removal of random and ``1000 worst samples'', impact of the different sizes of deletion requests (size of unlearn batch) on the performance of unlearned model have not been reported. Thus even though we selected our benchmark datasets based on \cite{brophy2021UnlearnRF} and \cite{nguyen2025surveyMU}, and closely followed the preprocessing as described in DaRE-RF, we could not directly compare our proposed unlearning frameworks for GLVQ against it. 
Nevertheless, the theoretical concepts and metrics for evaluating unlearning discussed in these works motivated our problem formulation and scope definition. To the best of our knowledge, unlearning strategies for the LVQ family of classifiers have not yet been investigated at the time of writing this paper. 

\subsection{Unlearning in tree-based models}\label{ssec:unlearn RF}
\cite{li2025overviewMU, nguyen2025surveyMU} discuss tree-based unlearning strategies for extreme Random Forest (eRF), of which Hedgecut\cite{schelter2021hedgecut} introduced a reconfigured learning strategy for the constituent extreme randomized trees (ERTs), such that the splitting thresholds (at decision nodes) are robust against removal of $k$ random samples. This reconfiguration ensures that the nodes at higher levels are stable, and that any retraining due to change in the splitting thresholds in the lower-level decision nodes are limited to the much smaller sub-trees. However the design of Hedgecut is based on the assumption of very few and infrequent data deletion requests. \cite{brophy2021UnlearnRF} introduces \textbf{Data Removal-Enabled} (DaRE) RF as an alternative exact unlearning strategy, which also aligns with the logic of retraining of minimal subtrees, but capable of handling larger deletion requests. Also unlike Hedgecut, DaRE-RF is not limited to only the ERT, but is based on a the standard RF algorithm excluding the bootstrapping. In our work in this paper we tried to closely follow the design requirement of model, benchmark datasets (and their preprocessing) as defined in \cite{brophy2021UnlearnRF}.

\subsection{Prototype based classifiers}\label{ssec:LVQ}
Let $\vec{\mathcal{X}}_i^D,y_i$ where $i\in[1,..N]$ denote a $D-$dimensional dataset ($\vec{\mathcal{X}}_i \in \mathbb{R}^D$) with $N$ training instances, each with label (ground-truth) $y_i\in[1,..C]$, where $C$ refers to the classes that a model will be trained to distinguish. Generalized Learning Vector Quantization is a nearest prototype classifier (NPC) that assigns a presented sample $\vec{\mathcal{X}}_i$ the label associated with the prototype closest to it. This dissimilarity-based label assignment is similar to that in K-Nearest neighbor (KNN) classifiers. Each class $c$ that GLVQ is trained on is represented by one or more (as defined by user) representative examples called prototypes, $\vec{w}^c \in \mathbb{R}^D$ (i.e., $\vec{w}$ lie in the same $D-$dimensional input same as $\vec{\mathcal{X}}$). Therefore, unlike in KNN classification which requires pairwise dissimilarity computation between presented sample and all $N$ samples of training set, GLVQ's decision requires computation of dissimilarity between presented sample and only the $n$ learned prototypes, which makes it computationally more efficient during inference, and increases its robustness against outlier samples. Thus GLVQ belong to the family of prototype based classifiers (PBCs). Consequently this also curbs the need for storage of explicit training samples for inference, thereby reducing storage needs and improving resilience against external attacks aimed at access to explicit training dataset. 

At the end of training, the prototypes should be such that the distance between any sample belonging to a class $c$ and prototype(s) representing this class is much smaller than that between the sample and prototypes of the other (incorrect) classes (exhibiting large margin principle). This is achieved by minimization of the cost-function 
{\abovedisplayskip=0pt
\belowdisplayskip=1pt
\begin{align}
    E=\frac{1}{N}\sum_i f(\mu_i), \text{ where } \mu_i=\frac{d_i^J-d_i^K}{d_i^J+d_i^K} \enspace ,
    \label{eqn:cf-lvq}
\end{align}
}
and $d_i^L, L=[J,K]$ denotes the dissimilarity between sample $\vec{\mathcal{X}_i}$ and its closest correct ($y_i=c(\vec{w}^J)$) and incorrect ($y_i \neq c(\vec{w}^K)$) prototypes. The activation function $f$ can be identity ($\mu=f(\mu)$), soft-max, swish, sigmoid, among others, as explained in \cite{vanVeen2021}. However in this work we used `swish' only, to appropriately handle the log transformation in data normalization. This activation function is given by 
{\abovedisplayskip=0pt
\belowdisplayskip=1pt
\begin{align}
        f(\mathbf{\mu}_i) = \frac{\mu_i}{1 + e^{-\beta \cdot \mathbf{\mu}_i}} \enspace .
        \label{eqn:swish}
\end{align}
}
The cost-function represented by Eq. \eqref{eqn:cf-lvq} emphasizes the choice of dissimilarity metrics in influencing the success of the LVQ method in learning \cite{ghosh2025interpretable}. 
The dissimilarity can be the absolute (squared) Euclidean, cosine, Chordal, as well as partial distance strategy based variants\cite{NaNLVQ} of these and their feature-relevance weighted adapted version, as seen in Generalized Relevance LVQ \cite{Hammer20021059} and its matrix-based variants \cite{schneider2009, ghosh2025interpretable, mohammadi2024prototype}. For this paper we focus on Squared Euclidean distance, i.e., 
{\abovedisplayskip=0pt
\belowdisplayskip=2pt
\begin{align}
    d_i^L(\vec{\mathcal{X}}_i, \vec{w}^L)=(\vec{\mathcal{X}}_i-\vec{w}^L)^T(\vec{\mathcal{X}}_i-\vec{w}^L) \enspace .
    \label{eqn:distSE}
\end{align}
}
The prototypes can be initialized randomly as seen in \cite{mohammadi2024generalized, mohammadi2024prototype}, or they can be initialized near the class conditional means, such as in \cite{schneider2009, ghosh2020visualisation, ghosh2025interpretable, NaNLVQ}. 
During training, for every sample $\vec{\mathcal{X}_i}$ presented to a prototype $\vec{w}$ during training, $\vec{w}$ is updated by being moved towards $\vec{\mathcal{X}_i}$ if $\vec{w}=\vec{w}^J$, and moved away from the sample otherwise. The update step :
{\abovedisplayskip=1pt
\belowdisplayskip=2pt
\begin{equation}
 \vec{w}^{L,new}=\vec{w}^{L,old} - \alpha . \Delta \vec{w}^L \enspace  \text{where } L\supset [J,K].   \label{eqn:prot}
\end{equation}
}
and $\alpha$ denotes the learning rate, 
and $\Delta \vec{w}^L$ denotes the derivative of the cost-function from Eqn.\eqref{eqn:cf-lvq}, with respect to $\vec{w}^L$, such that $\Delta \vec{w}^L$ is scaled with $-1$ when $L=J$, to emulate attraction towards samples belonging to the same class that $\vec{w}$\cite{schneider2009}. Thus for squared euclidean distance,
{\abovedisplayskip=1pt
\belowdisplayskip=2pt
 \begin{align}
  \frac{\partial d_L(\vec{\mathcal{X}}_i)}{\partial \vec{w}^L}=& -2(\vec{\mathcal{X}}_i-\vec{w}^L) \text{ and, } \label{eqn:sampInfluence} \\
 \Delta \vec{w}^J=&\frac{\partial E_{GLVQ}}{\partial \vec{w}^{J}}= \frac{2 d_K(\vec{\mathcal{X}}_i)}{\big(d_J(\vec{\mathcal{X}}_i)+d_K(\vec{\mathcal{X}}_i)\big)^2} \cdot \frac{\partial d_J(\vec{\mathcal{X}}_i)}{\partial \vec{w}^J} =\underbrace{\frac{-4 d_K(\vec{\mathcal{X}}_i) (\vec{\mathcal{X}}_i-\vec{w}^J)}{\big(d_J(\vec{\mathcal{X}}_i)+d_K(\vec{\mathcal{X}}_i)\big)^2}}_{\text{ force pulling $\vec{w}^J$ in}} \enspace, \label{eqn:wLupdate} \\
  \Delta \vec{w}^K=&\frac{\partial E_{GLVQ}}{\partial \vec{w}^{K}}= \frac{-2 d_J(\vec{\mathcal{X}}_i)}{\big(d_J(\vec{\mathcal{X}}_i)+d_K(\vec{\mathcal{X}}_i) \big)^2} \cdot \frac{\partial d_K(\vec{\mathcal{X}}_i)}{\partial \vec{w}^K}=\underbrace{\frac{4 d_J(\vec{\mathcal{X}}_i) (\vec{\mathcal{X}}_i-\vec{w}^K)}{\big( d_J(\vec{\mathcal{X}}_i)+d_K(\vec{\mathcal{X}}_i) \big)^2}}_{\text{force pushing $\vec{w}^K$ out}} \label{eqn:wKupdate}
 \end{align}
 }
 Thus a prototype of a class-C is updated based on the cumulative of the attraction and repulsion forces of all the samples for which it is the closest correct and closest incorrect prototype respectively, i.e., $\vec{w}^{C,new}=\vec{w}^{C,old}-\alpha\cdot (\Delta \vec{w}^{J=C}+\Delta \vec{w}^{K=C})$. 
For (single) batch gradient descent based optimization, such as what we followed in this paper, the updates happens per epoch, where $\Delta \vec{w}^L=\frac{1}{N}\sum_{i=1}^N\frac{\partial E_{GLVQ}(\vec{\mathcal{X}}_i)}{\partial \vec{w}_{L}}$, thus order in which training samples were presented, become irrelevant. For the activation function given by Eq. \eqref{eqn:swish}, the corresponding gradient is given by 
{\abovedisplayskip=1pt
\belowdisplayskip=2pt
\begin{align}
\frac{\partial f}{\partial \mathbf{\mu}_i} = \beta \cdot f(\mathbf{\mu}_i) + \Big(\frac{1}{1 + e^{-\beta \cdot \mathbf{\mu}_i}}\Big) \cdot (1 - \beta \cdot f(\mathbf{\mu}_i)) \enspace ,
\label{eqn: swish-grad}
\end{align}
}
which should be multiplied to Eqn. \eqref{eqn:wLupdate} and \eqref{eqn:wKupdate} for updating the model parameters (here, the prototypes). 

\subsubsection{Interpretability of GLVQ.}
Being a PBC, GLVQ models are not only efficiently scalable, they enable users to verify whether the representation of each class that they trained on make sense (global interpretability), thereby enhancing reliability and trust. In the context of personalized recommendations and predictions, the computed relative proximity between a subject or sample $\vec{\mathcal{X}}_i$ and the learned prototypes for the outcome-of interest $y$ (local interpretability) can not only help end-users in understanding the model's decision about them, but also motivate them towards their desired outcome, such as in the use-case of treatment-monitoring (e.g., patient profile moving from prototype representing high right towards that representing low risk). In case of poor performance of models on the training sets, the proximity and gradients (contribution) of the training samples with respect to the learned prototypes, given by Eqns. \eqref{eqn:wLupdate} and \eqref{eqn:wKupdate}, enable efficient detection of the cause of model under-fitting. In case of consistent low performance of a previously high performing deployed GLVQ model, this property of GLVQ enables the swift detection and addressing of model drifts, outlier and biased samples. 
 In this paper we therefore exploit these sample-wise contribution towards each of the prototypes (similar to the \emph{Influence method} mentioned in \cite{zhang2023reviewMU}), for a fast, efficient, yet cost-effective unlearning. 

\subsection{Unlearning vs retraining setup}
\begin{figure}
    \centering
    \includegraphics[trim={0.1cm 3cm 1cm 0.3cm}, clip, width=\textwidth]{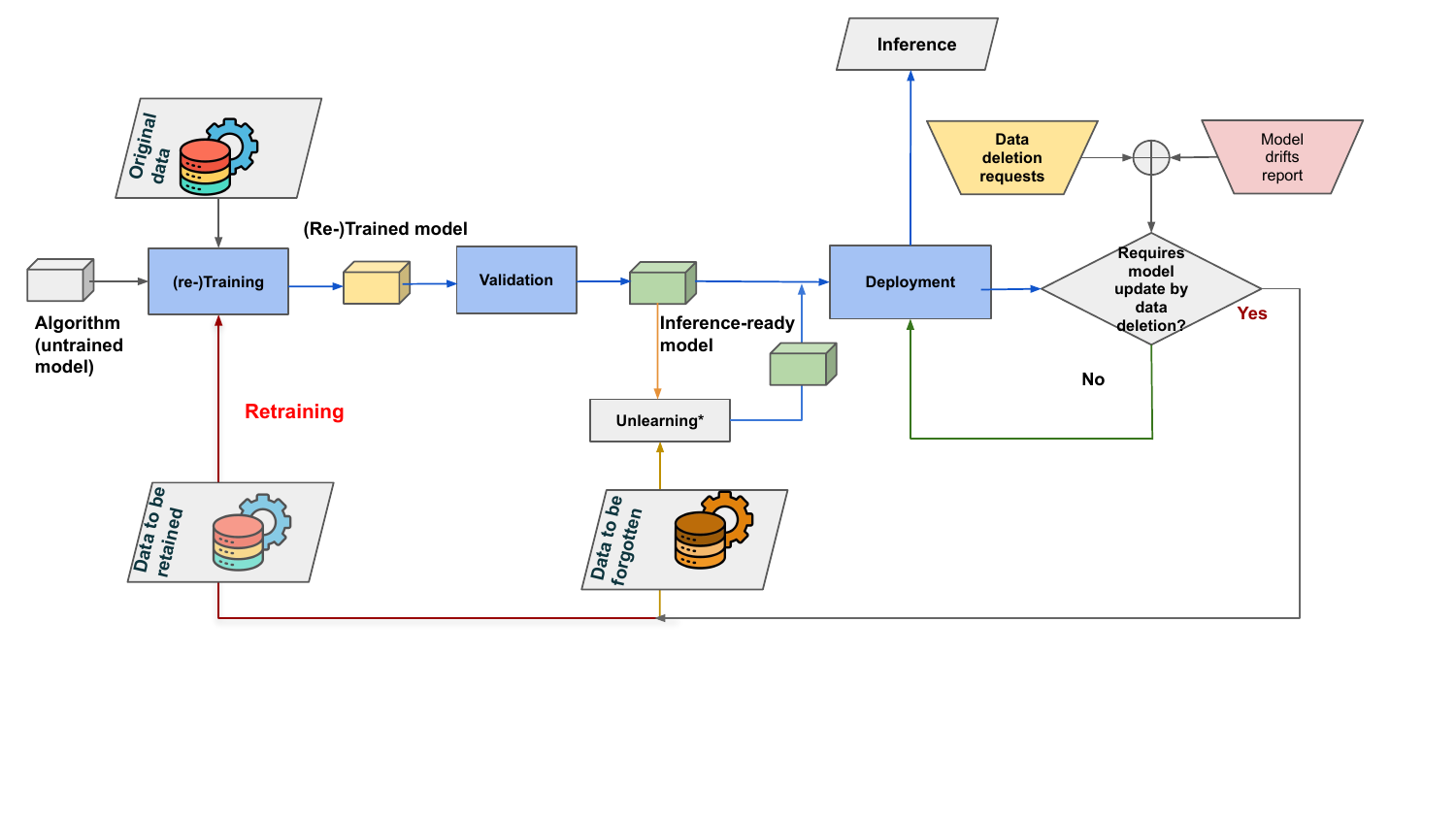}
    \caption{Unlearning vs retraining a model within the ML lifecycle.}
    \label{fig:unlearning}
\end{figure}

\autoref{fig:unlearning}, which is inspired from Figure 5 of \cite{zhang2023reviewMU}, illustrates the approximate points in the ML lifecycle where model retraining and unlearning takes place. 

\section{Unlearning for NPCs}\label{sec:Unlearning}
In this contribution we present a fast, low-cost model-intrinsic Machine Unlearning strategy for GLVQ models. The mathematical notations in this paper are based on those found in review and survey papers of MU \cite{li2025overviewMU, nguyen2025surveyMU, wang2024comprehensiveMU}, such that $\mathcal{Z}$ refers to the whole data space, which encompasses the training dataset $\mathcal{D}$ ($\mathcal{D}\subseteq \mathcal{Z}$), $\mathcal{A}$ refer to the inference model on which the unlearning is to be performed, $\mathcal{U}$ is the unlearning function for $\mathcal{A}$ to remove \emph{data to be forgotten, i.e., `forget'/ unlearned set}, $\mathcal{D}^f$. Note that $\mathcal{D}=\mathcal{D}^f\cup \mathcal{D}^r$, where $\mathcal{D}^r$ refers to the retained (or `remembered') data. 
While exact unlearning ensures Eqn. \eqref{eqn:unlearning-retraining} \cite{li2025overviewMU, wang2024comprehensiveMU, nguyen2025surveyMU}, for unlearning, 
{\abovedisplayskip=1pt
\belowdisplayskip=2pt
\begin{equation}
P(Y|\mathcal{U}(\mathcal{D}^f, \mathcal{A}(\mathcal{D})), \vec{\mathcal{X}})=P(Y|\mathcal{A}(\mathcal{D}^r), \vec{\mathcal{X}}) \enspace , \text{ where } (\vec{\mathcal{X}},Y)\subseteq \mathcal{D}
\label{eqn:unlearning-retraining}
\end{equation}
}
this condition can be relaxed from an exact equality to the minimization of the Kullbach Leibler divergence ($D_{KL}$) between outcomes predictions (estimated $Y$) from the unlearned GLVQ model $\mathcal{U}(\mathcal{D}^f, \mathcal{A}(\mathcal{D})$ and those from a GLVQ model retrained on $\mathcal{D}^r$, i.e., $\mathcal{A}(\mathcal{D}^r)$, given by:
{\abovedisplayskip=1pt
\belowdisplayskip=2pt
\begin{align}
    H=\frac{1}{N}\sum_{i=1}^N D_{KL} (P(Y_i|\mathcal{U}(\mathcal{D}^f, \mathcal{A}(\mathcal{D})), \vec{\mathcal{X}}_i) || P(Y_i|\mathcal{A}(\mathcal{D}^r), \vec{\mathcal{X}}_i)) \enspace ,
    \label{eqn:KLdiv}
\end{align}
}
which is Bayesian-based unlearning\cite{wang2024comprehensiveMU}.

\subsection{Unlearning contribution of specified samples}

As the prototypes of the GLVQ model presented here (and the Sklvq toolbox) are initialized near the class-means or medians (Eq. \eqref{eqn:prot_init}), unlearning involves first removing the impact of the samples in $\mathcal{D}^f$ from the prototypes. 
{\abovedisplayskip=1pt
\belowdisplayskip=1pt
\begin{align}
    \vec{w}_0^{L=c}&=\frac{1}{N_{Y=c}}\sum_{i=1}^{N_{Y=c}} \vec{\mathcal{X}}_i , \quad \forall  (\vec{\mathcal{X}}_i, Y_i) \subset \mathcal{D} \text{ and }\enspace Y_i= c .\label{eqn:prot_init}\\
    \vec{w}_{0,f}^{L=c}&=\frac{1}{N_{Y=c}-N_{Y=c}^f} \Big(\vec{w}_0^{L=c}\times N_{Y=c}-\sum_{f=1}^{N_{Y=c}^f} \vec{\mathcal{X}}_f\Big), \enspace \forall (\vec{\mathcal{X}}_f, Y_f) \subset \mathcal{D}^f, \quad \text{and }  Y_f=c
\end{align}
}
where $N_{Y=c}$ denotes the total number of samples with label $Y$=c in $\mathcal{D}$, $\vec{w}_0^{L=c}$, the initialized prototype of class-c in the original model $\mathcal{A}(\mathcal{D})$,  $\vec{w}_{0,u}$, the version of $\vec{w}_0^{L=c}$ with the influence of samples with this class label in $\mathcal{D}^f$, i.e,  $(\vec{\mathcal{X}}_f, Y_f) \forall Y_f=c$ removed. Thereafter, contribution of each training sample $\vec{\mathcal{X}}_i \subset \mathcal{D}^f$ to updating the initialized prototypes $\vec{w}_0^{L=c}$ (c=[1,C]) to their final state $\vec{w}^{L=c}$ (Eqn. \eqref{eqn:prot}) is unlearned via a \emph{gradient ascent step} based on Eqn. \eqref{eqn:wLupdate} and \eqref{eqn:wKupdate}, with the update scaling factor reversed for $\vec{w}^J$ and $\vec{w}^K$, as illustrated in Eqns. \eqref{eqn:wJascent} and \eqref{eqn:wKascent}. 
{\abovedisplayskip=1pt
\belowdisplayskip=2pt
\begin{align}
\vec{w}_f^{L}=& \vec{w}_o^L+\alpha \cdot \Delta \vec{w}^L\\
    \vec{w}_f^{J}=& \vec{w}_o^{J} + \alpha \cdot\Bigg[ \underbrace{\frac{2 d_K(\vec{\mathcal{X}}_i)}{d_J(\vec{\mathcal{X}}_i)+d_K(\vec{\mathcal{X}}_i)} \cdot (-2(\vec{\mathcal{X}}_i-\vec{w}_0^J))}_{\Delta \vec{w}^J}\Bigg] \enspace, \label{eqn:wJascent} \\
 \vec{w}_f^K =& \vec{w}_0^K +\alpha \cdot \Bigg[\underbrace{\frac{-2 d_J(\vec{\mathcal{X}}_i)}{d_J(\vec{\mathcal{X}}_i)+d_K(\vec{\mathcal{X}}_i)} \cdot (-2(\vec{\mathcal{X}}_i-\vec{w}_0^K))}_{\Delta \vec{w}^K}\Bigg] \label{eqn:wKascent}
\end{align}
}
In case of activation function other than identity, the Eqns. \eqref{eqn:wJascent} and \eqref{eqn:wKascent} need to be multiplied with the gradient of the corresponding activation function. For the work in this paper, this factor is given by Eqn. \eqref{eqn: swish-grad}. As the order of training samples presented do not have any impact in batch updates (with single batch), the challenges of \emph{incrementality} and \emph{stochasticity of training} \cite{nguyen2025surveyMU, wang2024comprehensiveMU} are not applicable in the unlearning scheme presented here. 

Although training a GLVQ as $\mathcal{A}$ requires iterating over more than a single epoch to ensure stabilization of the model parameters, which in our case are the prototypes $\vec{w}$, by optimizing the gradients (both value and direction), unlearning requires the reversal of the final gradient with respect to a sample. As seen in \cite{thudi2022unrollingSGD, kumar2026controllingMU, gupta2021adaptiveMU} our goal is to ensure that the deviations between the parameters from $\mathcal{U}(\mathcal{D}^f, \mathcal{A}(\mathcal{D}))$, i.e., $ \vec{w}_f^{L}$, and those from a GLVQ model retrained on just $\mathcal{D}^r$ is minimum, which is essentially extending Eqn.\eqref{eqn:KLdiv} from outcome predictions to parameter space: $D_{KL}(\mathcal{U}(\mathcal{D}^f, \mathcal{A}(\mathcal{D})) ||\mathcal{A}(\mathcal{D}^r))$. While such optimization enable verification of removal of all influences of samples from $\mathcal{D}^f$ from $\mathcal{A}(\mathcal{D})$, it requires access to a retrained model, which obviates the need for unlearning, as also noted in \cite{thudi2022unrollingSGD}. To prevent the degradation of the predictive performance of $\mathcal{U}(\mathcal{D}^f, \mathcal{A}(\mathcal{D}))$ the minimization of performance deviation (optimization) can be done with regards to the original model $\mathcal{A}(\mathcal{D})$ \cite{wang2024comprehensiveMU}, however, it cannot provide a reliable verification of successful unlearning, especially if $\mathcal{D}^r$ still consist of enough samples that are similar to the removed samples\cite{li2025overviewMU}, or if the removed samples contributed very less to the original training. As our objective in this paper has been to investigate whether unlearning was effective, rather than improve predictive performance of $\mathcal{U}(\mathcal{D}^f, \mathcal{A}(\mathcal{D}))$, we did not try this optimization. Note that, whether the `forgetting' of the samples in $\mathcal{D}^f$ has been unsuccessful due to presence of other similar samples in $\mathcal{D}^r$, can be verified by computing Eqns.\eqref{eqn:wLupdate} and \eqref{eqn:wKupdate} for the samples of interest. 

\subsection{A robust, fast unlearning framework}\label{ss:robustUnlearning}
\begin{figure}
    \centering
    \includegraphics[trim={0.1cm 3cm 1cm 0.3cm}, clip, width=\textwidth]{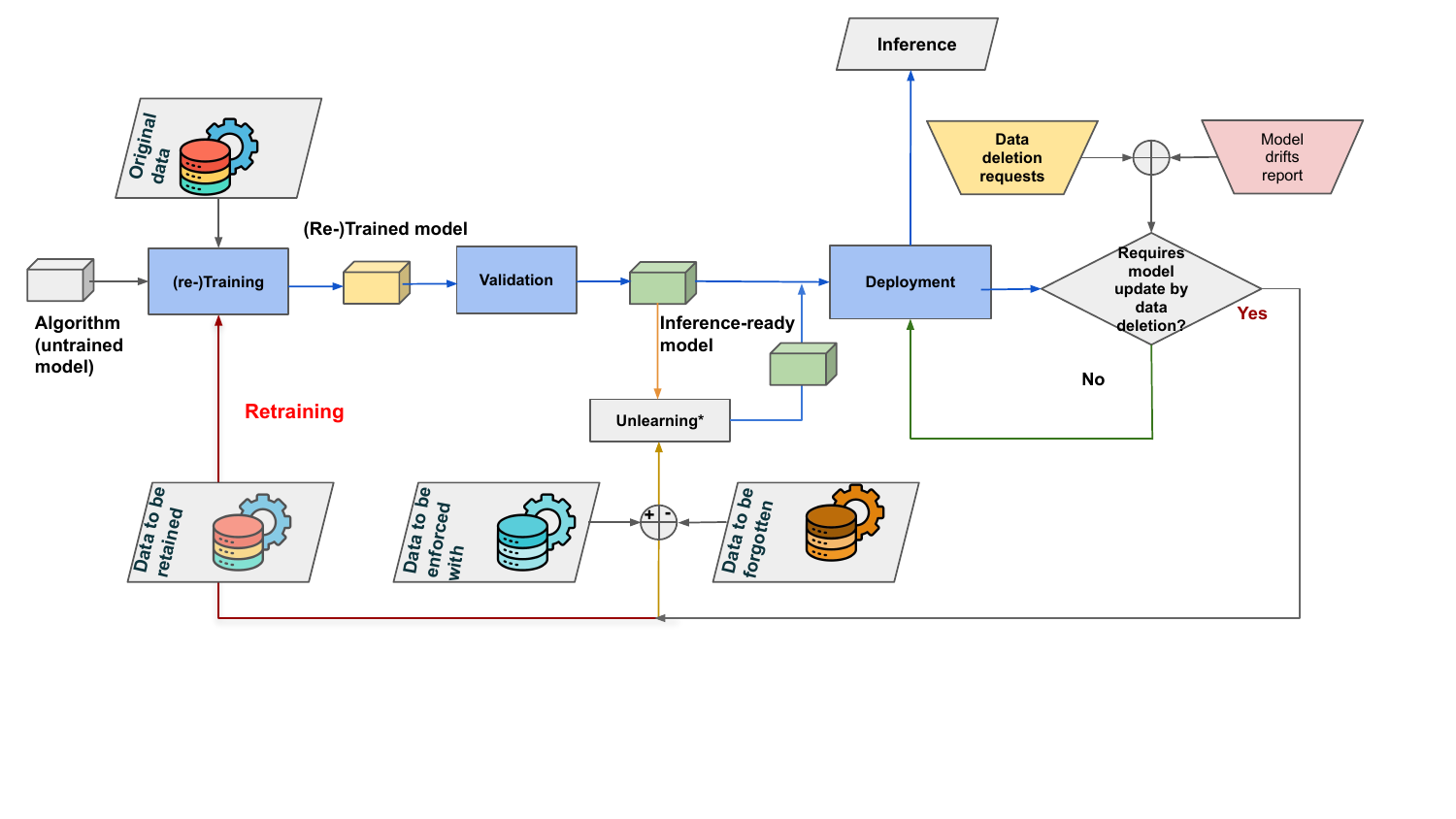}
    \caption{Preventing model deterioration in unlearning through `enforce' samples ($\mathcal{D}^e$).}
    \label{fig:enforcing}
\end{figure}

Eqn. \eqref{eqn:wJascent} and \eqref{eqn:wKascent} suggest that the relative proximity of the samples in $\mathcal{D}^f$, to the prototypes and class boundaries can could determine how their removal might impact the stability of the unlearned model, especially when $\mathcal{D}^f$ has class homogeneity. We investigate in this paper four such types of samples in $\mathcal{D}^f$, (individually, thus homogeneous class): (I) outlier; (II) located near the prototype of the original model; (III) located near the boundary of the class that is not the decision boundary; and (IV) located along/ near the decision boundary. When the size of $\mathcal{D}^f$ is large enough to contain samples from all classes and the mentioned categories, their effects are balanced out. However this is not so, when the size of $\mathcal{D}^f$ is very small ($<6$). To address these, we propose a reconfigured unlearning framework that is robust against the \emph{undesired, destabilizing} effects of smaller sample sizes of $\mathcal{D}^f$, during the unlearning-based update steps. We addressed this by introducing a set of `enforcer samples' from $\mathcal{D}$ \textbackslash $\mathcal{D}^f$ with similar class-distribution and sample size of $\mathcal{D}^f$, that will `protect' the model parameters from rapid deterioration by countering the mentioned de-stabilizing effect, as shown in \autoref{fig:enforcing}. This framework provides a trade-off between pure unlearning and retraining, as it uses a limited amount of data from $\mathcal{D}^r$ for enforcing, unlike retraining which uses the whole of $\mathcal{D}^r$. Thus it can be used for updating the model not just to `forget' specific samples, but also for compute-efficient and fast learning (compared to pure retraining) of a few new samples. Comparison of \emph{pure} Unlearning and the robust unlearning frameworks, on samples (I-IV) from the Breast Cancer dataset, can therefore show by the relative positions of the prototypes under these frameworks, that unlearning via the latter is more prominent. 
{\abovedisplayskip=1pt
\belowdisplayskip=2pt
\begin{align}
    \Delta \vec{w}_{f,e}^{J}=& \Bigg[
    \underbrace{\Bigg(-\alpha \cdot\frac{2 d_K(\vec{\mathcal{X}}_i^{\mathcal{D}^e})}{d_J(\vec{\mathcal{X}}_i^{\mathcal{D}^e})+d_K(\vec{\mathcal{X}}_i^{\mathcal{D}^e})} \cdot (-2(\vec{\mathcal{X}}_i^{\mathcal{D}^e}-\vec{w}_0^J))\Bigg)}_{\text{enforcing} \uparrow\uparrow \text{training}} +\underbrace{\Bigg(\alpha \cdot\frac{2 d_K(\vec{\mathcal{X}}_i^{\mathcal{D}^f})}{d_J(\vec{\mathcal{X}}_i^{\mathcal{D}^f})+d_K(\vec{\mathcal{X}}_i^{\mathcal{D}^f})} \cdot (-2(\vec{\mathcal{X}}_i^{\mathcal{D}^f}-\vec{w}_0^J))\Bigg)}_{\text{unlearning} \uparrow\downarrow \text{training}}
    \Bigg] \enspace, \label{eqn:wJrobustascent} \\
 \Delta \vec{w}_{f,e}^K =& \Bigg[ \underbrace{-\alpha \cdot \Bigg(\frac{-2 d_J(\vec{\mathcal{X}}_i^{\mathcal{D}^f})}{d_J(\vec{\mathcal{X}}_i^{\mathcal{D}^f})+d_K(\vec{\mathcal{X}}_i^{\mathcal{D}^f})} \cdot (-2(\vec{\mathcal{X}}_i^{\mathcal{D}^f}-\vec{w}_0^K))\Bigg)}_{\text{enforcing} \uparrow\uparrow \text{training}} +
 \underbrace{\Bigg(\alpha \cdot \frac{-2 d_J(\vec{\mathcal{X}}_i^{\mathcal{D}^f})}{d_J(\vec{\mathcal{X}}_i^{\mathcal{D}^f})+d_K(\vec{\mathcal{X}}_i^{\mathcal{D}^f})} \cdot (-2(\vec{\mathcal{X}}_i^{\mathcal{D}^f}-\vec{w}_0^K))\Bigg)}_{\text{unlearning} \uparrow\downarrow \text{training}}
   \Bigg] 
 \label{eqn:wKrobustascent}
\end{align}
}
This subset of data is hereby referred to as the \emph{enforcer set} given by $\mathcal{D}^e$). For this latter case, the $\mathcal{D}^f$ and $\mathcal{D}^e$ have their roles interchanged as the main and the `protector' process. The direction of the updates in `enforcing' is parallel to that in model training, i.e., opposite in direction to the updates in unlearning, as defined in Eqn. \eqref{eqn:wJrobustascent} and \eqref{eqn:wKrobustascent}. 

\subsection{Evaluation metrics}\label{ssec: eval}
Following \cite{zhang2023reviewMU, bourtoule2021MU, brophy2021UnlearnRF, thudi2022unrollingSGD}, the evaluation metrics relevant for GLVQ model within the scope of this paper are the $L^2$-norm between the parameters of the retrained and unlearned models, the gain in speed or computational efficiency when using unlearning versus retraining, percentage of predictions of the original model that are retained and that are lost, by the adapted models (retrained vs unlearned). 

\paragraph{Computational efficiency.} This is computed as Efficiency$=\frac{\text{Time taken to retrain the model on} \mathcal{D}^r}{\text{Time taken to unlearn }\mathcal{D}^f \text{from original model}}$ \cite{wang2024comprehensiveMU, zhang2023reviewMU, zhang2023reviewMU, brophy2021UnlearnRF}. A higher factor indicates a higher efficiency. 

\paragraph{$L^2$-norm.} Based on the definition of $L^2-$norm in \cite{wang2024comprehensiveMU, zhang2023reviewMU}, for GLVQ this term is computed based on the prototypes in the original ($\mathcal{A}(\mathcal{D})$), retrained, and the unlearned models. As model that unlearned the impacts of $\mathbf{D}^f$ should be as similar to its corresponding retrained model, a low $L^2$-norm based deviation between these models indicates successful unlearning. \cite{nguyen2025surveyMU} defines this as \emph{Layer-wise distance}.

\paragraph{Performance metrics.} 

When evaluating the performance of the unlearned models against the retrained models, in terms of their predictive ability, we consider the \% Retained, \%Lost, \%Improved upon, based on the \%Accuracy/ balanced accuracy. A successfully unlearned model should have high \% Retained and low \%Lost for the `remember' ($\mathcal{D}^r$) and the `Test' subsets of data, and the trends reversed on the `forget' subset ($\mathcal{D}^f$). In case of high \% Retained even for $\mathcal{D}^f$, it need not be considered an unsuccessful unlearning, if the performance of the unlearned model is very close to that of its corresponding retrained model (which never saw the samples in $\mathcal{D}^f$). In this case, it can imply that there are enough samples in $\mathcal{D}^r$ that are similar to those in $\mathcal{D}^f$, which masked the effect of unlearning. Mathematically, the \% Retained, Lost and Improved upon are given by:
\begin{itemize}
    \item $\% Retained$=$\frac{\#\text{samples that have been correctly classified by both original ($\mathcal{M}_0$) and the \emph{new} model ($\mathcal{M}_1$ or $\mathcal{M}_2)$}}{\#\text{samples that have been correctly classified by the original model}}$\cite{zhang2023reviewMU, wang2024comprehensiveMU}. 
    \item $\% Lost=\frac{\#\text{samples that have been correctly classified by $\mathcal{M}_0$ but misclassified by new model($\mathcal{M}_1$ or $\mathcal{M}_2)$}}{\#\text{samples that have been correctly classified by the original model}}$ \cite{zhang2023reviewMU, wang2024comprehensiveMU}. 
    \item $\% Improved=\frac{\#\text{samples that have been misclassified by $\mathcal{M}_0$ but correctly classified by new model($\mathcal{M}_1$ or $\mathcal{M}_2)$}}{\#\text{samples that have been correctly classified by the original model}}$ 
\end{itemize}

We use the metrics defined above to evaluate the performance of the retrained and unlearned models on the different sets, i.e., \emph{remembered}, test, unlearned subsets. ($\mathcal{A}(\mathcal{D}^r)$ or the unlearned model ($\mathcal{U}(\mathcal{A}(\mathcal{D}), \mathcal{D}^f)$, referred to \autoref{sec:Res} onward as $\mathcal{M}_1$ and $\mathcal{M}_2$ respectively). These evaluation metrics are based on \emph{Accuracy} and \emph{Completeness} defined in \cite{zhang2023reviewMU, wang2024comprehensiveMU}. However, since most of the datasets have significant class imbalance, we use Balanced accuracy instead of Accuracy for evaluation.

\section{Experiment design }\label{sec:Exp}
In this section first we mention the common experimental factors for all three research questions and lines of investigation mentioned in \autoref{ssec:contribution}, including benchmark datasets and evaluation metrics we used and why. Thereafter we explain the experimental setup for each of the research questions from \autoref{ssec:contribution}. 

\subsection{Datasets}
In this paper we used the following tabular datasets from the UCI repository and Kaggle, following \cite{wang2024comprehensiveMU, brophy2021UnlearnRF,nguyen2025surveyMU}: (i) Breast cancer ($\approx 570$ instances), (ii) Adult ($48K+$ instances), (iii) Diabetes (101766 instances), (iv) Surgical ($>14600$ instances), (v) Banking ($>41K$ instances) and (vi) Criteo ($\approx$ 1M instances) datasets. We specifically followed \cite{brophy2021UnlearnRF} for the exact source of the last five datasets mentioned here. 

\paragraph{Train- test split.} For datasets with prior split of training and set sets we used those as such in our experiments. For datasets without such split, we followed the recommendation in \cite{brophy2021UnlearnRF} and created a 80-20 (stratified) split for training and test sets respectively. The datasets whose target variables were originally multi-class have also been changed to binary class following the same paper. 

\paragraph{Selecting samples to be removed.} Samples in $\mathcal{D}^f$ were selected randomly. However, as the size and characteristics of $\mathcal{D}^f$ can impact the unlearned model, we vary size of $\mathcal{D}^f$ to include as few as 2 samples to a maximum of 2\% of the size of $\mathcal{D}$. We repeat this over 5 iterations\footnote{except in `Criteo' dataset, where we performed 3 iterations only}, to cover all possible characteristics of $\mathcal{D}^f$ and their respective impact on the update steps in unlearning. By $\mathcal{D}^f$ characteristics, we refer to whether its constituent $(\vec{\mathcal{X}},Y)$ are located near the class centers, decision boundaries, or are outliers. The deterioration due to these samples have been explained in \autoref{ss:robustUnlearning}. For the robust unlearning framework introduced in \autoref{ss:robustUnlearning}, $\mathcal{D}^e$ should have the same characteristics as $\mathcal{D}^f$, in terms of the subset size and class distribution of the constituent samples.

\subsection{Preprocessing}\label{ssec:prepro}
Although we tried to follow the preprocessing steps mentioned in \cite{brophy2021UnlearnRF} as closely as possible we decided not to follow their one-hot encoding of all categorical variables. Instead, wherever available, we preserved the ordinal category ranks, such as for the variable $Education$ from the `Adult' dataset, or converted to an approximate ranking as described in the IPython notebook called `dataset' in our toolbox. For categorical variables with nominal categories we did perform one-hot encoding, but preserved only those one-hot-encoded variables which had most variability. Although this is not the cleanest way for feature selection and could potentially introduce some bias, as we are not interested in the `absolute' predictive performance of GLVQ but the effectiveness of unlearning from a trained GLVQ, as a proof of concept, hence we proceeded with this such `hand-picked' feature engineering and selection. As Random Forest can implicitly perform feature importance, and since GLVQ unlike its extensions in \cite{Hammer20021059, schneider2009, ghosh2020visualisation, ghosh2025interpretable, mohammadi2024generalized} lack this ability, such a `hand-picked' removal of sparse features originating from one-hot-encoding was necessary to ensure that $\mathcal{M}_0$ learns enough for us to evaluate whether a poor performance of a $\mathcal{M}_2$ (i.e., $\mathcal{U}(\mathcal{D}^f,\mathcal{A}(\mathcal{D}))$) was due to removal of the 
$\mathcal{D}^f$ or simply due to the original model 
under-fitting. 

\subsection{Hyperparameters in GLVQ}\label{ssec:hyperparam}
The hyperparameter options within GLVQ are choice of the distance metrics, how the prototypes are initialized, the type of solvers, the type of activation function, and number of prototypes to represent each class. All experiments are based on the distance metric \emph{Squared Euclidean}, whereas for solvers we investigated \emph{Steepest gradient descent}, keeping step size (learning rate)
0.05 throughout all experiments, except for the dataset `Criteo', for which $\alpha=0.075$ to counter for the delay due to number of training samples. As we use the implementation of \cite{NaNLVQ}, the prototypes have been initialized near the class-means. Since we used log-transformation for pre-processing, we selected `Swish' as the activation function. This is because, based on its definition in \eqref{eqn:swish} we assume that it can handle data in log-scale more robustly. 

To confirm that this unlearning scheme is valid for GLVQ models with more than one prototype per class ($n_{prots/class}$ or ppc), we experimented with $ppc=[1,2]$ across all datasets and all research questions stated in \autoref{ssec:contribution}. The $\beta$ for the swish activation function for each dataset needed to be tuned separately, the specifics of which are in \autoref{sss:hyperparam}. 

\section{Results}\label{sec:Res}
In this section we first provide a brief overview of the characteristics of the publicly available datasets we used. Next we illustrate the efficiency of unlearning of a trained GLVQ model over its retraining, across all these datasets, and across different sizes of the 'forget' set $\mathcal{D}^f$, for the evaluation metrics described in \autoref{ssec: eval}, namely, speed-up factor, deviation of parameters of unlearned and retrained models. Thereafter we show the efficiency of unlearning in terms of $\%$ of correct predictions of original model ($\mathcal{M}_0$) retained by retrained model and unlearned model $\mathcal{U}(\mathcal{D}^f, \mathcal{A}(\mathcal{D}))$, which hereon will be referred to as $M2$. 

\autoref{tab:benchmark} describes the characteristics of the benchmark datasets on which we applied and evaluated the efficiency of our proposed unlearning strategy for GLVQ. $\#_{features}$ denotes 
the total number of features we considered following the preprocessing explained in \autoref{ssec:prepro} (dimensionality of these datasets). The amount of samples unlearned from each dataset varied based on the size of their respective training set size. 
\normalsize
\begin{table}[H]
\centering
\caption{Characteristics of the benchmark datasets in this paper}\label{ssec:datasets}
\label{tab:benchmark} 
\begin{tabularx}{\linewidth}{l l l l c r} 
\toprule
 \textbf{Dataset}& $\mathbf{N_{training}}$ & $\mathbf{N_{test}}$& \textbf{Class imbalance}& $\#_{features}$ & \textbf{samples in }$\mathbf{\mathcal{D}^f}$\\
  \toprule
    Breast cancer   &	381	    & 188	  & 145/236  ($\approx 0.614$)       & 30 & [2, 5, 20, 30, 50, 100]\\
    Surgical	   &  11708	    & 2927	  & 3000/8708 ($\approx 0.34$)      & 18 & [2, 12, 118, 586, 1171, 2342]\\
    Adult	       &  32561	    & 16281	  & 7841/24720 ($\approx 0.31$)     & 10 & [4, 33, 326, 1629, 3257, 6513]\\
    Banking	       &  32950	    & 8238	  & 3705/29245 ($\approx 0.13$)     & 18 & [4, 33, 330, 1648, 3295, 6590]\\
    Diabetes	   &  81412	    & 20354	  & 37500/43912  ($\approx 0.85$)   & 38 & [9, 82, 815, 4071, 8142, 16283]\\
    Criteo	       &  11463263  & 2516328 & 538681/10924582 ($\approx 0.05$)& 13 & [1147, 114633, 1146327]\\
    \bottomrule
\end{tabularx}
\end{table}
The models with ppc=1 and ppc=2 are denoted by the green and blue lines respectively in \autoref{fig:speedup} and \autoref{sfig:enforced-speedup}. In \autoref{fig:l2norm} and \autoref{sfig:enforced-l2norm} the green and yellow lines denote $\mathcal{M}_1=\mathcal{A}(\mathcal{D}^r)$ and $\mathcal{M}_2=\mathcal{U}(\mathcal{D}^f, \mathcal{A}(\mathcal{D}))$ each with ppc=1, respectively, and the blue and purple lines for their corresponding variants with ppc=2. If only a single color line is visible, it means that the number of prototypes do not have any significant impact in terms of the evaluation metric in focus. For ease of visualization in such almost overlapping plots, the marker shapes help in identifying the models in terms of the ppc: diamond for ppc=1, circle for ppc=2. As sample sizes of $\mathcal{D}^f$ increase by a factor of approximately 10, the x-axis for all datasets has been shown in log-scale in all figures, except mentioned otherwise. 
\subsection{Efficiency and $L^2-$distance}
\begin{figure}
    \centering
    \begin{subfigure}[!h]{0.49\textwidth}
    \centering
       \includegraphics[width=\textwidth]{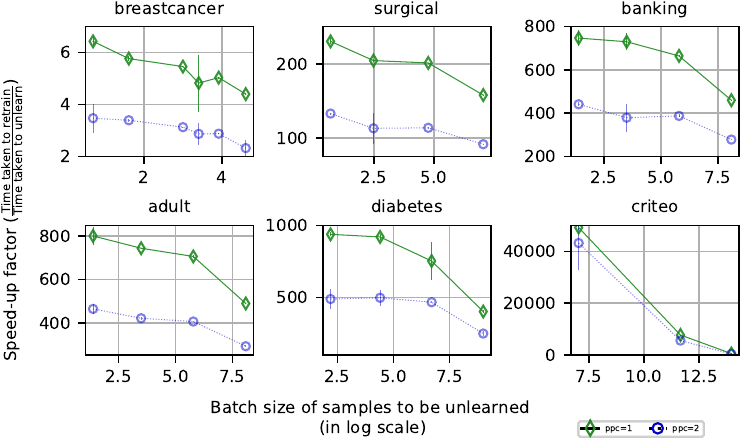}
    \caption{Speed-up factor of \emph{pure} unlearning of GLVQ models.}
         \label{fig:speedup}
    \end{subfigure}\hfill
     \begin{subfigure}[!h]{0.49\textwidth}
     \centering
    \centering
    \includegraphics[width=\textwidth]{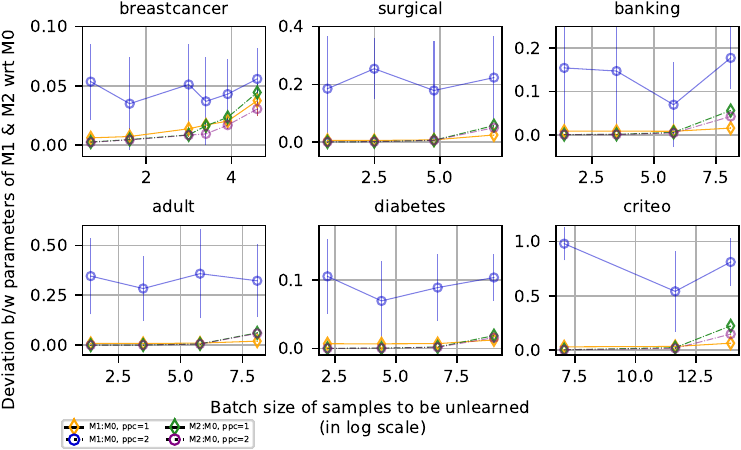}
    \caption{Deviation between prototypes of $\mathcal{M}_1$ and $\mathcal{M}_2$.}
    \label{fig:l2norm}
    \end{subfigure}
    \caption{Comparison of the unlearned and retrained models in terms of speed-up and relative deviation from original model.}
    \label{fig:unlearn-speed-dev}
\end{figure}

In this subsection we compare the unlearned models (both \emph{pure unlearning} and robust frameworks) and their corresponding retrained models in terms of their respective training time \emph{Efficiency} in terms of Speed-up, and deviation between model parameters (\autoref{fig:unlearn-speed-dev} and \autoref{fig:enforced-speed-dev}). 
\subsubsection{Gain in speed: Efficiency}\label{ssec:speedup}
\begin{figure}
    \centering
    \begin{subfigure}[!h]{0.49\textwidth}
    \centering
        \includegraphics[width=\textwidth]{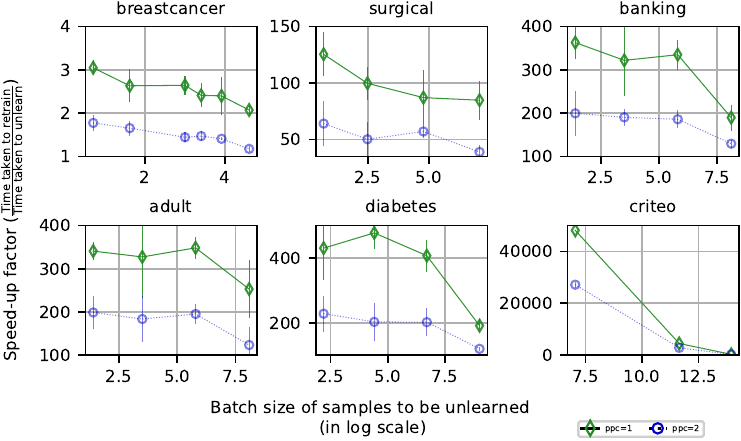}
        \caption{Speed-up factor of \emph{robust} unlearning of GLVQ models.}
        \label{sfig:enforced-speedup}
    \end{subfigure}\hfill
     \begin{subfigure}[!h]{0.49\textwidth}
     \centering
        \includegraphics[width=\textwidth]{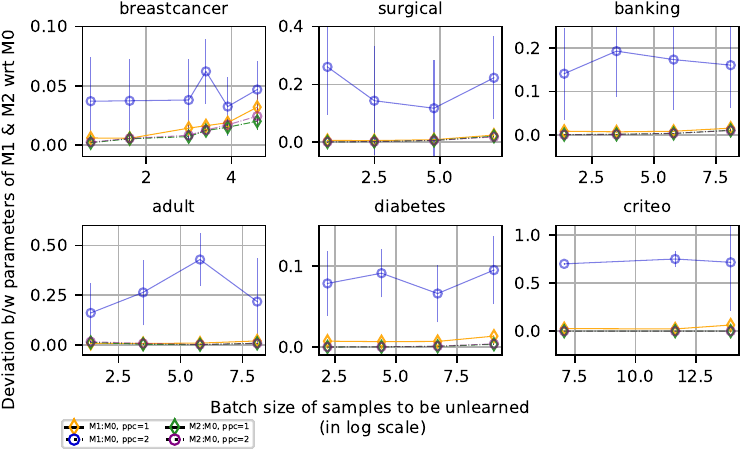}
        \caption{Deviation between prototypes of $\mathcal{M}_1$ and $\mathcal{M}_2$.}
        \label{sfig:enforced-l2norm}
    \end{subfigure}
    \caption{Comparison of the retrained model and unlearning-with enforcer in terms of speed-up and relative deviation from original model.}
    \label{fig:enforced-speed-dev}
\end{figure}

 The trends between the unlearned and retrained GLVQ models with ppc=1 and ppc=2 are almost collinear for both the mentioned frameworks, although the gain in speed is higher for models with ppc=1, as expected. This difference is however not visibly different for the largest dataset (`Criteo'). These also indicate that even though unlearning is faster than retraining, the speed-up is more prominent (i.e., unlearning is more efficient than retraining) for the larger datasets, and for lower batch size of $\mathcal{D}^f$. 
 Comparing \autoref{fig:speedup} against \autoref{sfig:enforced-speedup} we further notice that the speed-up of models with ppc=1 versus that with ppc=2 for the largest dataset is more significant for the \emph{robust} unlearning framework. This is because \emph{robust} unlearning models with ppc=2 involves the computation of gradients of 1147 `enforcer' samples (see `Criteo' in \autoref{tab:benchmark}) with respect to 4 prototypes instead of just 2. 

\subsubsection{Deviation between parameters of retrained and unlearned models}\label{ssec:Dev}
\autoref{fig:l2norm} and \autoref{sfig:enforced-l2norm} illustrate the $L^2$-norm deviation between the prototypes of $\mathcal{M}_1$ and $\mathcal{M}_2$ models with regards to the parameters of the original model $\mathcal{M}_0$, across the varying sizes of the $\mathcal{D}^f$. We see that the deviation of the unlearned models (green and purple dotted lines) are maximum for the larger sizes of $\mathcal{D}^f$, when applying the \emph{pure} unlearning strategy (\autoref{fig:l2norm}), which is addressed by the robust unlearning strategy(\autoref{sfig:enforced-l2norm}). The performance of this stabilized unlearning technique, based on the metrics described in \autoref{ssec: eval}, for the Breast cancer dataset are in \autoref{sfig:enf-breatscancer}, and for the other datasets in \autoref{fig:enf-performance}. For all datasets the deviation of the unlearned model's parameters from the original model lie between that of the retrained model's for ppc=1 and ppc=2. 

\begin{figure}
    \centering
    \begin{subfigure}[!h]{0.59\textwidth}
    \centering
    \includegraphics[trim={0.2cm 0.01cm 0.2cm 0.35cm}, clip,width=\textwidth]{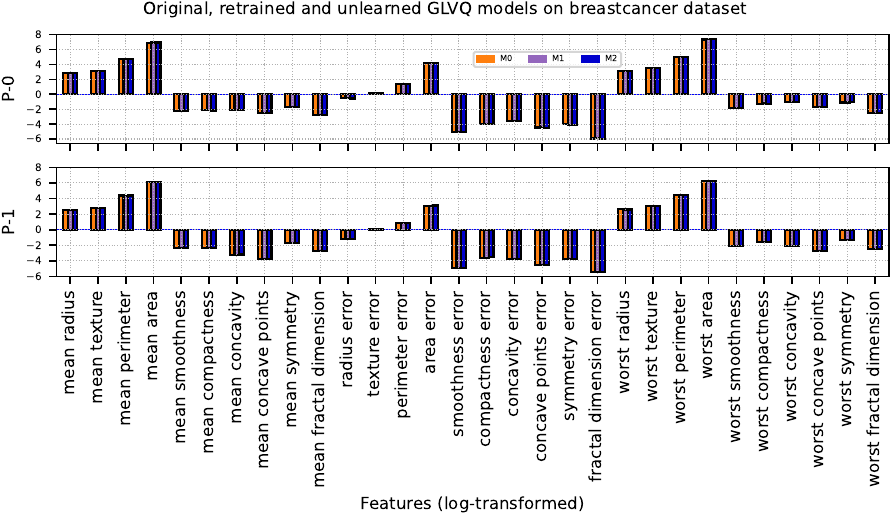}
    \caption{Deviation between the prototypes of the originally trained GLVQ model M0, its retrained $\mathcal{M}_1$ and unlearned $\mathcal{M}_2$ variants on the `Breast cancer' dataset.}
    \label{sfig:bc-prot-dev-bars}
    \end{subfigure}
    \hfill
    \begin{subfigure}[!h]{0.4\textwidth}
    \centering 
    \includegraphics[width=\textwidth]{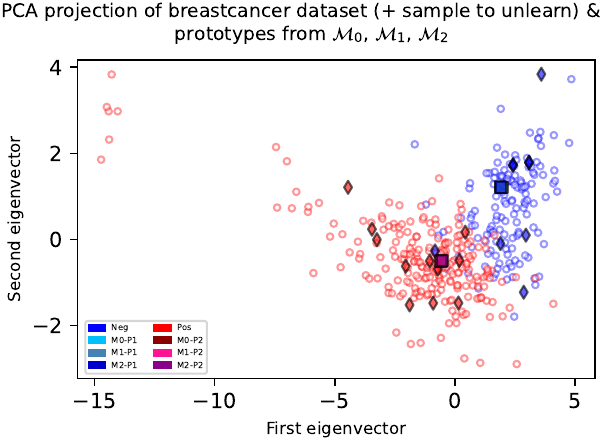}
    \caption{Distribution of samples from $\mathcal{D}^r$, $\mathcal{D}^f$, prototypes of the original, retrained and \emph{pure} unlearned models, in PCA-transformed space. }
    \label{fig:distr}
    \end{subfigure}
    \caption{Impact on unlearning vs retraining on their respective prototypes, and their distribution with respect to the `remembered' and `forget' sets of data.}
    \label{fig:bc-20-samples-distr-prots}
\end{figure}

While \autoref{fig:l2norm} and \autoref{sfig:enforced-l2norm} portray the overall deviation between the retrained and unlearned models relative to the original model, \autoref{fig:bc-20-samples-distr-prots} zooms into one of the model-triads from the `Breast cancer' dataset with sample size of $\mathcal{D}^f=20$ (9 samples from `Neg' and 11 from `Pos' classes), for GLVQ model trained originally with $ppc=1$. \autoref{fig:bc-20-samples-distr-prots} shows that the prototype sets from all the three models are close to each other for this iteration of twenty selected samples to be forgotten, not terms of the values of the prototypes (\autoref{sfig:bc-prot-dev-bars}) and the distribution of the prototypes and the samples from the `remember' and `forget' sets of data. \autoref{sfig:bc-prot-dev-bars} shows that for the feature `mean area', the value of prototype P-0 is 7, while for prototype P-1 it is 6 (thus a difference of $\approx$ 690sq mm in the input domain scale). For the feature `mean concave points', P-0 is -2.5 and P-1 -4.
Complementary to \autoref{sfig:bc-prot-dev-bars}, \autoref{fig:distr} provides an overview of the proximity of the samples in $\mathcal{D}^r$ and $\mathcal{D}^f$ relative to each of the prototypes of the original, retrained and unlearned models, and the impact of $\mathcal{D}^f$ on the prototype(s) in the unlearned model. The points with reddish shades indicate the class `positive diagnosis' (1), whereas the blueish shades indicate `negative' (0); the square markers indicate the prototypes.  
\subsection{Impact of location of samples in $\mathcal{D}^f$, on unlearning}\label{asec:sample location}
This section presents a systematic analysis of the impact of the location of a sample in $\mathcal{D}^f$, on the update of the prototypes from the original trained GLVQ model ($\mathcal{M}_0$) to those in the unlearned model ($\mathcal{M}_2$). Specifically, the findings from this analyses support the motivation for the robust unlearning framework introduced in \autoref{ss:robustUnlearning}. 
As the `Breast cancer' dataset is the smallest in size and with a `clear' decision boundary between the two classes, visible even in 2-dimensions (along first two Principal Components), we performed the analysis of impact of the $\mathcal{D}^f$ characteristics on the unlearned model $\mathcal{M}_2$, on this dataset. 
\begin{figure}
    \centering
    \begin{subfigure}[!h]{0.49\textwidth}
      \includegraphics[width=0.99\textwidth]{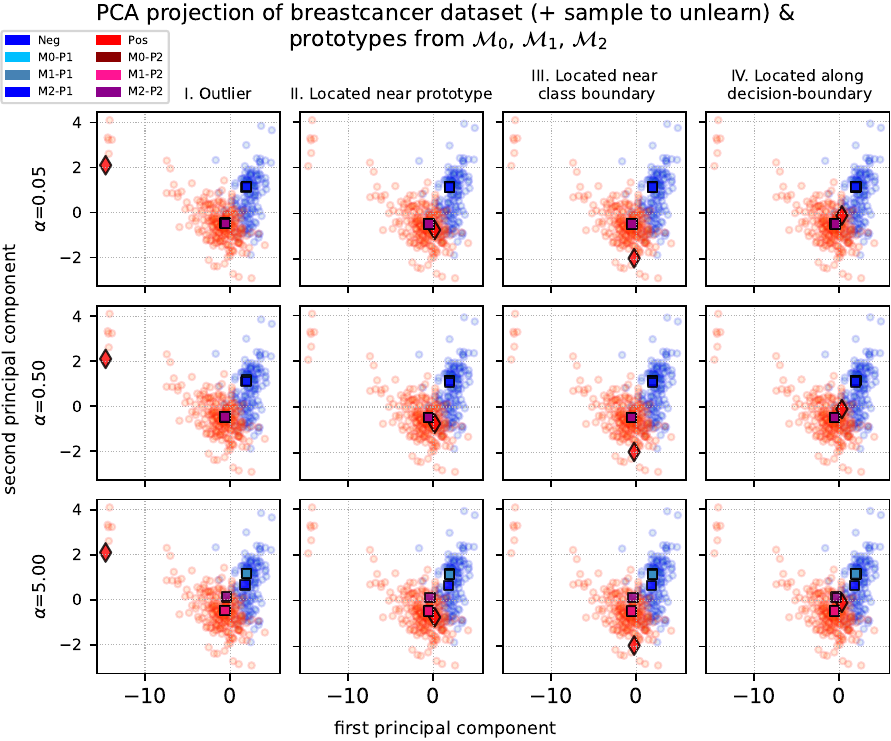}
    \caption{Based on the \emph{pure} unlearning strategy. }   
    \label{fig:sample_location_impact}
    \end{subfigure}\hfill
    \begin{subfigure}[!h]{0.49\textwidth}
      \includegraphics[width=0.99\textwidth]{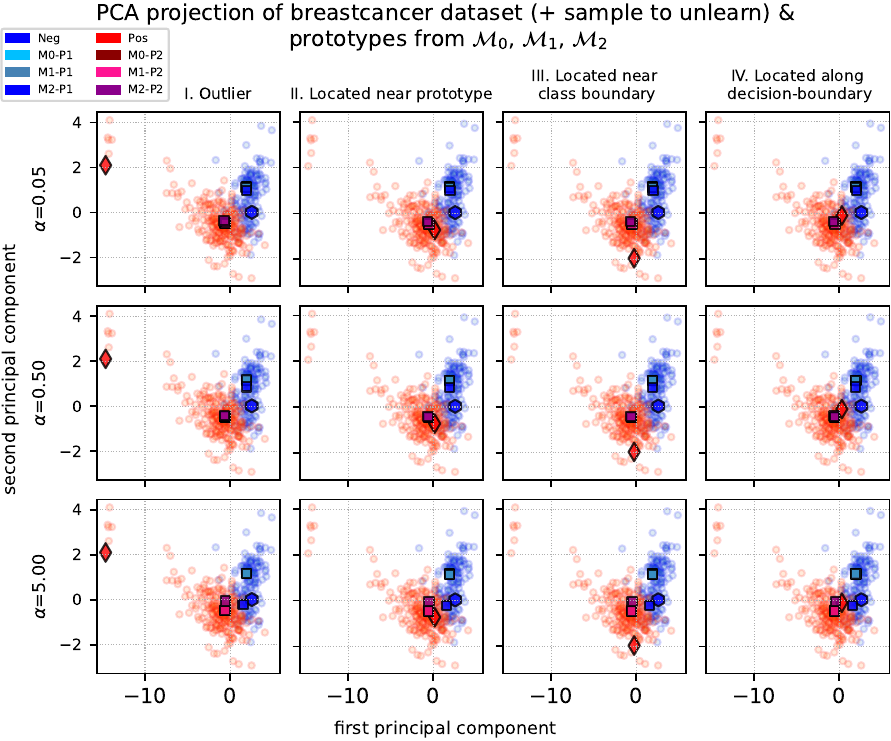}
      \caption{Based on the robust unlearning strategy. }
      \label{fig:sampleImpact_robust}
    \end{subfigure}
    \caption{Impact of location of samples in $\mathcal{D}^f$ on unlearning( $\mathcal{M}_0\rightarrow \mathcal{M}_2$), without and with `enforcer' set (the hexagon represents the `enforcer' sample.).}
    \label{fig:sample_location}
\end{figure}
 For this analysis we focus on the impact of removing the following types of samples from  $\mathcal{M}_0$, varying $\alpha$ by $10\times$ and $100 \times$ during only the unlearning phase, to amplify the \emph{push-and-pull} effects for better visualization: 
(I) An outlier sample; (II) A sample near the prototype of a class; 
(III) A sample near the boundary of a class that is not the decision boundary; and (IV) A sample on the decision boundary of the 2 classes. Each of these samples were unlearned individually, without the balancing effect of `enforcer' samples (\autoref{fig:sample_location_impact}) and with `enforcer' (\autoref{fig:sampleImpact_robust}), so that we could investigate the exact impact of each of these types of samples, and the role of `enforcer' samples in \emph{robust} unlearning. 

Each column in the sub-figures of \autoref{fig:sample_location} indicates the location-based category of sample, while each row correspond to the $\alpha$ used for that unlearning model. The effect of unlearning is represented by the movement of the prototype of the `Pos'-class from $\mathcal{M}_2$ away from that of the corresponding $\mathcal{M}_0$ and towards the decision-boundary (towards the opposite class), and the reverse for that of the prototype of the `Neg' class. Although unlearning $\mathcal{D}^f$ can already be noticed from the \emph{pure} unlearning framework (\autoref{fig:sampleImpact_robust}), the unlearning framework with `enforcer samples' makes the resulting $\mathcal{M}_2$ model robust by reducing the unbalanced \emph{pull and push effects} of the $\mathcal{D}^f$ on the prototypes of the incorrect and correct classes respectively, as seen in row-3 of \autoref{fig:sampleImpact_robust} (following the effects of unlearning and enforcing defined in Eqn. \eqref{eqn:wJrobustascent} and \eqref{eqn:wKrobustascent}). This is evident from comparing the second column of \autoref{fig:sample_location_impact} against that of \autoref{fig:sampleImpact_robust}, since in the latter, the enforcer sample (blue hexagon), balances out some of the undesired (and destabilizing) effects of \emph{pure} unlearning. This means that the `enforcer' sample from the blue class reduces the pushing force on the prototype of the red class imposed by the `forget' samples of the red class. Simultaneously this `enforcer' sample also reduces the pulling force on the blue class prototype towards the red class imposed by the same `forget' sample of the red class. This trend is also noticed for the third column. However for the fourth column, the unlearning with enforcers seem to be having the opposite effect. This could be due to the choice of the enforcer sample (also near the decision boundary), which was kept constant for all the investigations in this section. 

\subsection{Retained, lost and gained predictions}\label{ss:res_perf}
\begin{figure}
    \centering
    \begin{subfigure}[!h]{0.49\textwidth}
    \includegraphics[width=0.99\textwidth]{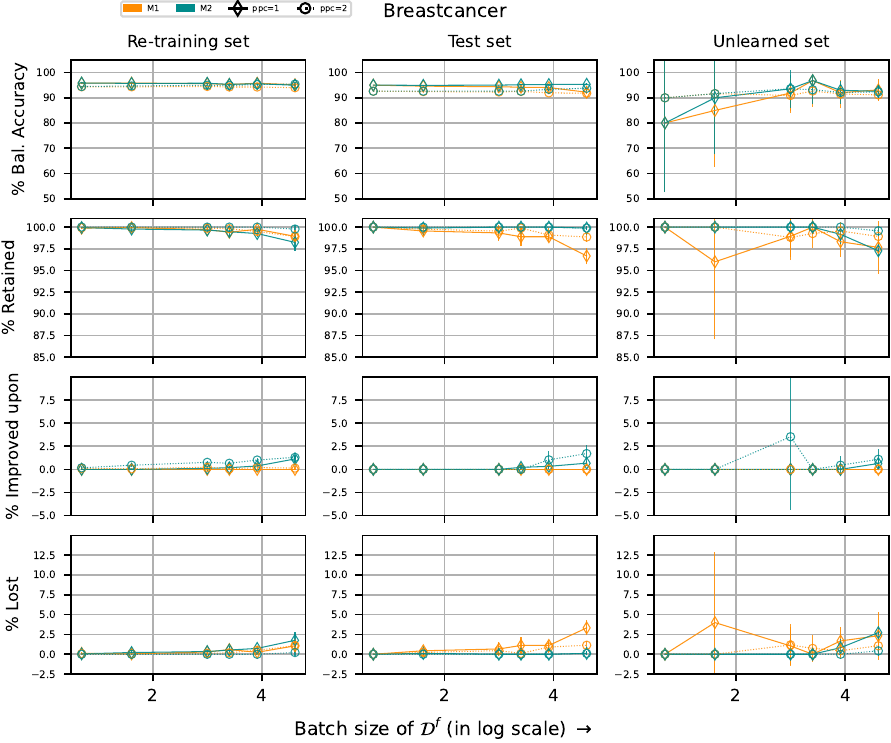}
    \caption{Based on the \emph{pure} unlearning framework, (corresponding \autoref{tab:_perf_breastcancer}).}
    \label{fig:perf-breastcancer}
    \end{subfigure}\hfill
    \begin{subfigure}[!h]{0.49\textwidth}
        \includegraphics[width=0.99\textwidth]{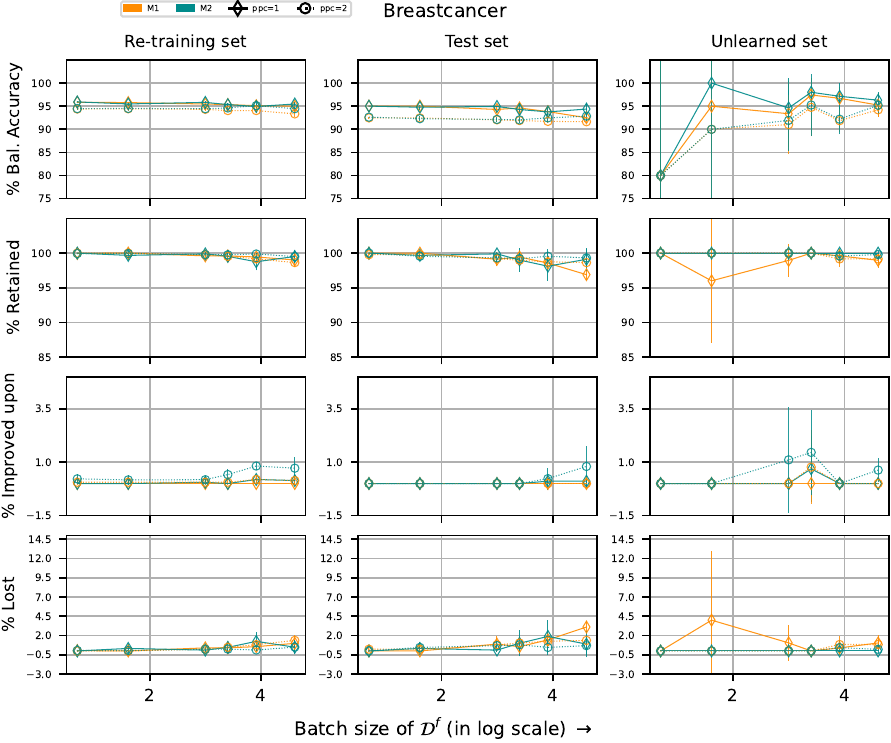}
        \caption{Based on \emph{robust} unlearning framework (corresponding \autoref{tab:enf_perf_bc}).}
        \label{sfig:enf-breatscancer}
    \end{subfigure}
\caption{Performance of the retrained and unlearned models on the samples from $\mathcal{D}^r$, test set, and $\mathcal{D}^f$ of the `Breast cancer' dataset, in terms of predictions retained, lost and gained with respect to the original trained model. }
\label{fig:perf-breastcancer-all}
\end{figure}

In this section we showcase in \autoref{fig:perf-breastcancer-all} and its corresponding \autoref{tab:_perf_breastcancer} and \autoref{tab:enf_perf_bc}, the performance of the \emph{pure} unlearning strategy (\autoref{fig:unlearning}) on the `Breast cancer' dataset, for GLVQ models with ppc=1, versus that from the robust unlearning strategy (\autoref{fig:enforcing}). The Appendix contains the performance plot for all other datasets we investigated in this paper. The performance across the varying sizes of $\mathcal{D}^f$ are based on the evaluation metrics defined in \autoref{ssec: eval}, each of which are plotted in each row of \autoref{fig:perf-breastcancer} and \autoref{sfig:enf-breatscancer}. 
Balanced accuracy appear to be mostly parallel on the `remembered' dataset $\mathcal{D}^r$ as well as the test subset. Comparing \autoref{sfig:enf-breatscancer} against \autoref{fig:perf-breastcancer} we find that the performance deviations corresponding to each size of $\mathcal{D}^f$ is smaller from the robust unlearning strategy, especially in terms of \% Retained (second row of both sub-plots). 
The decrease in performance deviation in the \emph{robust} unlearning framework is significant especially for larger datasets, such as the `Banking' dataset, where the \% lost across all three subsets of data are much lower, and \% retained is much higher for the \emph{robust} unlearning (\autoref{sfig:enf-banking}) compared to that in \emph{pure} unlearning ( \autoref{sfig:banking}), especially for higher size of $\mathcal{D}^f$. Similar trends are seen in the `Surgical dataset (\autoref{fig:perf_surgical} vs \autoref{sfig:enf-surgical}), `Adult dataset (\autoref{sfig:adult} vs \autoref{sfig:enf-adult}), `Diabetes' dataset (\autoref{sfig:diabetes} vs \autoref{sfig:enf-diabetes})), and `Criteo' (\autoref{sfig:criteo} vs \autoref{sfig:enf-criteo})).

\scriptsize{
    \centering
    \begin{longtable}{p{1cm}p{1.8cm}p{1.0cm}p{2cm}p{2cm}p{2cm}p{2cm}p{2cm}}
    \caption{Comparison performance of the retrained and unlearned models across different batch sizes of $\mathcal{D}^f$ (N($\mathcal{D}^f$)) in the `Breast cancer' dataset, reported as mean (std) over 5 iterations.}\label{tab:_perf_breastcancer}\\
    \toprule
ppc	&	N($\mathcal{D}^f$)	&	Subset & Model	& \% Bal. Acc	& \% Gained	&	\% Lost	& \% Retained	\\
\toprule%
1	&	2	&	retraining	&	M1	&	95.81 (0.16)	&	0.0 (0.0)	&	0.11 (0.15)	&	99.89 (0.15)	\\	
	&		&		&	M2	&	95.88 (0.21)	&	0.0 (0.0)	&	0.05 (0.12)	&	99.95 (0.12)	\\	\cmidrule{3-8}
	&		&	test	&	M1	&	95.03 (0.0)	&	0.0 (0.0)	&	0.0 (0.0)	&	100.0 (0.0)	\\	
	&		&		&	M2	&	95.03 (0.0)	&	0.0 (0.0)	&	0.0 (0.0)	&	100.0 (0.0)	\\	\cmidrule{3-8}
	&		&	unlearned	&	M1	&	80.0 (27.39)	&	0.0 (0.0)	&	0.0 (0.0)	&	100.0 (0.0)	\\	
	&		&		&	M2	&	80.0 (27.39)	&	0.0 (0.0)	&	0.0 (0.0)	&	100.0 (0.0)	\\	\cmidrule{2-8}
	&	5	&	retraining	&	M1	&	95.84 (0.16)	&	0.0 (0.0)	&	0.05 (0.12)	&	99.95 (0.12)	\\	
	&		&		&	M2	&	95.74 (0.47)	&	0.0 (0.0)	&	0.22 (0.3)	&	99.78 (0.3)	\\	\cmidrule{3-8}
	&		&	test	&	M1	&	94.63 (0.56)	&	0.0 (0.0)	&	0.44 (0.61)	&	99.56 (0.61)	\\	
	&		&		&	M2	&	94.95 (0.18)	&	0.0 (0.0)	&	0.11 (0.25)	&	99.89 (0.25)	\\	\cmidrule{3-8}
	&		&	unlearned	&	M1	&	85.0 (22.36)	&	0.0 (0.0)	&	4.0 (8.94)	&	96.0 (8.94)	\\	
	&		&		&	M2	&	90.0 (22.36)	&	0.0 (0.0)	&	0.0 (0.0)	&	100.0 (0.0)	\\	\cmidrule{2-8}
	&	20	&	retraining	&	M1	&	95.63 (0.35)	&	0.0 (0.0)	&	0.29 (0.29)	&	99.71 (0.29)	\\	
	&		&		&	M2	&	95.78 (0.27)	&	0.11 (0.26)	&	0.34 (0.13)	&	99.66 (0.13)	\\	\cmidrule{3-8}
	&		&	test	&	M1	&	94.4 (0.86)	&	0.0 (0.0)	&	0.67 (0.91)	&	99.33 (0.91)	\\	
	&		&		&	M2	&	95.03 (0.0)	&	0.0 (0.0)	&	0.0 (0.0)	&	100.0 (0.0)	\\	\cmidrule{3-8}
	&		&	unlearned	&	M1	&	91.91 (7.23)	&	0.0 (0.0)	&	1.05 (2.35)	&	98.95 (2.35)	\\	
	&		&		&	M2	&	93.58 (5.51)	&	0.0 (0.0)	&	0.0 (0.0)	&	100.0 (0.0)	\\	\cmidrule{2-8}
	&	30	&	retraining	&	M1	&	95.28 (0.44)	&	0.0 (0.0)	&	0.59 (0.47)	&	99.41 (0.47)	\\	
	&		&		&	M2	&	95.34 (0.61)	&	0.18 (0.16)	&	0.53 (0.56)	&	99.47 (0.56)	\\	\cmidrule{3-8}
	&		&	test	&	M1	&	94.2 (0.83)	&	0.0 (0.0)	&	1.11 (1.11)	&	98.89 (1.11)	\\	
	&		&		&	M2	&	95.19 (0.22)	&	0.22 (0.3)	&	0.0 (0.0)	&	100.0 (0.0)	\\	\cmidrule{3-8}
	&		&	unlearned	&	M1	&	96.85 (0.83)	&	0.0 (0.0)	&	0.0 (0.0)	&	100.0 (0.0)	\\	
	&		&		&	M2	&	96.85 (0.83)	&	0.0 (0.0)	&	0.0 (0.0)	&	100.0 (0.0)	\\	\cmidrule{2-8}
	&	50	&	retraining	&	M1	&	95.79 (0.27)	&	0.0 (0.0)	&	0.25 (0.26)	&	99.75 (0.26)	\\	
	&		&		&	M2	&	95.58 (0.34)	&	0.37 (0.14)	&	0.74 (0.35)	&	99.26 (0.35)	\\	\cmidrule{3-8}
	&		&	test	&	M1	&	94.07 (0.43)	&	0.0 (0.0)	&	1.11 (0.39)	&	98.89 (0.39)	\\	
	&		&		&	M2	&	95.28 (0.22)	&	0.33 (0.3)	&	0.0 (0.0)	&	100.0 (0.0)	\\	\cmidrule{3-8}
	&		&	unlearned	&	M1	&	92.03 (2.61)	&	0.0 (0.0)	&	1.68 (1.75)	&	98.32 (1.75)	\\	
	&		&		&	M2	&	92.98 (1.81)	&	0.0 (0.0)	&	0.83 (1.14)	&	99.17 (1.14)	\\	\cmidrule{2-8}
	&	100	&	retraining	&	M1	&	95.19 (0.78)	&	0.0 (0.0)	&	1.06 (0.26)	&	98.94 (0.26)	\\	
	&		&		&	M2	&	94.99 (1.39)	&	1.13 (0.39)	&	1.76 (0.99)	&	98.24 (0.99)	\\	\cmidrule{3-8}
	&		&	test	&	M1	&	92.22 (0.71)	&	0.0 (0.0)	&	3.33 (0.96)	&	96.67 (0.96)	\\	
	&		&		&	M2	&	95.38 (0.42)	&	0.67 (0.25)	&	0.11 (0.25)	&	99.89 (0.25)	\\	\cmidrule{3-8}
	&		&	unlearned	&	M1	&	93.14 (4.33)	&	0.0 (0.0)	&	2.34 (3.05)	&	97.66 (3.05)	\\	
	&		&		&	M2	&	92.59 (1.94)	&	0.64 (0.58)	&	2.73 (0.6)	&	97.27 (0.6)	\\	\hline
2	&	2	&	retraining	&	M1	&	94.39 (0.28)	&	0.11 (0.15)	&	0.06 (0.12)	&	99.94 (0.12)	\\	
	&		&		&	M2	&	94.47 (0.27)	&	0.17 (0.25)	&	0.0 (0.0)	&	100.0 (0.0)	\\	\cmidrule{3-8}
	&		&	test	&	M1	&	92.63 (0.0)	&	0.0 (0.0)	&	0.0 (0.0)	&	100.0 (0.0)	\\	
	&		&		&	M2	&	92.63 (0.0)	&	0.0 (0.0)	&	0.0 (0.0)	&	100.0 (0.0)	\\	\cmidrule{3-8}
	&		&	unlearned	&	M1	&	90.0 (22.36)	&	0.0 (0.0)	&	0.0 (0.0)	&	100.0 (0.0)	\\	
	&		&		&	M2	&	90.0 (22.36)	&	0.0 (0.0)	&	0.0 (0.0)	&	100.0 (0.0)	\\	\cmidrule{2-8}
	&	5	&	retraining	&	M1	&	94.36 (0.24)	&	0.06 (0.13)	&	0.06 (0.13)	&	99.94 (0.13)	\\	
	&		&		&	M2	&	94.7 (0.2)	&	0.45 (0.15)	&	0.0 (0.0)	&	100.0 (0.0)	\\	\cmidrule{3-8}
	&		&	test	&	M1	&	92.47 (0.22)	&	0.0 (0.0)	&	0.23 (0.31)	&	99.77 (0.31)	\\	
	&		&		&	M2	&	92.63 (0.0)	&	0.0 (0.0)	&	0.0 (0.0)	&	100.0 (0.0)	\\	\cmidrule{3-8}
	&		&	unlearned	&	M1	&	91.67 (11.79)	&	0.0 (0.0)	&	0.0 (0.0)	&	100.0 (0.0)	\\	
	&		&		&	M2	&	91.67 (11.79)	&	0.0 (0.0)	&	0.0 (0.0)	&	100.0 (0.0)	\\	\cmidrule{2-8}
	&	20	&	retraining	&	M1	&	94.51 (0.33)	&	0.23 (0.13)	&	0.12 (0.16)	&	99.88 (0.16)	\\	
	&		&		&	M2	&	94.98 (0.43)	&	0.76 (0.16)	&	0.0 (0.0)	&	100.0 (0.0)	\\	\cmidrule{3-8}
	&		&	test	&	M1	&	92.3 (0.18)	&	0.0 (0.0)	&	0.46 (0.26)	&	99.54 (0.26)	\\	
	&		&		&	M2	&	92.63 (0.0)	&	0.0 (0.0)	&	0.0 (0.0)	&	100.0 (0.0)	\\	\cmidrule{3-8}
	&		&	unlearned	&	M1	&	90.93 (6.99)	&	0.0 (0.0)	&	1.18 (2.63)	&	98.82 (2.63)	\\	
	&		&		&	M2	&	93.6 (7.48)	&	3.53 (7.89)	&	0.0 (0.0)	&	100.0 (0.0)	\\	\cmidrule{2-8}
	&	30	&	retraining	&	M1	&	94.3 (0.45)	&	0.06 (0.13)	&	0.18 (0.16)	&	99.82 (0.16)	\\	
	&		&		&	M2	&	94.9 (0.49)	&	0.66 (0.13)	&	0.0 (0.0)	&	100.0 (0.0)	\\	\cmidrule{3-8}
	&		&	test	&	M1	&	92.55 (0.18)	&	0.0 (0.0)	&	0.11 (0.26)	&	99.89 (0.26)	\\	
	&		&		&	M2	&	92.63 (0.0)	&	0.0 (0.0)	&	0.0 (0.0)	&	100.0 (0.0)	\\	\cmidrule{3-8}
	&		&	unlearned	&	M1	&	92.69 (6.3)	&	0.0 (0.0)	&	0.74 (1.66)	&	99.26 (1.66)	\\	
	&		&		&	M2	&	93.21 (5.64)	&	0.0 (0.0)	&	0.0 (0.0)	&	100.0 (0.0)	\\	\cmidrule{2-8}
	&	50	&	retraining	&	M1	&	94.28 (0.25)	&	0.19 (0.17)	&	0.5 (0.48)	&	99.5 (0.48)	\\	
	&		&		&	M2	&	95.27 (0.39)	&	1.01 (0.27)	&	0.0 (0.0)	&	100.0 (0.0)	\\	\cmidrule{3-8}
	&		&	test	&	M1	&	91.97 (0.56)	&	0.0 (0.0)	&	0.91 (0.77)	&	99.09 (0.77)	\\	
	&		&		&	M2	&	93.37 (0.68)	&	1.03 (0.94)	&	0.0 (0.0)	&	100.0 (0.0)	\\	\cmidrule{3-8}
	&		&	unlearned	&	M1	&	91.52 (5.66)	&	0.0 (0.0)	&	0.45 (1.02)	&	99.55 (1.02)	\\	
	&		&		&	M2	&	92.13 (4.45)	&	0.45 (1.02)	&	0.0 (0.0)	&	100.0 (0.0)	\\	\cmidrule{2-8}
	&	100	&	retraining	&	M1	&	94.02 (0.87)	&	0.14 (0.2)	&	1.08 (0.43)	&	98.92 (0.43)	\\	
	&		&		&	M2	&	95.43 (0.33)	&	1.3 (0.2)	&	0.22 (0.49)	&	99.78 (0.49)	\\	\cmidrule{3-8}
	&		&	test	&	M1	&	91.8 (0.42)	&	0.0 (0.0)	&	1.14 (0.57)	&	98.86 (0.57)	\\	
	&		&		&	M2	&	93.79 (0.75)	&	1.71 (0.9)	&	0.11 (0.26)	&	99.89 (0.26)	\\	\cmidrule{3-8}
	&		&	unlearned	&	M1	&	91.16 (2.16)	&	0.0 (0.0)	&	1.07 (1.06)	&	98.93 (1.06)	\\	
	&		&		&	M2	&	92.37 (2.61)	&	1.09 (1.09)	&	0.43 (0.59)	&	99.57 (0.59)	\\	\bottomrule
 \end{longtable}
}
\normalsize

\section{Discussion}\label{sec:Disc}
As number of samples to be unlearned simultaneously increases, the gain in speed from unlearning relative to retraining decreases, as is expected. However for smaller batches of data unlearning is clearly superior in terms of speed (\autoref{fig:speedup} and \autoref{sfig:enforced-speedup}). \autoref{fig:l2norm} and \autoref{sfig:enforced-l2norm} highlight the efficacy of unlearning, in terms of the low deviation between the parameters of retrained and unlearned model across the range of sample size of $\mathcal{D}^f$ with the added advantage of low computational resources. Due to the intrinsic interpretability of the LVQ-family of classifiers we can also investigate the impact of the unlearning on the model parameters (here the prototypes), as seen in \autoref{fig:bc-20-samples-distr-prots} and from \autoref{fig:sample_location} (both illustrated on the `Breast cancer' dataset). Additionally, comparison of \autoref{fig:sample_location_impact} and \autoref{fig:sampleImpact_robust} showcases the effect of unbalanced repulsion and attraction forces on the prototypes of the unlearned model, that the robust unlearning framework is better equipped to address. 

It is the intrinsic interpretability of GLVQ models that enable gaining the trust of the clients or use-case providers. 
In unlearning scenario, it is the same property which can assure the concerned parties and stakeholders that the impact of the samples flagged for deletion have been removed. For example, when the performance of the unlearned model and the deviation of its prototypes from that of its corresponding original model appear to be near-zero, it might raise suspicion about whether the samples to be forgotten by the model were indeed forgotten. While addressing such questions form the core of the challenges in machine unlearning effectiveness verification \cite{wang2024comprehensiveMU}, access to the relative proximity of these samples to the prototypes of the original and the unlearned model, provides reliable explanations and evidence, thereby contributing to interpretable machine unlearning. In other words, the relative proximity of these `forget' set samples and those belonging to $\mathcal{D}^r$, from the 
prototypes of the original and those of the unlearned models, as seen in \autoref{fig:sample_location}, can indicate the underlying reason behind the seemingly `unaltered' $\mathcal{M}_2$ model; e.g., availability of enough samples in $\mathcal{D}^r$ that are very similar to those $\mathcal{D}^f$, or the sample in $\mathcal{D}^f$ being an outlier. 
This is explored in depth in \autoref{asec:sample location}, which inspired the development of the more robust version of the unlearning scheme presented in \autoref{fig:enforcing}. 

Furthermore, \autoref{sfig:enforced-l2norm} suggest a low layer-wise (defined in \cite{nguyen2025surveyMU}) or $L^2$ distance (defined in \cite{zhang2023reviewMU, wang2024comprehensiveMU}), \autoref{sfig:enforced-speedup} indicates high efficiency (defined in \cite{zhang2023reviewMU}), while the\% Balanced accuracy, \% Retained and Lost from \autoref{fig:perf-breastcancer-all}, \autoref{fig:perf_bank_adult_diab} and \autoref{fig:enf-performance} suggest a higher (balanced) accuracy and completeness (defined in \cite{nguyen2025surveyMU}), as well as higher effectiveness and certifiability (defined in \cite{zhang2023reviewMU}), thereby indicating that the proposed unlearning strategy has been successful in its intended objective. Furthermore, our \emph{robust} unlearning strategy, which is a trade-off between \emph{pure} unlearning and retraining, provide higher reassurance about unlearning of the samples in $\mathcal{D}^f$, due to the added impact of other samples from the $\mathcal{D}^r$, albeit at a reduced speedup (\autoref{fig:speedup} vs \autoref{sfig:enforced-speedup}). This strategy is also useful for model adaptation by learning a few new samples, in a reliable but resource-efficient manner, without having to undergo the full retraining.
 
\section{Limitation and Future Work}\label{sec:Limits}
Although the evaluation of the presented strategy against external attacks is outside the scope of this paper (limitation of this work), as this unlearning technique does not require access to the whole of the original training data, its application does not elevate the risk of exposure for the `remembered' datasets ($\mathcal{D}^r$). Other two limitations of this paper are: (i) it has been evaluated only for binary class problems; and (ii) the proposed strategies have not been compared against any model-agnostic unlearning strategy. Although a conceptual similarity has been noted for unlearning sample effects in Support Vector Machines and KNN classifiers, a systematic comparison of these have not been performed in this work.

Future work should investigate this unlearning strategy not just in terms of attacks but also in terms of unlearning features, to optimize its application on the adaptive distance based variants of GLVQ. 

\section{Conclusion}\label{sec:Conclusion}
In this paper we present a single-step (batch) gradient based robust approximate unlearning strategy for prototype based models, for the purpose of sample deletion. Following the comparison strategies in the published works of \cite{brophy2021UnlearnRF, thudi2022unrollingSGD, kumar2026controllingMU} we compared the unlearning efficacy of our proposed techniques to their corresponding retrained models. Furthermore, unlearning strategies for RF require either that the RF is composed of extreme RTs or training is done via a specific unlearning-enabled learning scheme, such as in \cite{brophy2021UnlearnRF}. However the methods we present here can be directly applied for unlearning samples from any GLVQ model already trained with the Sklvq toolbox \cite{vanVeen2021}, or translated in a straightforward manner for any other LVQ toolbox in Python or in MATLAB, following \autoref{sec:code map}. 

We investigated the efficacy of these unlearning frameworks in terms of the speed-up it offers relative to a complete retraining of the model on the `remembered' dataset $\mathcal{D}^r$, as well as the deviation of model parameters as a result of removal of the impact of the samples from the forget set $\mathcal{D}^f$, across different sizes of this forget set. Our findings show that unless the size of the forget set is significantly large compared to the original size of the dataset, \emph{pure} unlearning provides a much more resource-efficient alternative to retraining; for the other scenarios, our \emph{robust} unlearning framework is more suitable alternative to both \emph{pure} unlearning as well as full retraining. Our robust \emph{unlearning} framework can be also be used for learning small batch of new samples as well, without having to perform a full retraining. 

Since GLVQ and our proposed unlearning strategy for it, are both intrinsically interpretable, it is possible to extract \emph{proof of sample effect deletion} after unlearning as shown in \autoref{sfig:bc-prot-dev-bars} and \autoref{fig:distr} for the `Breast cancer' dataset. Thus, when used together with our proposed unlearning strategy, the LVQ family of algorithms become more easily adaptable to relevant regulations and governance requirement. This consequently reduces the post-deployment technical and compute costs and hassles of any enterprise that adopted the LVQ-family of classifiers to address their needs. 

\section*{Acknowledgment}
M. M was supported by the project HAICu with file number NWA.1518.22.105. of the research programme Onderzoek op Routes door Consortia 2022 which is financed by the Dutch Research Council (NWO).
\bibliography{refs}
\appendix
\counterwithin{figure}{section} 
\setcounter{figure}{0}
\setcounter{table}{0}
\renewcommand{\thefigure}{A.\arabic{figure}}
\renewcommand{\thetable}{A\arabic{table}}
\renewcommand{\theequation}{A-\arabic{equation}}
\setcounter{equation}{0}

\section*{Appendix}

\counterwithin{figure}{section} 
\setcounter{table}{0}
\renewcommand{\thetable}{A\arabic{table}}
\setcounter{equation}{0}
\renewcommand{\theequation}{A-\arabic{equation}}

\section{Compute resource specifications.}
All experiments, including model design and development were carried out in a Victus by HP Gaming Laptop 15, running on 13th Gen Intel Core i7 processor, refresh rate of 144Hz, and memory of 32GiB.
\section{Hyperparameters.}\label{sss:hyperparam}
\begin{table}
    \caption{$\beta$ in the Swish activation function and learning rate ($alpha$) and max runs in Steepest Gradient descent across different datasets.}\label{tab:beta_activation}
    \centering
    \begin{tabular}{p{1.38cm}|p{1.1cm}p{1.4cm}p{1.4cm}p{1.1cm}p{1.2cm}p{0.9cm}}
          & \textbf{Breast cancer} & \textbf{Surgical} & \textbf{Banking} & \textbf{Adult} & \textbf{Diabetes} &\textbf{ Criteo } \\
         \toprule
        $\beta_{ppc=1}$   & 5 &      15&     18 &        15  &   10      & 10    \\ \hline
        $\beta_{ppc=2}$   & 5 &      5 &     20 &        10  &   10      & 10    \\ \hline
        $\alpha$ & 0.05 & 0.05 & 0.05   & 0.05 & 0.05 & 0.075\\ \hline
       max\_runs & 5 & 5 & 5   & 5 & 5 & 3\\
         \bottomrule
    \end{tabular}
\end{table}
The training of even the largest dataset (`Criteo') was performed in a single batch. Other hyperparameters are in \autoref{tab:beta_activation} and:
\begin{itemize}[label=$\bullet$]
    \item Activation function: Swish
    \item Dissimilarity: Squared euclidean
    \item Discriminant (cost) function: Relative distance. 
\end{itemize}
\section{Unlearning equation to Sklvq toolbox mapping}\label{sec:code map}
To help readers reproduce the results and understand the code we provided, we provide here a mapping of the equations related to GLVQ, including distances, activation functions, discriminant type, and their corresponding functions from the Sklvq toolbox\cite{vanVeen2021}. As we focussed were on Squared Euclidean distance and `Swish' activation function we provided the mapping with respect to these. For other distances in the Sklvq toolbox, one can replace the `SquaredEuclidean()' with the corresponding distance. Similarly for activation function.

\subsection{Equation to function mapping}
\begin{itemize}[label=$\bullet$]
    \item dist $\Big(d(\vec{w}, \vec{\mathcal{X}}) = (\vec{\mathcal{X}} - \vec{w})^{\top} (\vec{\mathcal{X}} - \vec{w})\Big)$ 
     =sklvq.distances.SquaredEuclidean()
    \item costFunc$\Big( \mu(\vec{\mathcal{X}}) = \frac{d(\vec{\mathcal{X}}, \vec{w}^J) - d(\vec{\mathcal{X}}, \vec{w}^K)}{d(
            \vec{\mathcal{X}}, \vec{w}^J) + d(\vec{\mathcal{X}}, \vec{w}^K)}\Big)$ =sklvq.discriminants.RelativeDistance()
    \item activation\_grad $\Bigg(\frac{\partial f}{\partial \mathbf{\mu}} = \beta \cdot f(\mathbf{\mu}) + (\frac{1}{1 + e^{-\beta \cdot \mathbf{\mu}}}) \cdot (1 - \beta \cdot f(\mathbf{\mu}))\Bigg)$\\ =sklvq.activations.Swish.gradient( sklvq.activations.Swish(model.activation\_params[`beta']),\\ unlearn\_data) 
    \item Updating closest correct prototype
        \begin{itemize}
            \item dist\_grad\_same $\Big(\frac{\partial d}{\partial \vec{w}^J} = -2 \cdot (\vec{\mathcal{X}} - \vec{w}^J)\Big)$ \\
            =sklvq.distances.SquaredEuclidean.gradient(sklvq.distances.SquaredEuclidean(), \\
            \indent data=unlearn\_data, model=glvq, i\_prototype=idx)    
            \item  discr\_grad\_same: $\Bigg(\frac{\partial \mu}{\partial \vec{w}^J} = \frac{2 \cdot d(\vec{\mathcal{X}},
                \vec{w}^K)}{(d(\vec{\mathcal{X}}, \vec{w}^J) + d(\vec{\mathcal{X}}, \vec{w}^K))^2}.\Bigg)$\\ =sklvq.discriminants.RelativeDistance.gradient(sklvq.discriminants.RelativeDistance(), \\
        \indent dist\_same=dist\_same, dist\_diff= dist\_diff, same\_label=True) 
        \item grad\_prot\_same $\Bigg(\partial \vec{w}^J=\frac{2 \cdot d(\vec{\mathcal{X}}, \vec{w}^K) (-2 \cdot (\vec{\mathcal{X}} - \vec{w}^J))}{(d(\vec{\mathcal{X}}, \vec{w}^J) + d(\vec{\mathcal{X}}, \vec{w}^K))^2}=\frac{-4\cdot(\vec{\mathcal{X}} - \vec{w}^J)d(\vec{\mathcal{X}},\vec{w}^K)}{(d(\vec{\mathcal{X}}, \vec{w}^J) + d(\vec{\mathcal{X}}, \vec{w}^K))^2}\Bigg)$ \\
            =dist\_grad\_same$\times$ discr\_grad\_same 
          \item objective\_prot\_grad\_same=activation\_grad$\times$grad\_prot\_same
        \end{itemize}
    \item Updating closest incorrect prototype
        \begin{itemize}
            \item dist\_grad\_diff $\Big(\frac{\partial d}{\partial \vec{w}^K} = -2 \cdot (\vec{\mathcal{X}} - \vec{w}^K)\Big)$\\
            =sklvq.distances.SquaredEuclidean.gradient(sklvq.distances.SquaredEuclidean(), \\
               \indent      data=unlearn\_data, model=glvq, i\_prototype=idx)
            \item discr\_grad\_diff  $\Big(\frac{\partial \mu}{\partial \vec{w}^J} = \frac{-2 \cdot d(\vec{\mathcal{X}},
                \vec{w}^K)}{(d(\vec{\mathcal{X}}, \vec{w}^J) + d(\vec{\mathcal{X}}, \vec{w}^K))^2}\Big)$\\
            =sklvq.discriminants.RelativeDistance.gradient(sklvq.discriminants.RelativeDistance(), \\
            \indent dist\_same=dist\_same, dist\_diff= dist\_diff, same\_label=False)
              \item grad\_prot\_diff $\Bigg(\partial \vec{w}^K=\frac{-2 \cdot d(\vec{\mathcal{X}},
                \vec{w}^J) (-2 \cdot (\vec{\mathcal{X}} - \vec{w}^K))}{(d(\vec{\mathcal{X}}, \vec{w}^K) + d(\vec{\mathcal{X}}, \vec{w}^J)^2}=\frac{4\cdot(\vec{\mathcal{X}}- \vec{w}^K)d(\vec{\mathcal{X}},
                \vec{w}^J)}{(d(\vec{\mathcal{X}}, \vec{w}^K) + d(\vec{\mathcal{X}}, \vec{w}^J))^2}\Bigg)$ \\
              =dist\_grad\_diff$\times$ discr\_grad\_diff
            \item objective\_prot\_grad\_diff=activation\_grad$\times$grad\_prot\_diff
        \end{itemize}
\end{itemize}

\section{Analysis of Unlearning-only framework on the benchmark datasets}
This section contains the tabulated performances of the Unlearned (M2) and Retrained (M1) models on the `Surgical', `Banking', `Adult', `Diabetes', and `Criteo' datasets.   
\subsection{Surgical, Banking, Adult, Diabetes, Criteo datasets}
\begin{figure}
 \centering
     \begin{subfigure}[!h]{0.49\textwidth}
     \includegraphics[width=\textwidth]{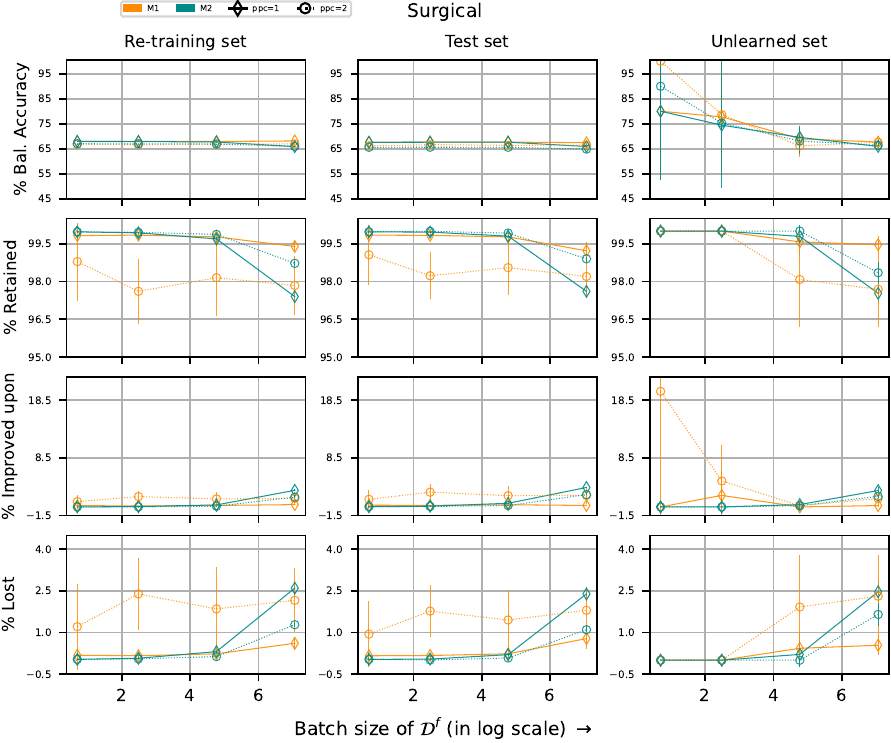}
     \caption{Performance of M1 vs M2 models on the `Surgical' dataset.}
     \label{fig:perf_surgical}
    \end{subfigure}\hfill
     \begin{subfigure}[!h]{0.49\textwidth}
     \centering
         \includegraphics[width=\textwidth]{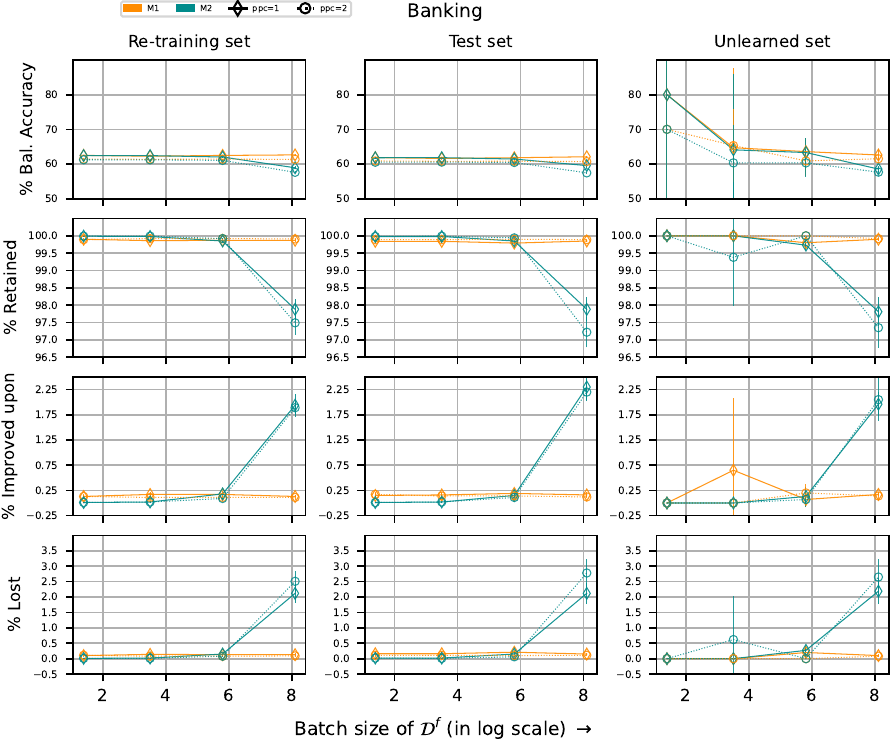}
         \caption{Banking dataset}
         \label{sfig:banking}
     \end{subfigure}\hfill
    \begin{subfigure}[!h]{0.49\textwidth}
     \centering
         \includegraphics[width=\textwidth]{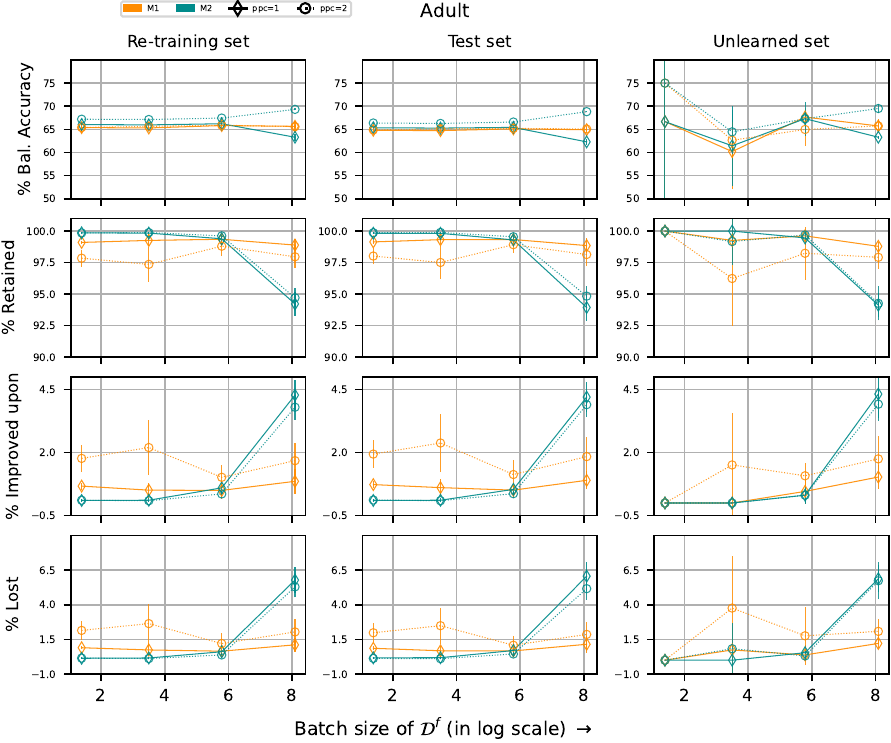}
         \caption{Adult dataset}
         \label{sfig:adult}
     \end{subfigure}\hfill
     \begin{subfigure}[!h]{0.49\textwidth}
     \centering
         \includegraphics[width=\textwidth]{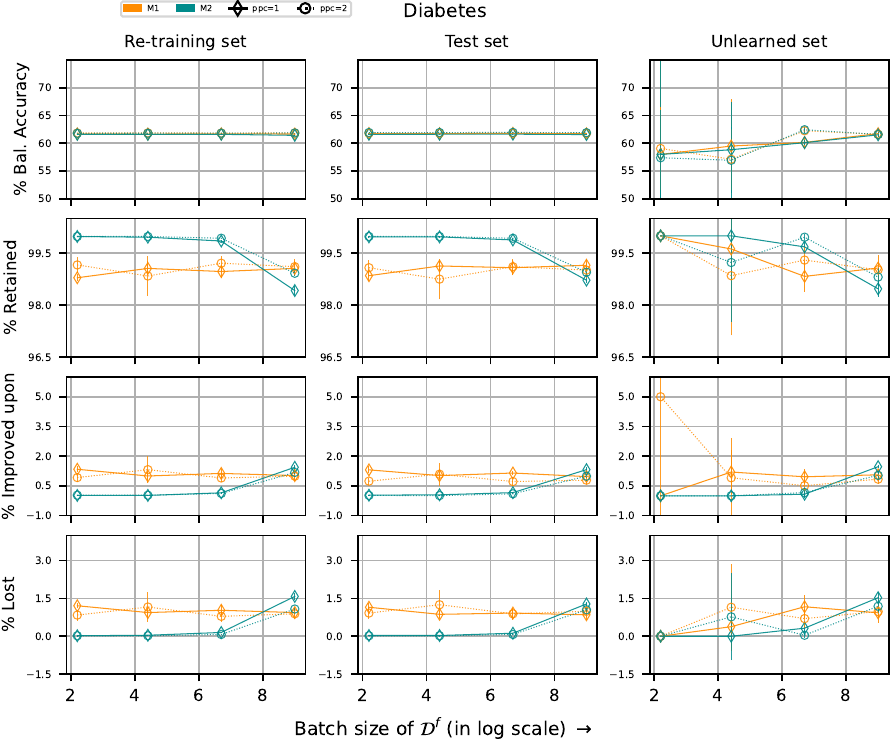}
         \caption{Diabetes dataset}
         \label{sfig:diabetes}
     \end{subfigure}\hfill
       \begin{subfigure}[!h]{0.49\textwidth}
     \centering
         \includegraphics[width=\textwidth]{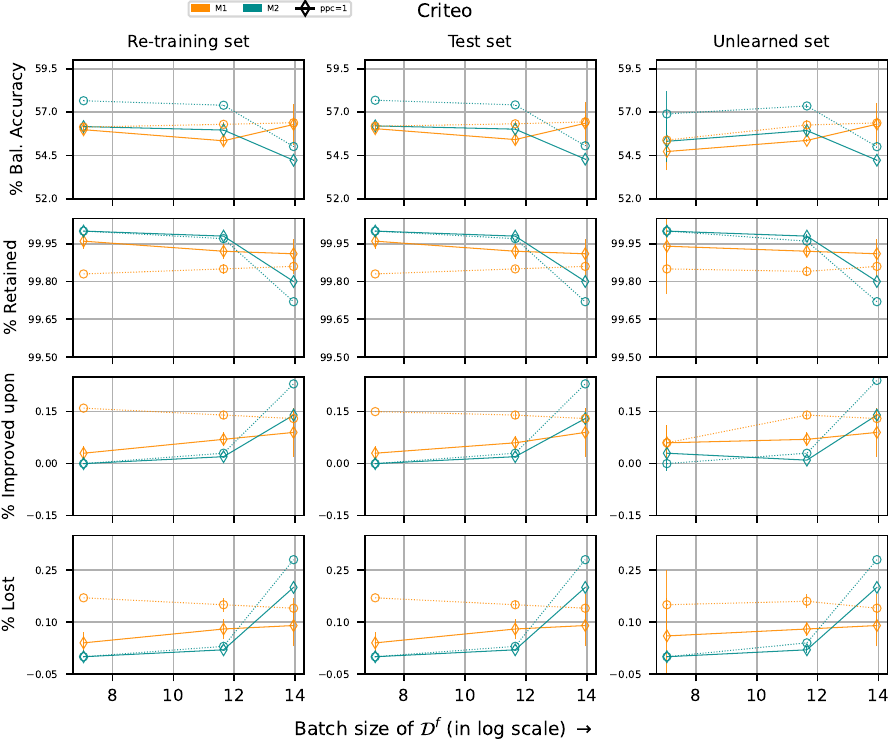}
         \caption{Criteo dataset}
         \label{sfig:criteo}
     \end{subfigure}
     \caption{Performances on the Banking, Adult, Diabetes and Criteo datasets. }
     \label{fig:perf_bank_adult_diab}
 \end{figure}

The performance of the retrained and unlearned models for the `Surgical', `Banking', 'Adult' and `Diabetes' and `Criteo' datasets are as illustrated in \autoref{fig:perf_bank_adult_diab}. The metrics are defined in \autoref{ssec: eval}. 
{\footnotesize
\begin{longtable}{p{1cm}p{1.8cm}p{1.0cm}p{2cm}p{2cm}p{2cm}p{2cm}p{2cm}}
    \caption{Comparison performance of the retrained and unlearned models across different batch sizes of $\mathcal{D}^f$ (N($\mathcal{D}^f$)) in the `Surgical' dataset, reported as mean (std) over 5 iterations.}\label{tab:_perf_surgical}\\
    \toprule
ppc	&	N($\mathcal{D}^f$)	&	Subset & Model	& \% Bal. Acc	& \% Gained	&	\% Lost	& \% Retained	\\
\toprule
1	&	2	&	retraining	&	M1	&	67.91 (0.07)	&	0.24 (0.06)	&	0.18 (0.03)	&	99.82 (0.03)	\\	
	&		&		&	M2	&	67.9 (0.03)	&	0.01 (0.01)	&	0.03 (0.02)	&	99.97 (0.02)	\\	\cmidrule{3-8}
	&		&	test	&	M1	&	67.62 (0.2)	&	0.3 (0.11)	&	0.16 (0.14)	&	99.84 (0.14)	\\	
	&		&		&	M2	&	67.55 (0.02)	&	0.03 (0.02)	&	0.03 (0.02)	&	99.97 (0.02)	\\	\cmidrule{3-8}
	&		&	unlearned	&	M1	&	80.0 (27.39)	&	0.0 (0.0)	&	0.0 (0.0)	&	100.0 (0.0)	\\	
	&		&		&	M2	&	80.0 (27.39)	&	0.0 (0.0)	&	0.0 (0.0)	&	100.0 (0.0)	\\	\cmidrule{2-8}
	&	12	&	retraining	&	M1	&	67.91 (0.01)	&	0.16 (0.04)	&	0.16 (0.03)	&	99.84 (0.03)	\\	
	&		&		&	M2	&	67.87 (0.04)	&	0.06 (0.04)	&	0.07 (0.04)	&	99.93 (0.04)	\\	\cmidrule{3-8}
	&		&	test	&	M1	&	67.63 (0.14)	&	0.27 (0.05)	&	0.17 (0.08)	&	99.83 (0.08)	\\	
	&		&		&	M2	&	67.6 (0.06)	&	0.13 (0.06)	&	0.04 (0.03)	&	99.96 (0.03)	\\	\cmidrule{3-8}
	&		&	unlearned	&	M1	&	77.75 (10.97)	&	2.0 (4.47)	&	0.0 (0.0)	&	100.0 (0.0)	\\	
	&		&		&	M2	&	74.42 (11.37)	&	0.0 (0.0)	&	0.0 (0.0)	&	100.0 (0.0)	\\	\cmidrule{2-8}
	&	118	&	retraining	&	M1	&	67.88 (0.11)	&	0.26 (0.09)	&	0.23 (0.1)	&	99.77 (0.1)	\\	
	&		&		&	M2	&	67.68 (0.06)	&	0.35 (0.11)	&	0.31 (0.07)	&	99.69 (0.07)	\\	\cmidrule{3-8}
	&		&	test	&	M1	&	67.65 (0.2)	&	0.38 (0.16)	&	0.23 (0.19)	&	99.77 (0.19)	\\	
	&		&		&	M2	&	67.62 (0.11)	&	0.63 (0.16)	&	0.2 (0.04)	&	99.8 (0.04)	\\	\cmidrule{3-8}
	&		&	unlearned	&	M1	&	68.98 (4.06)	&	0.0 (0.0)	&	0.43 (0.59)	&	99.57 (0.59)	\\	
	&		&		&	M2	&	69.54 (4.4)	&	0.41 (0.57)	&	0.21 (0.47)	&	99.79 (0.47)	\\	\cmidrule{2-8}
	&	1171	&	retraining	&	M1	&	68.11 (0.09)	&	0.4 (0.09)	&	0.61 (0.23)	&	99.39 (0.23)	\\	
	&		&		&	M2	&	65.85 (0.06)	&	2.87 (0.18)	&	2.6 (0.17)	&	97.4 (0.17)	\\	\cmidrule{3-8}
	&		&	test	&	M1	&	67.4 (0.12)	&	0.26 (0.1)	&	0.78 (0.36)	&	99.22 (0.36)	\\	
	&		&		&	M2	&	65.89 (0.29)	&	3.37 (0.15)	&	2.39 (0.19)	&	97.61 (0.19)	\\	\cmidrule{3-8}
	&		&	unlearned	&	M1	&	67.66 (0.98)	&	0.24 (0.18)	&	0.54 (0.34)	&	99.46 (0.34)	\\	
	&		&		&	M2	&	65.92 (1.02)	&	2.84 (0.58)	&	2.47 (0.37)	&	97.53 (0.37)	\\	\hline
2	&	2	&	retraining	&	M1	&	66.82 (0.18)	&	0.92 (1.19)	&	1.21 (1.55)	&	98.79 (1.55)	\\	
	&		&		&	M2	&	66.98 (0.02)	&	0.0 (0.0)	&	0.03 (0.02)	&	99.97 (0.02)	\\	\cmidrule{3-8}
	&		&	test	&	M1	&	66.07 (0.71)	&	1.32 (1.66)	&	0.94 (1.19)	&	99.06 (1.19)	\\	
	&		&		&	M2	&	65.56 (0.04)	&	0.04 (0.04)	&	0.03 (0.06)	&	99.97 (0.06)	\\	\cmidrule{3-8}
	&		&	unlearned	&	M1	&	100.0 (0.0)	&	20.0 (44.72)	&	0.0 (0.0)	&	100.0 (0.0)	\\	
	&		&		&	M2	&	90.0 (22.36)	&	0.0 (0.0)	&	0.0 (0.0)	&	100.0 (0.0)	\\	\cmidrule{2-8}
	&	12	&	retraining	&	M1	&	66.65 (0.21)	&	1.79 (0.97)	&	2.39 (1.3)	&	97.61 (1.3)	\\	
	&		&		&	M2	&	66.97 (0.03)	&	0.01 (0.01)	&	0.05 (0.02)	&	99.95 (0.02)	\\	\cmidrule{3-8}
	&		&	test	&	M1	&	66.63 (0.58)	&	2.56 (1.33)	&	1.77 (0.94)	&	98.23 (0.94)	\\	
	&		&		&	M2	&	65.59 (0.03)	&	0.09 (0.05)	&	0.01 (0.02)	&	99.99 (0.02)	\\	\cmidrule{3-8}
	&		&	unlearned	&	M1	&	78.66 (24.08)	&	4.5 (6.22)	&	0.0 (0.0)	&	100.0 (0.0)	\\	
	&		&		&	M2	&	75.25 (25.93)	&	0.0 (0.0)	&	0.0 (0.0)	&	100.0 (0.0)	\\	\cmidrule{2-8}
	&	118	&	retraining	&	M1	&	66.77 (0.24)	&	1.39 (1.18)	&	1.85 (1.52)	&	98.15 (1.52)	\\	
	&		&		&	M2	&	66.88 (0.04)	&	0.13 (0.04)	&	0.13 (0.01)	&	99.87 (0.01)	\\	\cmidrule{3-8}
	&		&	test	&	M1	&	66.34 (0.69)	&	1.95 (1.62)	&	1.45 (1.09)	&	98.55 (1.09)	\\	
	&		&		&	M2	&	65.55 (0.03)	&	0.24 (0.02)	&	0.08 (0.02)	&	99.92 (0.02)	\\	\cmidrule{3-8}
	&		&	unlearned	&	M1	&	66.13 (4.35)	&	0.21 (0.48)	&	1.92 (1.89)	&	98.08 (1.89)	\\	
	&		&		&	M2	&	68.08 (5.15)	&	0.21 (0.48)	&	0.0 (0.0)	&	100.0 (0.0)	\\	\cmidrule{2-8}
	&	1171	&	retraining	&	M1	&	66.92 (0.25)	&	1.49 (1.04)	&	2.16 (1.16)	&	97.84 (1.16)	\\	
	&		&		&	M2	&	65.92 (0.16)	&	1.7 (0.11)	&	1.28 (0.1)	&	98.72 (0.1)	\\	\cmidrule{3-8}
	&		&	test	&	M1	&	66.48 (0.78)	&	2.01 (1.5)	&	1.8 (0.84)	&	98.2 (0.84)	\\	
	&		&		&	M2	&	64.85 (0.19)	&	2.13 (0.11)	&	1.1 (0.13)	&	98.9 (0.13)	\\	\cmidrule{3-8}
	&		&	unlearned	&	M1	&	67.56 (1.43)	&	1.44 (1.06)	&	2.31 (1.48)	&	97.69 (1.48)	\\	
	&		&		&	M2	&	66.36 (1.58)	&	1.83 (0.29)	&	1.65 (0.43)	&	98.35 (0.43)	\\	\bottomrule
 \end{longtable}
}
{\footnotesize
    \begin{longtable}{p{1.5cm}p{1.8cm}p{1.2 cm}p{1.32cm}p{1.6cm}p{1.5cm}p{1.5cm}p{1.5cm}}
    \caption{\footnotesize{Comparison of performance of the retrained and unlearned models across different batch sizes of $\mathcal{D}^f$ (N($\mathcal{D}^f$)) in the `Banking' dataset, reported as mean (std) over 5 iterations.}}\label{tab:_perf_banking}\\
    \toprule
ppc	&	N($\mathcal{D}^f$)	&	Subset & Model	& \% Bal. Acc	& \% Gained	&	\% Lost	& \% Retained	\\
\toprule
1	&	4	&	retraining	&	M1	&	62.49 (0.23)	&	0.13 (0.03)	&	0.1 (0.04)	&	99.9 (0.04)	\\	
	&		&		&	M2	&	62.43 (0.04)	&	0.01 (0.01)	&	0.01 (0.01)	&	99.99 (0.01)	\\	\cmidrule{3-8}
	&		&	test	&	M1	&	61.85 (0.25)	&	0.15 (0.04)	&	0.16 (0.06)	&	99.84 (0.06)	\\	
	&		&		&	M2	&	61.81 (0.03)	&	0.01 (0.01)	&	0.02 (0.01)	&	99.98 (0.01)	\\	\cmidrule{3-8}
	&		&	unlearned	&	M1	&	80.0 (27.39)	&	0.0 (0.0)	&	0.0 (0.0)	&	100.0 (0.0)	\\	
	&		&		&	M2	&	80.0 (27.39)	&	0.0 (0.0)	&	0.0 (0.0)	&	100.0 (0.0)	\\	\cmidrule{2-8}
	&	33	&	retraining	&	M1	&	62.25 (0.32)	&	0.17 (0.04)	&	0.14 (0.04)	&	99.86 (0.04)	\\	
	&		&		&	M2	&	62.42 (0.04)	&	0.02 (0.01)	&	0.02 (0.01)	&	99.98 (0.01)	\\	\cmidrule{3-8}
	&		&	test	&	M1	&	61.61 (0.33)	&	0.16 (0.05)	&	0.16 (0.05)	&	99.84 (0.05)	\\	
	&		&		&	M2	&	61.83 (0.03)	&	0.02 (0.02)	&	0.02 (0.01)	&	99.98 (0.01)	\\	\cmidrule{3-8}
	&		&	unlearned	&	M1	&	64.7 (23.05)	&	0.65 (1.44)	&	0.0 (0.0)	&	100.0 (0.0)	\\	
	&		&		&	M2	&	64.09 (22.01)	&	0.0 (0.0)	&	0.0 (0.0)	&	100.0 (0.0)	\\	\cmidrule{2-8}
	&	330	&	retraining	&	M1	&	62.48 (0.17)	&	0.17 (0.07)	&	0.13 (0.05)	&	99.87 (0.05)	\\	
	&		&		&	M2	&	62.02 (0.04)	&	0.18 (0.01)	&	0.15 (0.01)	&	99.85 (0.01)	\\	\cmidrule{3-8}
	&		&	test	&	M1	&	61.81 (0.33)	&	0.19 (0.05)	&	0.21 (0.07)	&	99.79 (0.07)	\\	
	&		&		&	M2	&	61.44 (0.06)	&	0.15 (0.03)	&	0.15 (0.01)	&	99.85 (0.01)	\\	\cmidrule{3-8}
	&		&	unlearned	&	M1	&	63.58 (3.99)	&	0.07 (0.15)	&	0.2 (0.18)	&	99.8 (0.18)	\\	
	&		&		&	M2	&	63.31 (4.21)	&	0.13 (0.18)	&	0.27 (0.15)	&	99.73 (0.15)	\\	\cmidrule{2-8}
	&	3295	&	retraining	&	M1	&	62.63 (0.24)	&	0.13 (0.02)	&	0.13 (0.04)	&	99.87 (0.04)	\\	
	&		&		&	M2	&	58.91 (0.47)	&	1.94 (0.23)	&	2.12 (0.31)	&	97.88 (0.31)	\\	\cmidrule{3-8}
	&		&	test	&	M1	&	62.09 (0.34)	&	0.16 (0.05)	&	0.15 (0.04)	&	99.85 (0.04)	\\	
	&		&		&	M2	&	59.54 (0.56)	&	2.3 (0.28)	&	2.12 (0.35)	&	97.88 (0.35)	\\	\cmidrule{3-8}
	&		&	unlearned	&	M1	&	62.6 (1.75)	&	0.17 (0.04)	&	0.1 (0.04)	&	99.9 (0.04)	\\	
	&		&		&	M2	&	58.55 (0.92)	&	1.96 (0.24)	&	2.19 (0.42)	&	97.81 (0.42)	\\	\hline
2	&	4	&	retraining	&	M1	&	61.4 (0.34)	&	0.13 (0.05)	&	0.11 (0.02)	&	99.89 (0.02)	\\	
	&		&		&	M2	&	61.23 (0.02)	&	0.01 (0.01)	&	0.01 (0.0)	&	99.99 (0.0)	\\	\cmidrule{3-8}
	&		&	test	&	M1	&	60.88 (0.24)	&	0.17 (0.04)	&	0.11 (0.06)	&	99.89 (0.06)	\\	
	&		&		&	M2	&	60.52 (0.04)	&	0.01 (0.01)	&	0.02 (0.01)	&	99.98 (0.01)	\\	\cmidrule{3-8}
	&		&	unlearned	&	M1	&	70.0 (27.39)	&	0.0 (0.0)	&	0.0 (0.0)	&	100.0 (0.0)	\\	
	&		&		&	M2	&	70.0 (27.39)	&	0.0 (0.0)	&	0.0 (0.0)	&	100.0 (0.0)	\\	\cmidrule{2-8}
	&	33	&	retraining	&	M1	&	61.36 (0.13)	&	0.11 (0.04)	&	0.08 (0.04)	&	99.92 (0.04)	\\	
	&		&		&	M2	&	61.21 (0.02)	&	0.02 (0.02)	&	0.01 (0.01)	&	99.99 (0.01)	\\	\cmidrule{3-8}
	&		&	test	&	M1	&	60.77 (0.21)	&	0.14 (0.06)	&	0.11 (0.06)	&	99.89 (0.06)	\\	
	&		&		&	M2	&	60.55 (0.03)	&	0.02 (0.01)	&	0.01 (0.01)	&	99.99 (0.01)	\\	\cmidrule{3-8}
	&		&	unlearned	&	M1	&	65.33 (10.63)	&	0.0 (0.0)	&	0.0 (0.0)	&	100.0 (0.0)	\\	
	&		&		&	M2	&	60.33 (10.83)	&	0.0 (0.0)	&	0.62 (1.4)	&	99.38 (1.4)	\\	\cmidrule{2-8}
	&	330	&	retraining	&	M1	&	61.39 (0.12)	&	0.11 (0.03)	&	0.08 (0.03)	&	99.92 (0.03)	\\	
	&		&		&	M2	&	61.05 (0.09)	&	0.09 (0.01)	&	0.08 (0.02)	&	99.92 (0.02)	\\	\cmidrule{3-8}
	&		&	test	&	M1	&	60.81 (0.21)	&	0.14 (0.04)	&	0.1 (0.04)	&	99.9 (0.04)	\\	
	&		&		&	M2	&	60.45 (0.08)	&	0.11 (0.03)	&	0.06 (0.02)	&	99.94 (0.02)	\\	\cmidrule{3-8}
	&		&	unlearned	&	M1	&	60.89 (4.39)	&	0.2 (0.18)	&	0.0 (0.0)	&	100.0 (0.0)	\\	
	&		&		&	M2	&	60.32 (3.97)	&	0.07 (0.15)	&	0.0 (0.0)	&	100.0 (0.0)	\\	\cmidrule{2-8}
	&	3295	&	retraining	&	M1	&	61.35 (0.22)	&	0.11 (0.05)	&	0.09 (0.02)	&	99.91 (0.02)	\\	
	&		&		&	M2	&	57.61 (0.33)	&	1.89 (0.12)	&	2.51 (0.35)	&	97.49 (0.35)	\\	\cmidrule{3-8}
	&		&	test	&	M1	&	60.7 (0.25)	&	0.12 (0.03)	&	0.11 (0.03)	&	99.89 (0.03)	\\	
	&		&		&	M2	&	57.45 (0.34)	&	2.2 (0.16)	&	2.78 (0.44)	&	97.22 (0.44)	\\	\cmidrule{3-8}
	&		&	unlearned	&	M1	&	61.52 (0.64)	&	0.14 (0.08)	&	0.08 (0.04)	&	99.92 (0.04)	\\	
	&		&		&	M2	&	57.66 (0.45)	&	2.05 (0.42)	&	2.65 (0.58)	&	97.35 (0.58)	\\
\bottomrule
\end{longtable}

\begin{longtable}{p{1.5cm}p{1.8cm}p{1.2 cm}p{1.32cm}p{1.6cm}p{1.5cm}p{1.5cm}p{1.5cm}}
    \caption{\footnotesize{Comparison of performance of the retrained and unlearned models across different batch sizes of $\mathcal{D}^f$ (N($\mathcal{D}^f$)) in the `Adult' dataset, reported as mean (std) over 5 iterations.}}\label{tab:_perf_adult}\\
    \toprule
ppc	&	N($\mathcal{D}^f$)	&	Subset & Model	& \% Bal. Acc	& \% Gained	&	\% Lost	& \% Retained	\\
\toprule
1	&	4	&	retraining	&	M1	&	65.38 (0.7)	&	0.67 (0.14)	&	0.9 (0.12)	&	99.1 (0.12)	\\	
	&		&		&	M2	&	66.04 (0.17)	&	0.1 (0.06)	&	0.14 (0.05)	&	99.86 (0.05)	\\	\cmidrule{3-8}
	&		&	test	&	M1	&	64.79 (0.45)	&	0.73 (0.13)	&	0.86 (0.15)	&	99.14 (0.15)	\\	
	&		&		&	M2	&	65.34 (0.15)	&	0.1 (0.06)	&	0.17 (0.06)	&	99.83 (0.06)	\\	\cmidrule{3-8}
	&		&	unlearned	&	M1	&	66.67 (23.57)	&	0.0 (0.0)	&	0.0 (0.0)	&	100.0 (0.0)	\\	
	&		&		&	M2	&	66.67 (23.57)	&	0.0 (0.0)	&	0.0 (0.0)	&	100.0 (0.0)	\\	\cmidrule{2-8}
	&	33	&	retraining	&	M1	&	65.3 (0.6)	&	0.51 (0.3)	&	0.74 (0.47)	&	99.26 (0.47)	\\	
	&		&		&	M2	&	65.96 (0.2)	&	0.11 (0.07)	&	0.16 (0.09)	&	99.84 (0.09)	\\	\cmidrule{3-8}
	&		&	test	&	M1	&	64.73 (0.42)	&	0.61 (0.34)	&	0.68 (0.37)	&	99.32 (0.37)	\\	
	&		&		&	M2	&	65.27 (0.15)	&	0.11 (0.08)	&	0.18 (0.11)	&	99.82 (0.11)	\\	\cmidrule{3-8}
	&		&	unlearned	&	M1	&	60.19 (8.19)	&	0.0 (0.0)	&	0.74 (1.66)	&	99.26 (1.66)	\\	
	&		&		&	M2	&	61.44 (8.71)	&	0.0 (0.0)	&	0.0 (0.0)	&	100.0 (0.0)	\\	\cmidrule{2-8}
	&	326	&	retraining	&	M1	&	65.82 (0.64)	&	0.5 (0.33)	&	0.65 (0.35)	&	99.35 (0.35)	\\	
	&		&		&	M2	&	66.21 (0.21)	&	0.6 (0.22)	&	0.61 (0.21)	&	99.39 (0.21)	\\	\cmidrule{3-8}
	&		&	test	&	M1	&	65.1 (0.48)	&	0.51 (0.3)	&	0.67 (0.36)	&	99.33 (0.36)	\\	
	&		&		&	M2	&	65.42 (0.1)	&	0.54 (0.19)	&	0.7 (0.24)	&	99.3 (0.24)	\\	\cmidrule{3-8}
	&		&	unlearned	&	M1	&	67.66 (2.05)	&	0.46 (0.49)	&	0.38 (0.27)	&	99.62 (0.27)	\\	
	&		&		&	M2	&	67.31 (1.39)	&	0.31 (0.32)	&	0.54 (0.59)	&	99.46 (0.59)	\\	\cmidrule{2-8}
	&	3257	&	retraining	&	M1	&	65.65 (1.11)	&	0.86 (0.49)	&	1.11 (0.5)	&	98.89 (0.5)	\\	
	&		&		&	M2	&	63.29 (0.71)	&	4.28 (0.61)	&	5.78 (0.92)	&	94.22 (0.92)	\\	\cmidrule{3-8}
	&		&	test	&	M1	&	64.92 (0.77)	&	0.9 (0.35)	&	1.15 (0.62)	&	98.85 (0.62)	\\	
	&		&		&	M2	&	62.29 (0.6)	&	4.2 (0.61)	&	6.06 (1.05)	&	93.94 (1.05)	\\	\cmidrule{3-8}
	&		&	unlearned	&	M1	&	65.73 (0.75)	&	1.03 (0.47)	&	1.21 (0.35)	&	98.79 (0.35)	\\	
	&		&		&	M2	&	63.29 (0.32)	&	4.32 (0.95)	&	5.88 (0.95)	&	94.12 (0.95)	\\	\hline
2	&	4	&	retraining	&	M1	&	65.4 (1.11)	&	1.77 (0.55)	&	2.15 (0.66)	&	97.85 (0.66)	\\	
	&		&		&	M2	&	67.19 (0.03)	&	0.11 (0.03)	&	0.15 (0.05)	&	99.85 (0.05)	\\	\cmidrule{3-8}
	&		&	test	&	M1	&	64.87 (0.93)	&	1.94 (0.55)	&	1.98 (0.67)	&	98.02 (0.67)	\\	
	&		&		&	M2	&	66.37 (0.03)	&	0.12 (0.03)	&	0.13 (0.04)	&	99.87 (0.04)	\\	\cmidrule{3-8}
	&		&	unlearned	&	M1	&	75.0 (25.0)	&	0.0 (0.0)	&	0.0 (0.0)	&	100.0 (0.0)	\\	
	&		&		&	M2	&	75.0 (25.0)	&	0.0 (0.0)	&	0.0 (0.0)	&	100.0 (0.0)	\\	\cmidrule{2-8}
	&	33	&	retraining	&	M1	&	65.64 (0.97)	&	2.2 (1.09)	&	2.64 (1.39)	&	97.36 (1.39)	\\	
	&		&		&	M2	&	67.13 (0.13)	&	0.09 (0.03)	&	0.12 (0.05)	&	99.88 (0.05)	\\	\cmidrule{3-8}
	&		&	test	&	M1	&	65.11 (0.72)	&	2.38 (1.16)	&	2.49 (1.29)	&	97.51 (1.29)	\\	
	&		&		&	M2	&	66.28 (0.09)	&	0.09 (0.03)	&	0.11 (0.05)	&	99.89 (0.05)	\\	\cmidrule{3-8}
	&		&	unlearned	&	M1	&	62.64 (3.93)	&	1.51 (2.07)	&	3.76 (3.78)	&	96.24 (3.78)	\\	
	&		&		&	M2	&	64.46 (4.78)	&	0.0 (0.0)	&	0.83 (1.86)	&	99.17 (1.86)	\\	\cmidrule{2-8}
	&	326	&	retraining	&	M1	&	65.89 (1.28)	&	1.02 (0.48)	&	1.2 (0.78)	&	98.8 (0.78)	\\	
	&		&		&	M2	&	67.44 (0.21)	&	0.36 (0.16)	&	0.38 (0.16)	&	99.62 (0.16)	\\	\cmidrule{3-8}
	&		&	test	&	M1	&	65.21 (1.03)	&	1.13 (0.59)	&	1.08 (0.66)	&	98.92 (0.66)	\\	
	&		&		&	M2	&	66.58 (0.19)	&	0.37 (0.16)	&	0.45 (0.16)	&	99.55 (0.16)	\\	\cmidrule{3-8}
	&		&	unlearned	&	M1	&	64.97 (3.59)	&	1.08 (0.51)	&	1.76 (2.11)	&	98.24 (2.11)	\\	
	&		&		&	M2	&	67.28 (3.67)	&	0.31 (0.33)	&	0.31 (0.33)	&	99.69 (0.33)	\\	\cmidrule{2-8}
	&	3257	&	retraining	&	M1	&	65.59 (0.37)	&	1.68 (0.71)	&	2.04 (0.9)	&	97.96 (0.9)	\\	
	&		&		&	M2	&	69.32 (0.22)	&	3.8 (0.49)	&	5.29 (0.76)	&	94.71 (0.76)	\\	\cmidrule{3-8}
	&		&	test	&	M1	&	65.07 (0.45)	&	1.84 (0.76)	&	1.86 (0.91)	&	98.14 (0.91)	\\	
	&		&		&	M2	&	68.83 (0.16)	&	3.9 (0.48)	&	5.16 (0.8)	&	94.84 (0.8)	\\	\cmidrule{3-8}
	&		&	unlearned	&	M1	&	65.74 (1.42)	&	1.75 (0.92)	&	2.08 (0.92)	&	97.92 (0.92)	\\	
	&		&		&	M2	&	69.51 (1.05)	&	3.92 (0.65)	&	5.73 (1.35)	&	94.27 (1.35)	\\	
\bottomrule
\end{longtable}
}

\section{Analysis of Unlearning with enforcer samples}
\begin{figure}
    \centering
     \begin{subfigure}[!h]{0.49\textwidth}
     \centering
        \includegraphics[
        width=\textwidth]{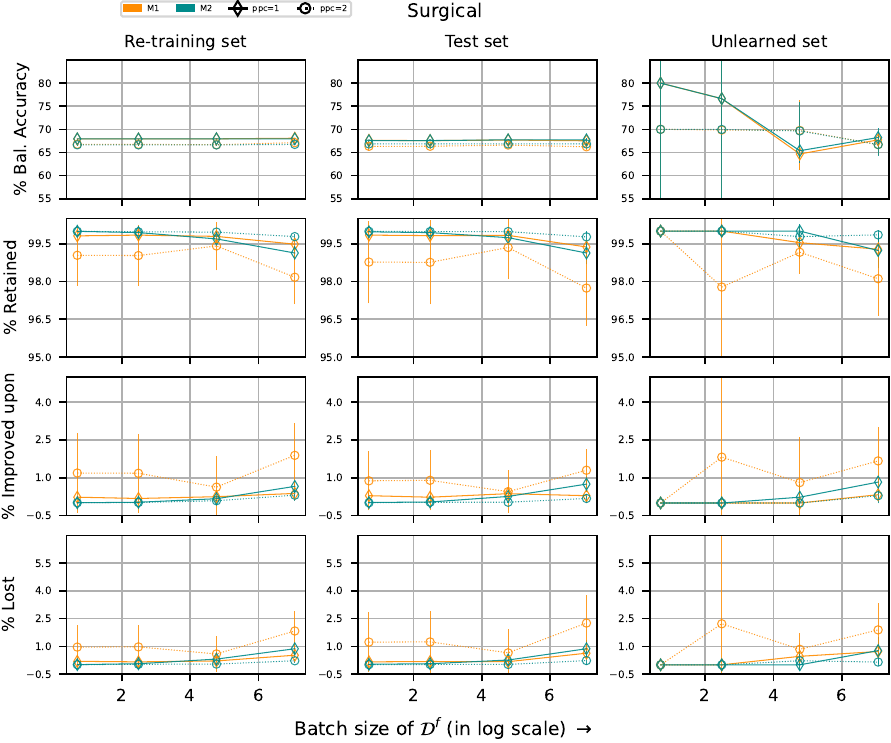}
        \caption{Surgical}
        \label{sfig:enf-surgical}
    \end{subfigure}\hfill
    \begin{subfigure}[!h]{0.49\textwidth}
    \centering
        \includegraphics[width=\textwidth]{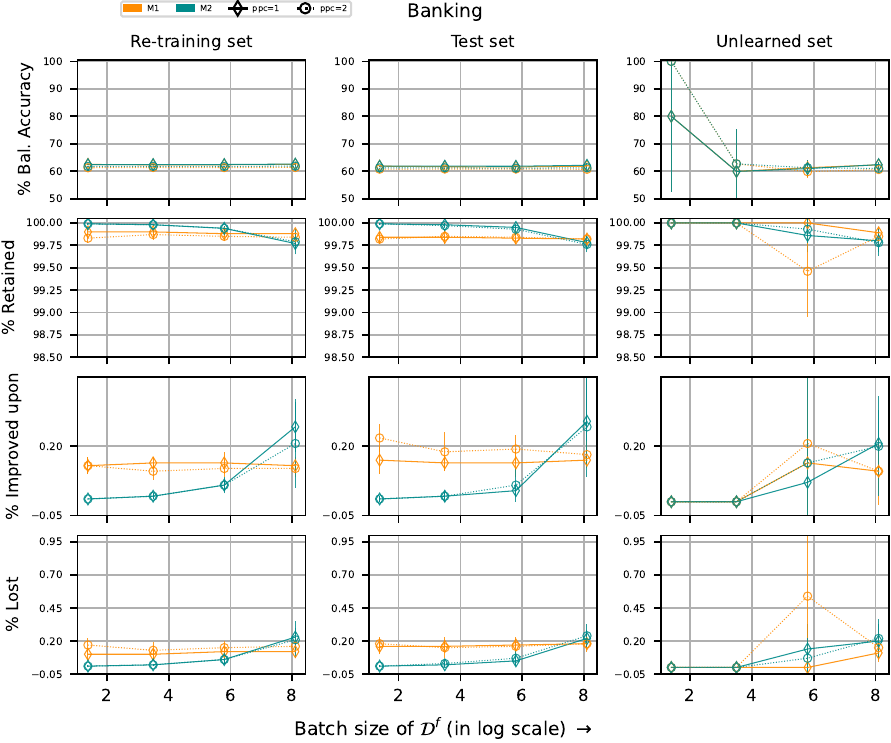}
        \caption{Banking}
        \label{sfig:enf-banking}
    \end{subfigure}\hfill
    \begin{subfigure}[!h]{0.49\textwidth}
    \centering
        \includegraphics[width=\textwidth]{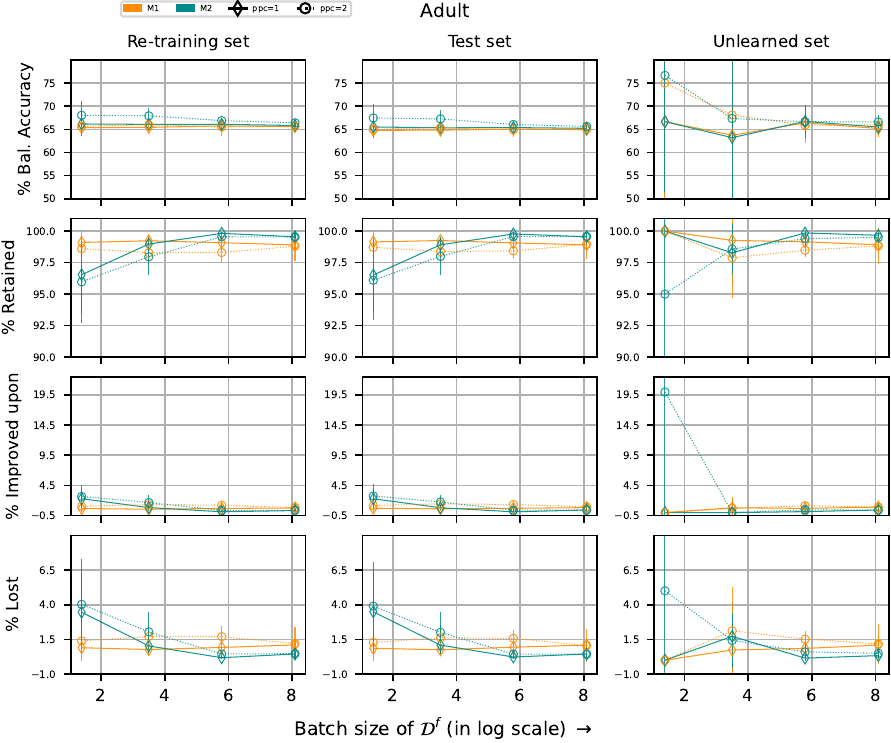}
        \caption{Adult}
        \label{sfig:enf-adult}
    \end{subfigure} \hfill
    \begin{subfigure}[!h]{0.49\textwidth}
    \centering
        \includegraphics[width=\textwidth]{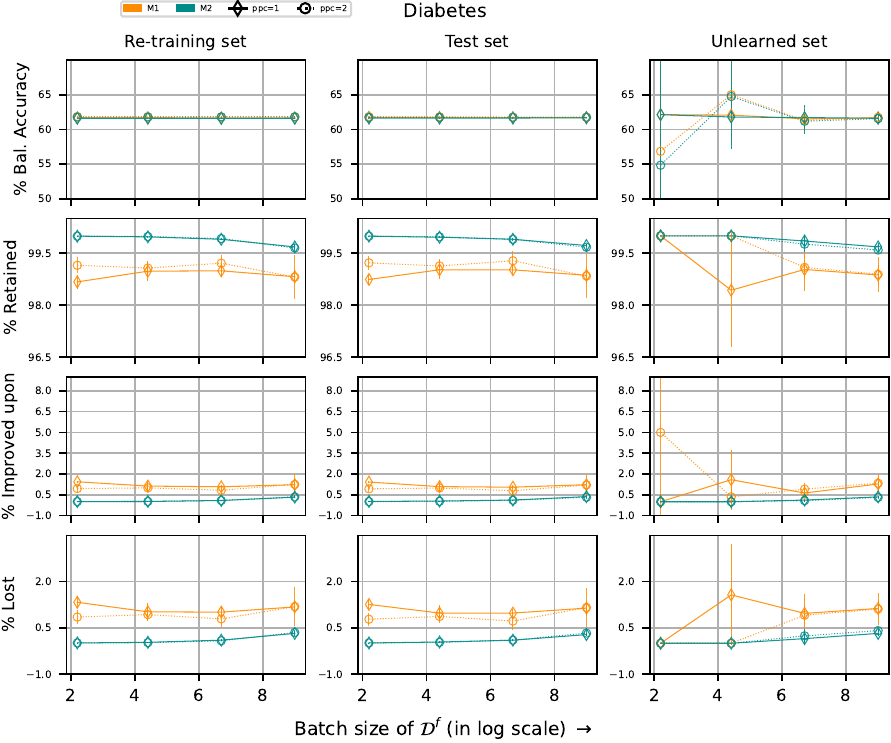}
        \caption{Diabetes}
        \label{sfig:enf-diabetes}
    \end{subfigure} \hfill
        \begin{subfigure}[!h]{0.49\textwidth}
    \centering
        \includegraphics[width=\textwidth]{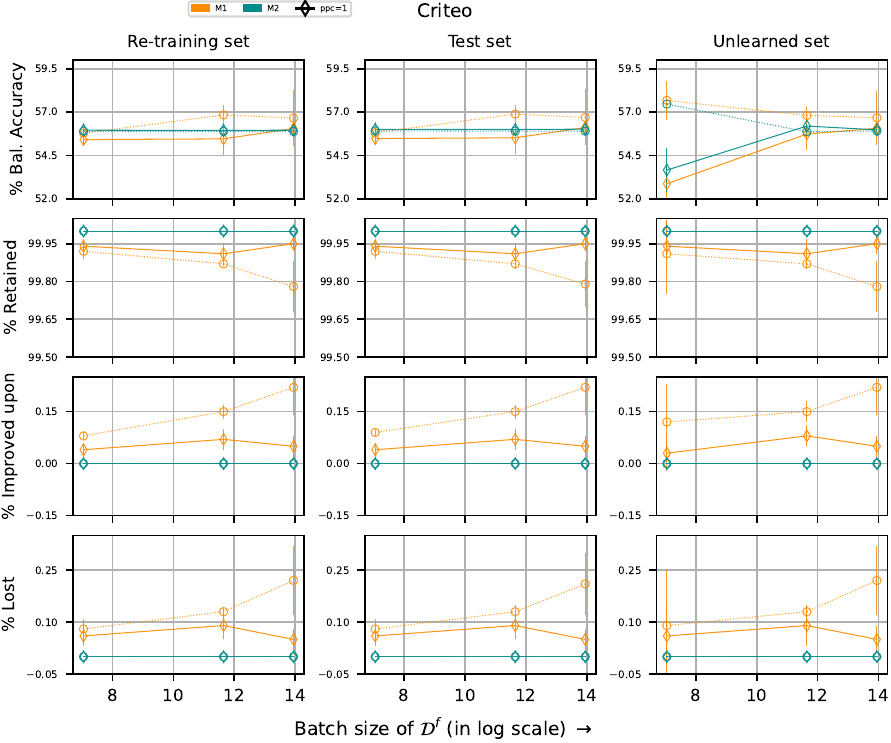}
        \caption{Criteo}
        \label{sfig:enf-criteo}
    \end{subfigure} 
    \caption{Performance of the stabilized unlearning framework for GLVQ, across the benchmark datasets.}
    \label{fig:enf-performance}
\end{figure}

\normalsize{
Within the robust unlearning framework the deteriorating impact of the types of samples discussed in \autoref{ss:robustUnlearning} and \autoref{asec:sample location} are much less severe, especially when the sample to forget is an outlier, as seen in \autoref{fig:sampleImpact_robust}. While \autoref{fig:enf-performance} illustrates the performance of the \emph{Unlearning with enforcer samples}, the corresponding tables are \autoref{tab:enf_perf_surg}, \autoref{tab:enf_perf_bank}, \autoref{tab:enf_perf_adult}, \autoref{tab:enf_perf_diab} and \autoref{tab:enf_perf_criteo}. }

\scriptsize{
\begin{longtable}{p{1.5cm}p{1.8cm}p{1.2 cm}p{1.5cm}p{1.5cm}p{1.5cm}p{1.5cm}p{1.5cm}}
    \caption{\footnotesize{Comparison of performance of the retrained and unlearned (with enforcer) models across different batch sizes of $\mathcal{D}^f$ (N($\mathcal{D}^f$)) in the `Breast cancer' dataset, reported as mean (std) over 5 iterations.}}\label{tab:enf_perf_bc}\\
    \toprule
ppc	&	N($\mathcal{D}^f$)	&	Subset & Model	& \% Bal. Acc	& \% Gained	&	\% Lost	& \% Retained	\\
\toprule
1	&	2	&	retraining	&	M1	&	95.86 (0.1)	&	0.0 (0.0)	&	0.05 (0.12)	&	99.95 (0.12)	\\	
	&		&		&	M2	&	95.93 (0.15)	&	0.0 (0.0)	&	0.0 (0.0)	&	100.0 (0.0)	\\	\cmidrule{3-8}
	&		&	test	&	M1	&	95.03 (0.0)	&	0.0 (0.0)	&	0.0 (0.0)	&	100.0 (0.0)	\\	
	&		&		&	M2	&	95.03 (0.0)	&	0.0 (0.0)	&	0.0 (0.0)	&	100.0 (0.0)	\\	\cmidrule{3-8}
	&		&	unlearned	&	M1	&	80.0 (27.39)	&	0.0 (0.0)	&	0.0 (0.0)	&	100.0 (0.0)	\\	
	&		&		&	M2	&	80.0 (27.39)	&	0.0 (0.0)	&	0.0 (0.0)	&	100.0 (0.0)	\\	\cmidrule{2-8}
	&	5	&	retraining	&	M1	&	95.78 (0.03)	&	0.0 (0.0)	&	0.0 (0.0)	&	100.0 (0.0)	\\	
	&		&		&	M2	&	95.49 (0.46)	&	0.0 (0.0)	&	0.33 (0.6)	&	99.67 (0.6)	\\	\cmidrule{3-8}
	&		&	test	&	M1	&	95.03 (0.0)	&	0.0 (0.0)	&	0.0 (0.0)	&	100.0 (0.0)	\\	
	&		&		&	M2	&	94.78 (0.55)	&	0.0 (0.0)	&	0.33 (0.75)	&	99.67 (0.75)	\\	\cmidrule{3-8}
	&		&	unlearned	&	M1	&	95.0 (11.18)	&	0.0 (0.0)	&	4.0 (8.94)	&	96.0 (8.94)	\\	
	&		&		&	M2	&	100.0 (0.0)	&	0.0 (0.0)	&	0.0 (0.0)	&	100.0 (0.0)	\\	\cmidrule{2-8}
	&	20	&	retraining	&	M1	&	95.48 (0.38)	&	0.0 (0.0)	&	0.4 (0.39)	&	99.6 (0.39)	\\	
	&		&		&	M2	&	95.83 (0.22)	&	0.06 (0.13)	&	0.11 (0.26)	&	99.89 (0.26)	\\	\cmidrule{3-8}
	&		&	test	&	M1	&	94.3 (0.74)	&	0.0 (0.0)	&	0.89 (1.01)	&	99.11 (1.01)	\\	
	&		&		&	M2	&	94.95 (0.18)	&	0.0 (0.0)	&	0.11 (0.25)	&	99.89 (0.25)	\\	\cmidrule{3-8}
	&		&	unlearned	&	M1	&	93.37 (7.19)	&	0.0 (0.0)	&	1.05 (2.35)	&	98.95 (2.35)	\\	
	&		&		&	M2	&	94.62 (6.41)	&	0.0 (0.0)	&	0.0 (0.0)	&	100.0 (0.0)	\\	\cmidrule{2-8}
	&	30	&	retraining	&	M1	&	95.35 (0.3)	&	0.0 (0.0)	&	0.41 (0.34)	&	99.59 (0.34)	\\	
	&		&		&	M2	&	95.35 (0.4)	&	0.0 (0.0)	&	0.47 (0.54)	&	99.53 (0.54)	\\	\cmidrule{3-8}
	&		&	test	&	M1	&	94.62 (0.71)	&	0.0 (0.0)	&	0.56 (0.96)	&	99.44 (0.96)	\\	
	&		&		&	M2	&	94.29 (1.25)	&	0.0 (0.0)	&	1.0 (1.69)	&	99.0 (1.69)	\\	\cmidrule{3-8}
	&		&	unlearned	&	M1	&	97.43 (1.62)	&	0.0 (0.0)	&	0.0 (0.0)	&	100.0 (0.0)	\\	
	&		&		&	M2	&	98.02 (1.95)	&	0.69 (1.54)	&	0.0 (0.0)	&	100.0 (0.0)	\\	\cmidrule{2-8}
	&	50	&	retraining	&	M1	&	95.11 (0.29)	&	0.0 (0.0)	&	0.56 (0.34)	&	99.44 (0.34)	\\	
	&		&		&	M2	&	94.94 (1.02)	&	0.19 (0.17)	&	1.25 (1.23)	&	98.75 (1.23)	\\	\cmidrule{3-8}
	&		&	test	&	M1	&	93.82 (0.94)	&	0.0 (0.0)	&	1.44 (1.08)	&	98.56 (1.08)	\\	
	&		&		&	M2	&	93.77 (1.3)	&	0.11 (0.25)	&	1.89 (2.1)	&	98.11 (2.1)	\\	\cmidrule{3-8}
	&		&	unlearned	&	M1	&	96.71 (2.42)	&	0.0 (0.0)	&	0.4 (0.89)	&	99.6 (0.89)	\\	
	&		&		&	M2	&	97.15 (2.83)	&	0.0 (0.0)	&	0.0 (0.0)	&	100.0 (0.0)	\\	\cmidrule{2-8}
	&	100	&	retraining	&	M1	&	95.02 (0.93)	&	0.0 (0.0)	&	0.99 (0.46)	&	99.01 (0.46)	\\	
	&		&		&	M2	&	95.48 (0.5)	&	0.14 (0.19)	&	0.49 (0.52)	&	99.51 (0.52)	\\	\cmidrule{3-8}
	&		&	test	&	M1	&	92.51 (0.7)	&	0.0 (0.0)	&	3.11 (0.75)	&	96.89 (0.75)	\\	
	&		&		&	M2	&	94.38 (1.32)	&	0.11 (0.25)	&	0.89 (1.69)	&	99.11 (1.69)	\\	\cmidrule{3-8}
	&		&	unlearned	&	M1	&	95.24 (2.6)	&	0.0 (0.0)	&	1.04 (1.05)	&	98.96 (1.05)	\\	
	&		&		&	M2	&	96.3 (1.76)	&	0.0 (0.0)	&	0.0 (0.0)	&	100.0 (0.0)	\\	\hline
2	&	2	&	retraining	&	M1	&	94.39 (0.12)	&	0.06 (0.12)	&	0.0 (0.0)	&	100.0 (0.0)	\\	
	&		&		&	M2	&	94.48 (0.34)	&	0.22 (0.23)	&	0.06 (0.12)	&	99.94 (0.12)	\\	\cmidrule{3-8}
	&		&	test	&	M1	&	92.47 (0.22)	&	0.0 (0.0)	&	0.23 (0.31)	&	99.77 (0.31)	\\	
	&		&		&	M2	&	92.63 (0.0)	&	0.0 (0.0)	&	0.0 (0.0)	&	100.0 (0.0)	\\	\cmidrule{3-8}
	&		&	unlearned	&	M1	&	80.0 (27.39)	&	0.0 (0.0)	&	0.0 (0.0)	&	100.0 (0.0)	\\	
	&		&		&	M2	&	80.0 (27.39)	&	0.0 (0.0)	&	0.0 (0.0)	&	100.0 (0.0)	\\	\cmidrule{2-8}
	&	5	&	retraining	&	M1	&	94.45 (0.39)	&	0.06 (0.12)	&	0.0 (0.0)	&	100.0 (0.0)	\\	
	&		&		&	M2	&	94.52 (0.28)	&	0.17 (0.25)	&	0.06 (0.13)	&	99.94 (0.13)	\\	\cmidrule{3-8}
	&		&	test	&	M1	&	92.46 (0.37)	&	0.0 (0.0)	&	0.23 (0.51)	&	99.77 (0.51)	\\	
	&		&		&	M2	&	92.3 (0.54)	&	0.0 (0.0)	&	0.46 (0.75)	&	99.54 (0.75)	\\	\cmidrule{3-8}
	&		&	unlearned	&	M1	&	90.0 (14.91)	&	0.0 (0.0)	&	0.0 (0.0)	&	100.0 (0.0)	\\	
	&		&		&	M2	&	90.0 (14.91)	&	0.0 (0.0)	&	0.0 (0.0)	&	100.0 (0.0)	\\	\cmidrule{2-8}
	&	20	&	retraining	&	M1	&	94.32 (0.41)	&	0.06 (0.13)	&	0.23 (0.24)	&	99.77 (0.24)	\\	
	&		&		&	M2	&	94.46 (0.3)	&	0.18 (0.16)	&	0.23 (0.24)	&	99.77 (0.24)	\\	\cmidrule{3-8}
	&		&	test	&	M1	&	92.05 (0.37)	&	0.0 (0.0)	&	0.8 (0.51)	&	99.2 (0.51)	\\	
	&		&		&	M2	&	92.14 (0.45)	&	0.0 (0.0)	&	0.69 (0.63)	&	99.31 (0.63)	\\	\cmidrule{3-8}
	&		&	unlearned	&	M1	&	91.03 (6.42)	&	0.0 (0.0)	&	0.0 (0.0)	&	100.0 (0.0)	\\	
	&		&		&	M2	&	91.94 (6.54)	&	1.11 (2.48)	&	0.0 (0.0)	&	100.0 (0.0)	\\	\cmidrule{2-8}
	&	30	&	retraining	&	M1	&	94.06 (0.4)	&	0.06 (0.14)	&	0.48 (0.34)	&	99.52 (0.34)	\\	
	&		&		&	M2	&	94.61 (0.64)	&	0.42 (0.27)	&	0.24 (0.25)	&	99.76 (0.25)	\\	\cmidrule{3-8}
	&		&	test	&	M1	&	91.89 (0.54)	&	0.0 (0.0)	&	1.03 (0.75)	&	98.97 (0.75)	\\	
	&		&		&	M2	&	92.05 (0.63)	&	0.0 (0.0)	&	0.8 (0.87)	&	99.2 (0.87)	\\	\cmidrule{3-8}
	&		&	unlearned	&	M1	&	94.76 (6.29)	&	0.77 (1.72)	&	0.0 (0.0)	&	100.0 (0.0)	\\	
	&		&		&	M2	&	95.26 (6.65)	&	1.46 (2.0)	&	0.0 (0.0)	&	100.0 (0.0)	\\	\cmidrule{2-8}
	&	50	&	retraining	&	M1	&	94.06 (0.56)	&	0.19 (0.17)	&	0.82 (0.65)	&	99.18 (0.65)	\\	
	&		&		&	M2	&	95.13 (0.27)	&	0.82 (0.17)	&	0.13 (0.17)	&	99.87 (0.17)	\\	\cmidrule{3-8}
	&		&	test	&	M1	&	91.72 (0.45)	&	0.0 (0.0)	&	1.26 (0.63)	&	98.74 (0.63)	\\	
	&		&		&	M2	&	92.47 (0.91)	&	0.23 (0.51)	&	0.46 (1.02)	&	99.54 (1.02)	\\	\cmidrule{3-8}
	&		&	unlearned	&	M1	&	91.87 (2.45)	&	0.0 (0.0)	&	0.84 (1.15)	&	99.16 (1.15)	\\	
	&		&		&	M2	&	92.19 (3.24)	&	0.0 (0.0)	&	0.43 (0.95)	&	99.57 (0.95)	\\	\cmidrule{2-8}
	&	100	&	retraining	&	M1	&	93.34 (0.7)	&	0.14 (0.2)	&	1.37 (0.7)	&	98.63 (0.7)	\\	
	&		&		&	M2	&	94.47 (0.58)	&	0.72 (0.51)	&	0.5 (0.93)	&	99.5 (0.93)	\\	\cmidrule{3-8}
	&		&	test	&	M1	&	91.64 (0.37)	&	0.0 (0.0)	&	1.37 (0.51)	&	98.63 (0.51)	\\	
	&		&		&	M2	&	92.85 (0.89)	&	0.8 (0.96)	&	0.69 (1.24)	&	99.31 (1.24)	\\	\cmidrule{3-8}
	&		&	unlearned	&	M1	&	94.22 (1.47)	&	0.0 (0.0)	&	0.84 (0.47)	&	99.16 (0.47)	\\	
	&		&		&	M2	&	95.2 (1.78)	&	0.63 (0.57)	&	0.21 (0.47)	&	99.79 (0.47)	\\	\bottomrule

\end{longtable}

\begin{longtable}{p{1.5cm}p{1.8cm}p{1.2 cm}p{1.5cm}p{1.8cm}p{1.5cm}p{1.8cm}p{1.5cm}}
    \caption{\footnotesize{Comparison of performance of the retrained and unlearned (with enforcer) models across different batch sizes of $\mathcal{D}^f$ (N($\mathcal{D}^f$)) in the `Surgical' dataset, reported as mean (std) over 5 iterations.}}\label{tab:enf_perf_surg}\\
    \toprule
    ppc	&	N($\mathcal{D}^f$)	&	Subset & Model	& \% Bal. Acc	& \% Gained	&	\% Lost	& \% Retained	\\
    \toprule
    1	&	2	&	retraining	&	M1	&	67.93 (0.06)	&	0.23 (0.05)	&	0.19 (0.03)	&	99.81 (0.03)	\\	
	&		&	      	&	M2	&	67.94 (0.01)	&	0.01 (0.02)	&	0.01 (0.01)	&	99.99 (0.01)	\\	\cmidrule{3-8}
	&		&	test	&	M1	&	67.62 (0.18)	&	0.29 (0.1)	&	0.16 (0.14)	&	99.84 (0.14)	\\	
	&		&	      &	M2	&	67.55 (0.02)	&	0.02 (0.02)	&	0.03 (0.02)	&	99.97 (0.02)	\\	\cmidrule{3-8}
	&		&	unlearned	&	M1	&	80.0 (27.39)	&	0.0 (0.0)	&	0.0 (0.0)	&	100.0 (0.0)	\\	
	&		&	      &	M2	&	80.0 (27.39)	&	0.0 (0.0)	&	0.0 (0.0)	&	100.0 (0.0)	\\	\cmidrule{2-8}
	&	12	&	retraining	&	M1	&	67.92 (0.03)	&	0.18 (0.05)	&	0.16 (0.02)	&	99.84 (0.02)	\\	
	&		&	      &	M2	&	67.92 (0.02)	&	0.03 (0.02)	&	0.06 (0.06)	&	99.94 (0.06)	\\	\cmidrule{3-8}
	&		&	test	&	M1	&	67.56 (0.12)	&	0.23 (0.08)	&	0.18 (0.06)	&	99.82 (0.06)	\\	
	&		&	      &	M2	&	67.57 (0.1)	&	0.04 (0.04)	&	0.06 (0.07)	&	99.94 (0.07)	\\	\cmidrule{3-8}
	&		&	unlearned	&	M1	&	76.64 (11.91)	&	0.0 (0.0)	&	0.0 (0.0)	&	100.0 (0.0)	\\	
	&		&	      &	M2	&	76.64 (11.91)	&	0.0 (0.0)	&	0.0 (0.0)	&	100.0 (0.0)	\\	\cmidrule{2-8}
	&	118	&	retraining	&	M1	&	67.95 (0.02)	&	0.25 (0.08)	&	0.21 (0.06)	&	99.79 (0.06)	\\	
	&		&	      &	M2	&	67.96 (0.11)	&	0.17 (0.06)	&	0.31 (0.09)	&	99.69 (0.09)	\\	\cmidrule{3-8}
	&		&	test	&	M1	&	67.69 (0.17)	&	0.37 (0.16)	&	0.17 (0.06)	&	99.83 (0.06)	\\	
	&		&	      &	M2	&	67.73 (0.11)	&	0.26 (0.08)	&	0.26 (0.17)	&	99.74 (0.17)	\\	\cmidrule{3-8}
	&		&	unlearned	&	M1	&	64.67 (3.37)	&	0.0 (0.0)	&	0.46 (0.63)	&	99.54 (0.63)	\\	
	&		&	      &	M2	&	65.38 (2.61)	&	0.23 (0.51)	&	0.0 (0.0)	&	100.0 (0.0)	\\	\cmidrule{2-8}
	&	1171	&	retraining	&	M1	&	68.08 (0.09)	&	0.38 (0.1)	&	0.52 (0.17)	&	99.48 (0.17)	\\	
	&		&	      &	M2	&	67.95 (0.26)	&	0.66 (0.11)	&	0.87 (0.29)	&	99.13 (0.29)	\\	\cmidrule{3-8}
	&		&	test	&	M1	&	67.52 (0.15)	&	0.29 (0.08)	&	0.63 (0.26)	&	99.37 (0.26)	\\	
	&		&	      &	M2	&	67.73 (0.41)	&	0.75 (0.31)	&	0.87 (0.45)	&	99.13 (0.45)	\\	\cmidrule{3-8}
	&		&	unlearned	&	M1	&	67.72 (1.93)	&	0.33 (0.13)	&	0.72 (0.52)	&	99.28 (0.52)	\\	
	&		&	      &	M2	&	68.23 (2.12)	&	0.83 (0.12)	&	0.77 (0.71)	&	99.23 (0.71)	\\	\hline
2	&	2	&	retraining	&	M1	&	66.77 (0.23)	&	1.19 (1.58)	&	0.96 (1.21)	&	99.04 (1.21)	\\	
	&		&	      &	M2	&	66.65 (0.01)	&	0.01 (0.01)	&	0.01 (0.01)	&	99.99 (0.01)	\\	\cmidrule{3-8}
	&		&	test	&	M1	&	66.33 (0.71)	&	0.88 (1.17)	&	1.23 (1.64)	&	98.77 (1.64)	\\	
	&		&	      &	M2	&	66.84 (0.04)	&	0.01 (0.02)	&	0.01 (0.02)	&	99.99 (0.02)	\\	\cmidrule{3-8}
	&		&	unlearned	&	M1	&	70.0 (27.39)	&	0.0 (0.0)	&	0.0 (0.0)	&	100.0 (0.0)	\\	
	&		&	      &	M2	&	70.0 (27.39)	&	0.0 (0.0)	&	0.0 (0.0)	&	100.0 (0.0)	\\	\cmidrule{2-8}
	&	12	&	retraining	&	M1	&	66.75 (0.19)	&	1.18 (1.56)	&	0.97 (1.2)	&	99.03 (1.2)	\\	
	&		&	      &	M2	&	66.64 (0.04)	&	0.02 (0.01)	&	0.03 (0.02)	&	99.97 (0.02)	\\	\cmidrule{3-8}
	&		&	test	&	M1	&	66.34 (0.7)	&	0.9 (1.19)	&	1.24 (1.67)	&	98.76 (1.67)	\\	
	&		&	      &	M2	&	66.87 (0.03)	&	0.03 (0.02)	&	0.02 (0.02)	&	99.98 (0.02)	\\	\cmidrule{3-8}
	&		&	unlearned	&	M1	&	69.86 (24.06)	&	1.82 (4.07)	&	2.22 (4.97)	&	97.78 (4.97)	\\	
	&		&	      &	M2	&	69.97 (21.86)	&	0.0 (0.0)	&	0.0 (0.0)	&	100.0 (0.0)	\\	\cmidrule{2-8}
	&	118	&	retraining	&	M1	&	66.67 (0.09)	&	0.63 (1.23)	&	0.59 (0.97)	&	99.41 (0.97)	\\	
	&		&	      &	M2	&	66.63 (0.09)	&	0.09 (0.02)	&	0.04 (0.03)	&	99.96 (0.03)	\\	\cmidrule{3-8}
	&		&	test	&	M1	&	66.6 (0.58)	&	0.45 (0.84)	&	0.65 (1.27)	&	99.35 (1.27)	\\	
	&		&	      &	M2	&	66.85 (0.04)	&	0.03 (0.02)	&	0.02 (0.02)	&	99.98 (0.02)	\\	\cmidrule{3-8}
	&		&	unlearned	&	M1	&	69.76 (6.63)	&	0.81 (1.81)	&	0.84 (0.86)	&	99.16 (0.86)	\\	
	&		&	      &	M2	&	69.66 (6.15)	&	0.0 (0.0)	&	0.22 (0.49)	&	99.78 (0.49)	\\	\cmidrule{2-8}
	&	1171	&	retraining	&	M1	&	67.12 (0.39)	&	1.89 (1.28)	&	1.83 (1.07)	&	98.17 (1.07)	\\	
	&		&	          &	M2	&	66.74 (0.16)	&	0.31 (0.23)	&	0.22 (0.1)	&	99.78 (0.1)	\\	\cmidrule{3-8}
	&		&	test	&	M1	&	66.23 (0.77)	&	1.3 (0.85)	&	2.26 (1.49)	&	97.74 (1.49)	\\	
	&		&	      &	M2	&	66.84 (0.18)	&	0.18 (0.17)	&	0.23 (0.25)	&	99.77 (0.25)	\\	\cmidrule{3-8}
	&		&	unlearned	&	M1	&	66.71 (2.35)	&	1.67 (1.33)	&	1.89 (1.46)	&	98.11 (1.46)	\\	
	&		&	          &	M2	&	66.71 (2.39)	&	0.29 (0.3)	&	0.15 (0.19)	&	99.85 (0.19)	\\	\bottomrule
\end{longtable}

\begin{longtable}{p{1.5cm}p{1.8cm}p{1.2 cm}p{1.5cm}p{1.8cm}p{1.5cm}p{1.8cm}p{1.5cm}}
    \caption{\footnotesize{Comparison of performance of the retrained and unlearned (with enforcer) models across different batch sizes of $\mathcal{D}^f$ (N($\mathcal{D}^f$)) in the `Banking' dataset, reported as mean (std) over 5 iterations.}}\label{tab:enf_perf_bank}\\
    \toprule
    ppc	&	N($\mathcal{D}^f$)	&	Subset & Model	& \% Bal. Acc	& \% Gained	&	\% Lost	& \% Retained	\\
    \toprule
1	&	2	&	retraining	&	M1	&	67.93 (0.06)	&	0.23 (0.05)	&	0.19 (0.03)	&	99.81 (0.03)	\\	
	&		&		&	M2	&	67.94 (0.01)	&	0.01 (0.02)	&	0.01 (0.01)	&	99.99 (0.01)	\\	\cmidrule{3-8}
	&		&	test	&	M1	&	67.62 (0.18)	&	0.29 (0.1)	&	0.16 (0.14)	&	99.84 (0.14)	\\	
	&		&		&	M2	&	67.55 (0.02)	&	0.02 (0.02)	&	0.03 (0.02)	&	99.97 (0.02)	\\	\cmidrule{3-8}
	&		&	unlearned	&	M1	&	80.0 (27.39)	&	0.0 (0.0)	&	0.0 (0.0)	&	100.0 (0.0)	\\	
	&		&		&	M2	&	80.0 (27.39)	&	0.0 (0.0)	&	0.0 (0.0)	&	100.0 (0.0)	\\	\cmidrule{2-8}
	&	12	&	retraining	&	M1	&	67.92 (0.03)	&	0.18 (0.05)	&	0.16 (0.02)	&	99.84 (0.02)	\\	
	&		&		&	M2	&	67.92 (0.02)	&	0.03 (0.02)	&	0.06 (0.06)	&	99.94 (0.06)	\\	\cmidrule{3-8}
	&		&	test	&	M1	&	67.56 (0.12)	&	0.23 (0.08)	&	0.18 (0.06)	&	99.82 (0.06)	\\	
	&		&		&	M2	&	67.57 (0.1)	&	0.04 (0.04)	&	0.06 (0.07)	&	99.94 (0.07)	\\	\cmidrule{3-8}
	&		&	unlearned	&	M1	&	76.64 (11.91)	&	0.0 (0.0)	&	0.0 (0.0)	&	100.0 (0.0)	\\	
	&		&		&	M2	&	76.64 (11.91)	&	0.0 (0.0)	&	0.0 (0.0)	&	100.0 (0.0)	\\	\cmidrule{2-8}
	&	118	&	retraining	&	M1	&	67.95 (0.02)	&	0.25 (0.08)	&	0.21 (0.06)	&	99.79 (0.06)	\\	
	&		&		&	M2	&	67.96 (0.11)	&	0.17 (0.06)	&	0.31 (0.09)	&	99.69 (0.09)	\\	\cmidrule{3-8}
	&		&	test	&	M1	&	67.69 (0.17)	&	0.37 (0.16)	&	0.17 (0.06)	&	99.83 (0.06)	\\	
	&		&		&	M2	&	67.73 (0.11)	&	0.26 (0.08)	&	0.26 (0.17)	&	99.74 (0.17)	\\	\cmidrule{3-8}
	&		&	unlearned	&	M1	&	64.67 (3.37)	&	0.0 (0.0)	&	0.46 (0.63)	&	99.54 (0.63)	\\	
	&		&		&	M2	&	65.38 (2.61)	&	0.23 (0.51)	&	0.0 (0.0)	&	100.0 (0.0)	\\	\cmidrule{2-8}
	&	1171	&	retraining	&	M1	&	68.08 (0.09)	&	0.38 (0.1)	&	0.52 (0.17)	&	99.48 (0.17)	\\	
	&		&		&	M2	&	67.95 (0.26)	&	0.66 (0.11)	&	0.87 (0.29)	&	99.13 (0.29)	\\	\cmidrule{3-8}
	&		&	test	&	M1	&	67.52 (0.15)	&	0.29 (0.08)	&	0.63 (0.26)	&	99.37 (0.26)	\\	
	&		&		&	M2	&	67.73 (0.41)	&	0.75 (0.31)	&	0.87 (0.45)	&	99.13 (0.45)	\\	\cmidrule{3-8}
	&		&	unlearned	&	M1	&	67.72 (1.93)	&	0.33 (0.13)	&	0.72 (0.52)	&	99.28 (0.52)	\\	
	&		&		&	M2	&	68.23 (2.12)	&	0.83 (0.12)	&	0.77 (0.71)	&	99.23 (0.71)	\\	\hline
2	&	2	&	retraining	&	M1	&	66.77 (0.23)	&	1.19 (1.58)	&	0.96 (1.21)	&	99.04 (1.21)	\\	
	&		&		&	M2	&	66.65 (0.01)	&	0.01 (0.01)	&	0.01 (0.01)	&	99.99 (0.01)	\\	\cmidrule{3-8}
	&		&	test	&	M1	&	66.33 (0.71)	&	0.88 (1.17)	&	1.23 (1.64)	&	98.77 (1.64)	\\	
	&		&		&	M2	&	66.84 (0.04)	&	0.01 (0.02)	&	0.01 (0.02)	&	99.99 (0.02)	\\	\cmidrule{3-8}
	&		&	unlearned	&	M1	&	70.0 (27.39)	&	0.0 (0.0)	&	0.0 (0.0)	&	100.0 (0.0)	\\	
	&		&		&	M2	&	70.0 (27.39)	&	0.0 (0.0)	&	0.0 (0.0)	&	100.0 (0.0)	\\	\cmidrule{2-8}
	&	12	&	retraining	&	M1	&	66.75 (0.19)	&	1.18 (1.56)	&	0.97 (1.2)	&	99.03 (1.2)	\\	
	&		&		&	M2	&	66.64 (0.04)	&	0.02 (0.01)	&	0.03 (0.02)	&	99.97 (0.02)	\\	\cmidrule{3-8}
	&		&	test	&	M1	&	66.34 (0.7)	&	0.9 (1.19)	&	1.24 (1.67)	&	98.76 (1.67)	\\	
	&		&		&	M2	&	66.87 (0.03)	&	0.03 (0.02)	&	0.02 (0.02)	&	99.98 (0.02)	\\	\cmidrule{3-8}
	&		&	unlearned	&	M1	&	69.86 (24.06)	&	1.82 (4.07)	&	2.22 (4.97)	&	97.78 (4.97)	\\	
	&		&		&	M2	&	69.97 (21.86)	&	0.0 (0.0)	&	0.0 (0.0)	&	100.0 (0.0)	\\	\cmidrule{2-8}
	&	118	&	retraining	&	M1	&	66.67 (0.09)	&	0.63 (1.23)	&	0.59 (0.97)	&	99.41 (0.97)	\\	
	&		&		&	M2	&	66.63 (0.09)	&	0.09 (0.02)	&	0.04 (0.03)	&	99.96 (0.03)	\\	\cmidrule{3-8}
	&		&	test	&	M1	&	66.6 (0.58)	&	0.45 (0.84)	&	0.65 (1.27)	&	99.35 (1.27)	\\	
	&		&		&	M2	&	66.85 (0.04)	&	0.03 (0.02)	&	0.02 (0.02)	&	99.98 (0.02)	\\	\cmidrule{3-8}
	&		&	unlearned	&	M1	&	69.76 (6.63)	&	0.81 (1.81)	&	0.84 (0.86)	&	99.16 (0.86)	\\	
	&		&		&	M2	&	69.66 (6.15)	&	0.0 (0.0)	&	0.22 (0.49)	&	99.78 (0.49)	\\	\cmidrule{2-8}
	&	1171	&	retraining	&	M1	&	67.12 (0.39)	&	1.89 (1.28)	&	1.83 (1.07)	&	98.17 (1.07)	\\	
	&		&		&	M2	&	66.74 (0.16)	&	0.31 (0.23)	&	0.22 (0.1)	&	99.78 (0.1)	\\	\cmidrule{3-8}
	&		&	test	&	M1	&	66.23 (0.77)	&	1.3 (0.85)	&	2.26 (1.49)	&	97.74 (1.49)	\\	
	&		&		&	M2	&	66.84 (0.18)	&	0.18 (0.17)	&	0.23 (0.25)	&	99.77 (0.25)	\\	\cmidrule{3-8}
	&		&	unlearned	&	M1	&	66.71 (2.35)	&	1.67 (1.33)	&	1.89 (1.46)	&	98.11 (1.46)	\\	
	&		&		&	M2	&	66.71 (2.39)	&	0.29 (0.3)	&	0.15 (0.19)	&	99.85 (0.19)	\\	\bottomrule
\end{longtable}

\begin{longtable}{p{1.5cm}p{1.8cm}p{1.2 cm}p{1.32cm}p{1.6cm}p{1.5cm}p{1.5cm}p{1.5cm}}
    \caption{\footnotesize{Comparison of performance of the retrained and unlearned (with enforcer) models across different batch sizes of $\mathcal{D}^f$ (N($\mathcal{D}^f$)) in the `Adult' dataset, reported as mean (std) over 5 iterations.}}\label{tab:enf_perf_adult}\\
    \toprule
    ppc	&	N($\mathcal{D}^f$)	&	Subset & Model	& \% Bal. Acc	& \% Gained	&	\% Lost	& \% Retained	\\
    \toprule
1	&	4	&	retraining	&	M1	&	65.37 (0.69)	&	0.67 (0.14)	&	0.9 (0.12)	&	99.1 (0.12)	\\	
	&		&		&	M2	&	66.16 (2.68)	&	2.3 (1.56)	&	3.46 (3.1)	&	96.54 (3.1)	\\	\cmidrule{3-8}
	&		&	test	&	M1	&	64.79 (0.46)	&	0.73 (0.14)	&	0.86 (0.15)	&	99.14 (0.15)	\\	
	&		&		&	M2	&	65.52 (2.4)	&	2.29 (1.62)	&	3.49 (3.04)	&	96.51 (3.04)	\\	\cmidrule{3-8}
	&		&	unlearned	&	M1	&	66.67 (23.57)	&	0.0 (0.0)	&	0.0 (0.0)	&	100.0 (0.0)	\\	
	&		&		&	M2	&	66.67 (23.57)	&	0.0 (0.0)	&	0.0 (0.0)	&	100.0 (0.0)	\\	\cmidrule{2-8}
	&	33	&	retraining	&	M1	&	65.43 (0.76)	&	0.56 (0.25)	&	0.77 (0.44)	&	99.23 (0.44)	\\	
	&		&		&	M2	&	66.08 (1.15)	&	0.83 (0.5)	&	1.03 (0.52)	&	98.97 (0.52)	\\	\cmidrule{3-8}
	&		&	test	&	M1	&	64.83 (0.54)	&	0.63 (0.31)	&	0.75 (0.31)	&	99.25 (0.31)	\\	
	&		&		&	M2	&	65.33 (0.85)	&	0.81 (0.39)	&	1.09 (0.65)	&	98.91 (0.65)	\\	\cmidrule{3-8}
	&		&	unlearned	&	M1	&	63.72 (13.54)	&	0.8 (1.79)	&	0.74 (1.66)	&	99.26 (1.66)	\\	
	&		&		&	M2	&	63.2 (13.44)	&	0.0 (0.0)	&	1.72 (2.45)	&	98.28 (2.45)	\\	\cmidrule{2-8}
	&	326	&	retraining	&	M1	&	65.74 (0.99)	&	0.69 (0.33)	&	0.92 (0.43)	&	99.08 (0.43)	\\	
	&		&		&	M2	&	66.09 (0.17)	&	0.14 (0.08)	&	0.18 (0.09)	&	99.82 (0.09)	\\	\cmidrule{3-8}
	&		&	test	&	M1	&	65.05 (0.71)	&	0.71 (0.31)	&	0.94 (0.38)	&	99.06 (0.38)	\\	
	&		&		&	M2	&	65.38 (0.16)	&	0.13 (0.07)	&	0.23 (0.12)	&	99.77 (0.12)	\\	\cmidrule{3-8}
	&		&	unlearned	&	M1	&	66.42 (2.71)	&	0.7 (0.34)	&	0.85 (0.74)	&	99.15 (0.74)	\\	
	&		&		&	M2	&	66.72 (3.23)	&	0.16 (0.35)	&	0.15 (0.21)	&	99.85 (0.21)	\\	\cmidrule{2-8}
	&	3257	&	retraining	&	M1	&	65.6 (0.94)	&	0.77 (0.37)	&	1.1 (0.44)	&	98.9 (0.44)	\\	
	&		&		&	M2	&	65.83 (0.33)	&	0.34 (0.11)	&	0.45 (0.17)	&	99.55 (0.17)	\\	\cmidrule{3-8}
	&		&	test	&	M1	&	64.94 (0.74)	&	0.82 (0.36)	&	1.09 (0.43)	&	98.91 (0.43)	\\	
	&		&		&	M2	&	65.22 (0.17)	&	0.43 (0.1)	&	0.45 (0.15)	&	99.55 (0.15)	\\	\cmidrule{3-8}
	&		&	unlearned	&	M1	&	65.18 (1.77)	&	0.91 (0.47)	&	1.09 (0.67)	&	98.91 (0.67)	\\	
	&		&		&	M2	&	65.5 (0.93)	&	0.43 (0.18)	&	0.33 (0.17)	&	99.67 (0.17)	\\	\hline
2	&	4	&	retraining	&	M1	&	65.67 (1.15)	&	1.06 (1.04)	&	1.39 (1.46)	&	98.61 (1.46)	\\	
	&		&		&	M2	&	68.05 (3.02)	&	2.68 (1.95)	&	4.03 (3.25)	&	95.97 (3.25)	\\	\cmidrule{3-8}
	&		&	test	&	M1	&	65.02 (0.92)	&	1.17 (1.15)	&	1.29 (1.38)	&	98.71 (1.38)	\\	
	&		&		&	M2	&	67.46 (2.97)	&	2.76 (1.96)	&	3.9 (3.19)	&	96.1 (3.19)	\\	\cmidrule{3-8}
	&		&	unlearned	&	M1	&	75.0 (35.36)	&	0.0 (0.0)	&	0.0 (0.0)	&	100.0 (0.0)	\\	
	&		&		&	M2	&	76.67 (25.28)	&	20.0 (44.72)	&	5.0 (11.18)	&	95.0 (11.18)	\\	\cmidrule{2-8}
	&	33	&	retraining	&	M1	&	65.96 (1.28)	&	1.33 (0.72)	&	1.71 (0.94)	&	98.29 (0.94)	\\	
	&		&		&	M2	&	67.93 (1.76)	&	1.68 (1.16)	&	2.04 (1.46)	&	97.96 (1.46)	\\	\cmidrule{3-8}
	&		&	test	&	M1	&	65.34 (1.11)	&	1.44 (0.76)	&	1.62 (0.9)	&	98.38 (0.9)	\\	
	&		&		&	M2	&	67.26 (1.83)	&	1.76 (1.19)	&	2.01 (1.46)	&	97.99 (1.46)	\\	\cmidrule{3-8}
	&		&	unlearned	&	M1	&	68.09 (12.75)	&	0.69 (1.54)	&	2.12 (3.18)	&	97.88 (3.18)	\\	
	&		&		&	M2	&	67.34 (13.41)	&	0.0 (0.0)	&	1.43 (1.96)	&	98.57 (1.96)	\\	\cmidrule{2-8}
	&	326	&	retraining	&	M1	&	65.54 (1.9)	&	1.21 (0.42)	&	1.7 (0.76)	&	98.3 (0.76)	\\	
	&		&		&	M2	&	66.89 (0.52)	&	0.37 (0.2)	&	0.45 (0.23)	&	99.55 (0.23)	\\	\cmidrule{3-8}
	&		&	test	&	M1	&	64.95 (1.62)	&	1.35 (0.51)	&	1.57 (0.6)	&	98.43 (0.6)	\\	
	&		&		&	M2	&	66.05 (0.43)	&	0.39 (0.24)	&	0.44 (0.28)	&	99.56 (0.28)	\\	\cmidrule{3-8}
	&		&	unlearned	&	M1	&	65.91 (3.87)	&	1.14 (0.44)	&	1.52 (0.57)	&	98.48 (0.57)	\\	
	&		&		&	M2	&	66.66 (3.63)	&	0.46 (0.33)	&	0.6 (0.57)	&	99.4 (0.57)	\\	\cmidrule{2-8}
	&	3257	&	retraining	&	M1	&	65.68 (0.72)	&	0.85 (0.89)	&	1.2 (1.2)	&	98.8 (1.2)	\\	
	&		&		&	M2	&	66.41 (0.65)	&	0.38 (0.28)	&	0.47 (0.39)	&	99.53 (0.39)	\\	\cmidrule{3-8}
	&		&	test	&	M1	&	65.1 (0.59)	&	0.96 (0.99)	&	1.05 (1.17)	&	98.95 (1.17)	\\	
	&		&		&	M2	&	65.64 (0.54)	&	0.42 (0.33)	&	0.43 (0.36)	&	99.57 (0.36)	\\	\cmidrule{3-8}
	&		&	unlearned	&	M1	&	65.98 (1.54)	&	0.92 (1.03)	&	1.18 (1.46)	&	98.82 (1.46)	\\	
	&		&		&	M2	&	66.6 (1.58)	&	0.4 (0.21)	&	0.5 (0.41)	&	99.5 (0.41)	\\	\bottomrule
\end{longtable}
\begin{longtable}{p{1.5cm}p{1.8cm}p{1.2 cm}p{1.32cm}p{1.6cm}p{1.5cm}p{1.5cm}p{1.5cm}}
    \caption{\footnotesize{Comparison of performance of the retrained and unlearned (with enforcer) models across different batch sizes of $\mathcal{D}^f$ (N($\mathcal{D}^f$)) in the `Diabetes' dataset, reported as mean (std) over 5 iterations.}}\label{tab:enf_perf_diab}\\
    \toprule
    ppc	&	N($\mathcal{D}^f$)	&	Subset & Model	& \% Bal. Acc	& \% Gained	&	\% Lost	& \% Retained	\\
    \toprule
1	&	9	&	retraining	&	M1	&	61.68 (0.08)	&	1.44 (0.09)	&	1.33 (0.19)	&	98.67 (0.19)	\\	
	&		&		&	M2	&	61.59 (0.0)	&	0.01 (0.0)	&	0.01 (0.01)	&	99.99 (0.01)	\\	\cmidrule{3-8}
	&		&	test	&	M1	&	61.73 (0.05)	&	1.42 (0.1)	&	1.26 (0.16)	&	98.74 (0.16)	\\	
	&		&		&	M2	&	61.62 (0.01)	&	0.02 (0.01)	&	0.01 (0.01)	&	99.99 (0.01)	\\	\cmidrule{3-8}
	&		&	unlearned	&	M1	&	62.12 (19.14)	&	0.0 (0.0)	&	0.0 (0.0)	&	100.0 (0.0)	\\	
	&		&		&	M2	&	62.12 (19.14)	&	0.0 (0.0)	&	0.0 (0.0)	&	100.0 (0.0)	\\	\cmidrule{2-8}
	&	82	&	retraining	&	M1	&	61.68 (0.04)	&	1.13 (0.34)	&	1.02 (0.29)	&	98.98 (0.29)	\\	
	&		&		&	M2	&	61.59 (0.01)	&	0.02 (0.01)	&	0.03 (0.01)	&	99.97 (0.01)	\\	\cmidrule{3-8}
	&		&	test	&	M1	&	61.71 (0.09)	&	1.09 (0.31)	&	0.98 (0.27)	&	99.02 (0.27)	\\	
	&		&		&	M2	&	61.62 (0.01)	&	0.05 (0.01)	&	0.04 (0.02)	&	99.96 (0.02)	\\	\cmidrule{3-8}
	&		&	unlearned	&	M1	&	62.05 (3.88)	&	1.59 (2.17)	&	1.57 (1.64)	&	98.43 (1.64)	\\	
	&		&		&	M2	&	61.78 (4.57)	&	0.0 (0.0)	&	0.0 (0.0)	&	100.0 (0.0)	\\	\cmidrule{2-8}
	&	815	&	retraining	&	M1	&	61.65 (0.04)	&	1.08 (0.14)	&	1.01 (0.17)	&	98.99 (0.17)	\\	
	&		&		&	M2	&	61.58 (0.02)	&	0.09 (0.04)	&	0.1 (0.03)	&	99.9 (0.03)	\\	\cmidrule{3-8}
	&		&	test	&	M1	&	61.69 (0.05)	&	1.05 (0.13)	&	0.98 (0.14)	&	99.02 (0.14)	\\	
	&		&		&	M2	&	61.63 (0.03)	&	0.12 (0.04)	&	0.1 (0.04)	&	99.9 (0.04)	\\	\cmidrule{3-8}
	&		&	unlearned	&	M1	&	61.51 (1.53)	&	0.63 (0.35)	&	0.97 (0.62)	&	99.03 (0.62)	\\	
	&		&		&	M2	&	61.7 (1.83)	&	0.12 (0.11)	&	0.15 (0.16)	&	99.85 (0.16)	\\	\cmidrule{2-8}
	&	8142	&	retraining	&	M1	&	61.64 (0.08)	&	1.23 (0.22)	&	1.18 (0.21)	&	98.82 (0.21)	\\	
	&		&		&	M2	&	61.6 (0.04)	&	0.33 (0.07)	&	0.32 (0.07)	&	99.68 (0.07)	\\	\cmidrule{3-8}
	&		&	test	&	M1	&	61.69 (0.08)	&	1.23 (0.18)	&	1.14 (0.24)	&	98.86 (0.24)	\\	
	&		&		&	M2	&	61.67 (0.04)	&	0.37 (0.09)	&	0.28 (0.05)	&	99.72 (0.05)	\\	\cmidrule{3-8}
	&		&	unlearned	&	M1	&	61.67 (0.26)	&	1.28 (0.23)	&	1.13 (0.29)	&	98.87 (0.29)	\\	
	&		&		&	M2	&	61.59 (0.24)	&	0.35 (0.09)	&	0.32 (0.1)	&	99.68 (0.1)	\\	\hline
2	&	9	&	retraining	&	M1	&	61.85 (0.03)	&	0.95 (0.24)	&	0.85 (0.23)	&	99.15 (0.23)	\\	
	&		&		&	M2	&	61.77 (0.0)	&	0.01 (0.01)	&	0.01 (0.01)	&	99.99 (0.01)	\\	\cmidrule{3-8}
	&		&	test	&	M1	&	61.83 (0.06)	&	0.93 (0.26)	&	0.78 (0.21)	&	99.22 (0.21)	\\	
	&		&		&	M2	&	61.72 (0.01)	&	0.01 (0.01)	&	0.01 (0.01)	&	99.99 (0.01)	\\	\cmidrule{3-8}
	&		&	unlearned	&	M1	&	56.83 (6.14)	&	5.0 (11.18)	&	0.0 (0.0)	&	100.0 (0.0)	\\	
	&		&		&	M2	&	54.83 (8.17)	&	0.0 (0.0)	&	0.0 (0.0)	&	100.0 (0.0)	\\	\cmidrule{2-8}
	&	82	&	retraining	&	M1	&	61.83 (0.02)	&	1.0 (0.2)	&	0.93 (0.21)	&	99.07 (0.21)	\\	
	&		&		&	M2	&	61.77 (0.01)	&	0.02 (0.01)	&	0.02 (0.01)	&	99.98 (0.01)	\\	\cmidrule{3-8}
	&		&	test	&	M1	&	61.81 (0.06)	&	0.99 (0.25)	&	0.87 (0.2)	&	99.13 (0.2)	\\	
	&		&		&	M2	&	61.73 (0.01)	&	0.03 (0.01)	&	0.03 (0.02)	&	99.97 (0.02)	\\	\cmidrule{3-8}
	&		&	unlearned	&	M1	&	64.99 (6.52)	&	0.35 (0.78)	&	0.0 (0.0)	&	100.0 (0.0)	\\	
	&		&		&	M2	&	64.74 (6.28)	&	0.0 (0.0)	&	0.0 (0.0)	&	100.0 (0.0)	\\	\cmidrule{2-8}
	&	815	&	retraining	&	M1	&	61.79 (0.02)	&	0.83 (0.24)	&	0.79 (0.25)	&	99.21 (0.25)	\\	
	&		&		&	M2	&	61.79 (0.03)	&	0.09 (0.04)	&	0.08 (0.02)	&	99.92 (0.02)	\\	\cmidrule{3-8}
	&		&	test	&	M1	&	61.77 (0.05)	&	0.79 (0.23)	&	0.72 (0.29)	&	99.28 (0.29)	\\	
	&		&		&	M2	&	61.73 (0.02)	&	0.11 (0.04)	&	0.1 (0.01)	&	99.9 (0.01)	\\	\cmidrule{3-8}
	&		&	unlearned	&	M1	&	61.29 (1.76)	&	0.91 (0.46)	&	0.91 (0.32)	&	99.09 (0.32)	\\	
	&		&		&	M2	&	61.17 (1.78)	&	0.08 (0.11)	&	0.24 (0.22)	&	99.76 (0.22)	\\	\cmidrule{2-8}
	&	8142	&	retraining	&	M1	&	61.83 (0.08)	&	1.25 (0.74)	&	1.19 (0.63)	&	98.81 (0.63)	\\	
	&		&		&	M2	&	61.8 (0.06)	&	0.35 (0.11)	&	0.35 (0.09)	&	99.65 (0.09)	\\	\cmidrule{3-8}
	&		&	test	&	M1	&	61.75 (0.05)	&	1.21 (0.71)	&	1.16 (0.64)	&	98.84 (0.64)	\\	
	&		&		&	M2	&	61.74 (0.04)	&	0.32 (0.09)	&	0.33 (0.09)	&	99.67 (0.09)	\\	\cmidrule{3-8}
	&		&	unlearned	&	M1	&	61.74 (0.59)	&	1.34 (0.68)	&	1.11 (0.51)	&	98.89 (0.51)	\\	
	&		&		&	M2	&	61.55 (0.56)	&	0.31 (0.08)	&	0.41 (0.05)	&	99.59 (0.05)	\\	\bottomrule

\end{longtable}

\begin{longtable}{p{1.5cm}p{1.8cm}p{1.2 cm}p{1.32cm}p{1.6cm}p{1.5cm}p{1.5cm}p{1.5cm}}
    \caption{\footnotesize{Comparison of performance of the retrained and unlearned (with enforcer) models across different batch sizes of $\mathcal{D}^f$ (N($\mathcal{D}^f$)) in the `Criteo' dataset, reported as mean (std) over 5 iterations.}}\label{tab:enf_perf_criteo}\\
   \toprule
    ppc	&	N($\mathcal{D}^f$)	&	Subset & Model	& \% Bal. Acc	& \% Gained	&	\% Lost	& \% Retained	\\
\toprule
1	&	1147	&	retraining	&	M1	&	55.41 (0.35)	&	0.04 (0.02)	&	0.06 (0.03)	&	99.94 (0.03)	\\	
	&		&		&	M2	&	55.95 (0.01)	&	0.0 (0.0)	&	0.0 (0.0)	&	100.0 (0.0)	\\	\cmidrule{3-8}
	&		&	test	&	M1	&	55.47 (0.36)	&	0.04 (0.02)	&	0.06 (0.03)	&	99.94 (0.03)	\\	
	&		&		&	M2	&	56.0 (0.01)	&	0.0 (0.0)	&	0.0 (0.0)	&	100.0 (0.0)	\\	\cmidrule{3-8}
	&		&	unlearned	&	M1	&	52.86 (0.91)	&	0.03 (0.05)	&	0.06 (0.1)	&	99.94 (0.1)	\\	
	&		&		&	M2	&	53.65 (1.28)	&	0.0 (0.0)	&	0.0 (0.0)	&	100.0 (0.0)	\\	\cmidrule{2-8}
	&	114633	&	retraining	&	M1	&	55.45 (0.92)	&	0.07 (0.03)	&	0.09 (0.04)	&	99.91 (0.04)	\\	
	&		&		&	M2	&	55.94 (0.01)	&	0.0 (0.0)	&	0.0 (0.0)	&	100.0 (0.0)	\\	\cmidrule{3-8}
	&		&	test	&	M1	&	55.51 (0.93)	&	0.07 (0.03)	&	0.09 (0.04)	&	99.91 (0.04)	\\	
	&		&		&	M2	&	55.99 (0.01)	&	0.0 (0.0)	&	0.0 (0.0)	&	100.0 (0.0)	\\	\cmidrule{3-8}
	&		&	unlearned	&	M1	&	55.73 (0.95)	&	0.08 (0.03)	&	0.09 (0.06)	&	99.91 (0.06)	\\	
	&		&		&	M2	&	56.19 (0.14)	&	0.0 (0.0)	&	0.0 (0.0)	&	100.0 (0.0)	\\	\cmidrule{2-8}
	&	1146327	&	retraining	&	M1	&	56.05 (0.19)	&	0.05 (0.03)	&	0.05 (0.03)	&	99.95 (0.03)	\\	
	&		&		&	M2	&	55.95 (0.04)	&	0.0 (0.0)	&	0.0 (0.0)	&	100.0 (0.0)	\\	\cmidrule{3-8}
	&		&	test	&	M1	&	56.1 (0.19)	&	0.05 (0.03)	&	0.05 (0.03)	&	99.95 (0.03)	\\	
	&		&		&	M2	&	56.01 (0.03)	&	0.0 (0.0)	&	0.0 (0.0)	&	100.0 (0.0)	\\	\cmidrule{3-8}
	&		&	unlearned	&	M1	&	56.06 (0.22)	&	0.05 (0.03)	&	0.05 (0.04)	&	99.95 (0.04)	\\	
	&		&		&	M2	&	55.97 (0.06)	&	0.0 (0.0)	&	0.0 (0.0)	&	100.0 (0.0)	\\	\hline
2	&	1147	&	retraining	&	M1	&	55.78 (0.61)	&	0.08 (0.01)	&	0.08 (0.03)	&	99.92 (0.03)	\\	
	&		&		&	M2	&	55.89 (0.01)	&	0.0 (0.0)	&	0.0 (0.0)	&	100.0 (0.0)	\\	\cmidrule{3-8}
	&		&	test	&	M1	&	55.8 (0.61)	&	0.09 (0.01)	&	0.08 (0.03)	&	99.92 (0.03)	\\	
	&		&		&	M2	&	55.91 (0.01)	&	0.0 (0.0)	&	0.0 (0.0)	&	100.0 (0.0)	\\	\cmidrule{3-8}
	&		&	unlearned	&	M1	&	57.67 (1.11)	&	0.12 (0.11)	&	0.09 (0.16)	&	99.91 (0.16)	\\	
	&		&		&	M2	&	57.46 (0.08)	&	0.0 (0.0)	&	0.0 (0.0)	&	100.0 (0.0)	\\	\cmidrule{2-8}
	&	114633	&	retraining	&	M1	&	56.83 (0.57)	&	0.15 (0.02)	&	0.13 (0.01)	&	99.87 (0.01)	\\	
	&		&		&	M2	&	55.88 (0.02)	&	0.0 (0.0)	&	0.0 (0.0)	&	100.0 (0.0)	\\	\cmidrule{3-8}
	&		&	test	&	M1	&	56.88 (0.54)	&	0.15 (0.02)	&	0.13 (0.02)	&	99.87 (0.02)	\\	
	&		&		&	M2	&	55.9 (0.02)	&	0.0 (0.0)	&	0.0 (0.0)	&	100.0 (0.0)	\\	\cmidrule{3-8}
	&		&	unlearned	&	M1	&	56.8 (0.47)	&	0.15 (0.03)	&	0.13 (0.02)	&	99.87 (0.02)	\\	
	&		&		&	M2	&	55.86 (0.16)	&	0.0 (0.0)	&	0.0 (0.0)	&	100.0 (0.0)	\\	\cmidrule{2-8}
	&	1146327	&	retraining	&	M1	&	56.65 (1.61)	&	0.22 (0.08)	&	0.22 (0.1)	&	99.78 (0.1)	\\	
	&		&		&	M2	&	55.88 (0.04)	&	0.0 (0.0)	&	0.0 (0.0)	&	100.0 (0.0)	\\	\cmidrule{3-8}
	&		&	test	&	M1	&	56.69 (1.62)	&	0.22 (0.08)	&	0.21 (0.09)	&	99.79 (0.09)	\\	
	&		&		&	M2	&	55.9 (0.03)	&	0.0 (0.0)	&	0.0 (0.0)	&	100.0 (0.0)	\\	\cmidrule{3-8}
	&		&	unlearned	&	M1	&	56.67 (1.55)	&	0.22 (0.08)	&	0.22 (0.1)	&	99.78 (0.1)	\\	
	&		&		&	M2	&	55.91 (0.13)	&	0.0 (0.0)	&	0.0 (0.0)	&	100.0 (0.0)	\\	\bottomrule

\end{longtable}
}

\end{document}